%% file: main.tex
\documentclass[10pt,twocolumn,letterpaper]{article}

\usepackage[pagenumbers]{paperstyle}

\input{preamble}

\definecolor{linkblue}{rgb}{0.21,0.49,0.74}
\usepackage[pagebackref,breaklinks,colorlinks,allcolors=linkblue]{hyperref}

\title{LeM\'emoire: Memory-Conditioned JEPA World Models for Planning in Changing Worlds}

\author{Sadman Rakib Pinon$^{1,2}$ \quad Adamo Orsini$^{1,2}$ \quad Linlian Jiang$^{1,2}$ \quad Zhixiang Chi$^{3}$\\
Jacob Kingi$^{1,2}$ \quad Yang Wang$^{1,2\dagger}$ \quad Xinxin Zuo$^{1\dagger}$\\[2pt]
$^{1}$Concordia University \qquad $^{2}$Mila -- Quebec AI Institute \qquad $^{3}$University of Toronto\\[1pt]
{\small $^{\dagger}$Corresponding authors. \quad \texttt{sadmanrakib.pinon@mail.concordia.ca}}\\[1pt]
{\small Project page: \url{https://sadmanpinon.github.io/lememoire/}}
}
\hypersetup{pdftitle={LeM\'emoire: Memory-Conditioned JEPA World Models for Planning in Changing Worlds},pdfauthor={Sadman Rakib Pinon, Adamo Orsini, Linlian Jiang, Zhixiang Chi, Jacob Kingi, Yang Wang, Xinxin Zuo},pdfsubject={},pdfkeywords={}}

\begin{document}
\maketitle
\addtocontents{toc}{\protect\setcounter{tocdepth}{-1}}
\input{figures/teaser}
\input{sections/0_abstract}
\input{sections/1_intro}
\input{sections/2_related}
\input{sections/3_method}

\input{sections/4_experiments}
\input{sections/5_conclusion}

{
    \small
    \bibliographystyle{ieeenat_fullname}
    \bibliography{main}
}

\input{sections/X_suppl}
\end{document}

%% file: preamble.tex
\usepackage{cuted}

\usepackage{currfile}

\usepackage{caption}

\usepackage{tocloft}
\newcommand{\pmstd}[1]{{\scriptsize\,$\pm$\,#1}}
\usepackage{colortbl}
\newcommand{\grpbest}{\cellcolor{gray!15}}
\usepackage{array}
\newcolumntype{L}[1]{>{\raggedright\arraybackslash}p{#1}}

\usepackage{xspace}
\newcommand{\lememoire}{LeM\'emoire\xspace}
\newcommand{\bench}{ShiftedWorlds\xspace}
\newcommand{\dcode}{dynamics code\xspace}
\newcommand{\tcode}{teacher code\xspace}

\usepackage{pgfplots}
\pgfplotsset{compat=1.16}
\usepgfplotslibrary{fillbetween}
\usepackage{pifont}
\usetikzlibrary{arrows.meta}
\usepgfplotslibrary{groupplots}
\usepackage{algorithm}
\usepackage{algpseudocode}

\makeatletter
\renewcommand\paragraph{\@startsection{paragraph}{4}{\z@}{1.5ex \@plus .5ex \@minus .2ex}{-1em}{\normalfont\normalsize\bfseries}}
\makeatother

\AtBeginDocument{%
  \crefname{appendix}{App.}{Apps.}%
  \Crefname{appendix}{App.}{Apps.}%
  \crefname{section}{Sec.}{Secs.}%
  \Crefname{section}{Sec.}{Secs.}%
}

\usepackage{needspace}
\makeatletter
\def\paperstylesection{\@startsection {section}{1}{\z@}{-10pt}{7pt} {\large\bf}}
\def\paperstylesubsection{\@startsection {subsection}{2}{\z@}{-8pt}{5pt} {\elvbf}}
\renewcommand\paragraph{\@startsection{paragraph}{4}{\z@}{1.5ex}{-1em}{\normalfont\normalsize\bfseries}}
\makeatother

%% file: figures/teaser.tex
\begin{strip}
\centering
\includegraphics[width=\linewidth]{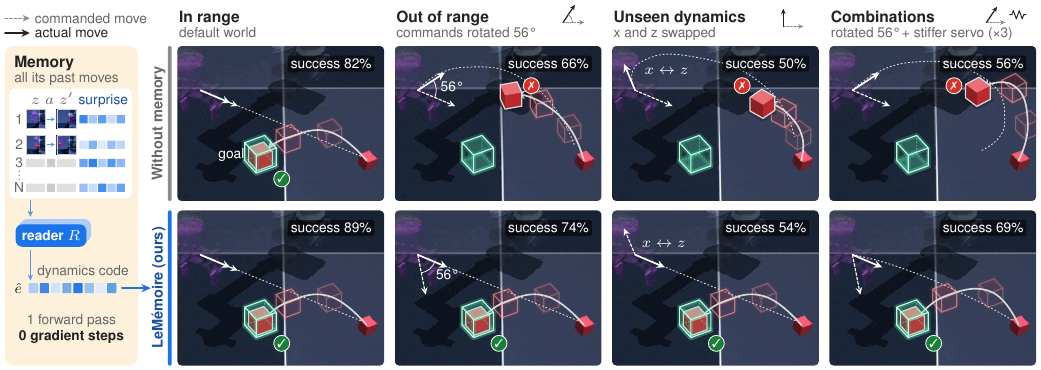}
\captionsetup{hypcap=false}
\captionof{figure}{\textbf{\lememoire} adapts a JEPA world model to dynamics shifts by reading a memory of its own transitions.
Each column is one illustrative condition of OGBench-Cube~\cite{ogbench_park2025}, with the same start and goal (green cube): the training condition and three dynamics shifts.
Top: the same world model trained without memory. Under rotated or swapped commands, its executed moves (solid arrows) deviate from the commanded ones (dashed), and the cube misses the goal.
Bottom: \lememoire infers each condition's \dcode from the transitions in its memory $(z_{t-1}, a_{t-1}) \rightarrow z_t$ in a single forward pass, without gradient steps, and adjusts its commands.
}
\captionsetup{hypcap=true}
\label{fig:\currfilebase}%
\end{strip}

%% file: sections/0_abstract.tex
\begin{abstract}
JEPA world models plan by predicting the consequences of actions in a learned latent space.
However, the dynamics of an environment often depend on factors the agent cannot observe, such as friction or the response of its actuators.
Because these models predict from only a few recent frames, they cannot distinguish environments that differ only in these factors, and their plans degrade under dynamics shifts.
We propose \lememoire, a JEPA world model that plans by conditioning its predictor on a memory of every transition it has executed in its current world, kept across episodes.
As the agent acts, the memory stores each action together with the outcome it produced. From these transitions the model reads a \dcode, which conditions its predictions of the effects of its next actions.
At test time, the model adapts by writing its transitions into its memory, without {accessing} the hidden factors or changing any weight.
In three environments, memory raises planning success under dynamics shifts outside the training range above that of the same model trained on the same data without memory.
On PushT, it raises out-of-range success from 50\% to 72\% and success under unseen dynamics by 18 points, and it significantly outperforms test-time training out of range and under combined shifts in 11 of 12 paired comparisons across environments and backbones. A history encoder limited to the current episode recovers less than half of the gain, so the gain comes from the experience kept across episodes.
A JEPA world model can thus adapt to dynamics shifts from a memory of its own transitions.
\end{abstract}

%% file: sections/1_intro.tex
\section{Introduction}
\label{sec:intro}

Latent world models built on joint-embedding predictive architectures (JEPAs) learn from reward-free offline trajectories to predict how an action changes a compact latent state.
They solve navigation and manipulation tasks from pixels by searching with model-predictive control (MPC) for the actions whose predicted outcome reaches a goal image~\cite{lecun2022path,zhou2025dinowm,sobal2025pldm,assran2025vjepa2,maes2026lewm,wang2026straightening,terver2026drives}.
Most of them are trained and tested under one fixed set of dynamics~\cite{liu2026sgjepa}, and their generalization tests change the layout of walls and mazes or the shape of the pushed object~\cite{zhou2025dinowm,sobal2025pldm}.
Shifts in the dynamics themselves have been studied only recently and on narrow sets of shifts: mass and damping in one maze~\cite{wang2026adajepa,mete2026optta}, remapped actions and contacts~\cite{zhang2026jepattt}, or a single physical parameter given to the model~\cite{liu2026sgjepa}, which leaves inferring such parameters from observation open.

In practice, dynamics can shift far more widely: any actuator, contact or object may change its response~\cite{zhang2026jepattt}, for instance when a floor becomes more slippery or a motor responds along a rotated axis (\cref{fig:teaser}).
Neither change is visible in a single frame, so the same observation and action lead to a different outcome, and MPC driven by the old predictions optimizes the wrong objective~\cite{wang2026adajepa}.
Training under many dynamics, as in dynamics randomization~\cite{peng2018dr}, does not remove the problem, because a deterministic JEPA predictor trained with a regression loss learns the conditional mean of the futures that its inputs allow.
Its prediction averages over every setting of the hidden dynamics that a few frames of history cannot distinguish~\cite{smirnov2026headroom,luo2026physicalworld}, so it is biased wherever the dynamics differ from that average.

A deployed agent does not know its dynamics, yet every action it executes is followed by an outcome, and the gap between the predicted and the observed outcome is evidence about its current world.
Test-time training uses that gap to update the model's weights on recent transitions, within each episode~\cite{nagabandi2019learning,wang2026adajepa,mete2026optta} or persistently across episodes~\cite{zhang2026jepattt}, which requires optimization while acting.
Context inference infers a latent context from past transitions in a forward pass: from a short window of recent transitions~\cite{yu2017uposi,lee2020cadm,kumar2021rma,roeder2025dali}, from a few transitions sampled in the test environment~\cite{evans2022iida}, or from earlier episodes in meta-reinforcement learning~\cite{rakelly2019pearl}.
Yet the JEPA predictors we build on condition only on the last few latent states~\cite{zhou2025dinowm,maes2026lewm}.

We argue that hidden dynamics are a fixed property of the current {\emph{world}, an environment under one fixed value of its hidden dynamics}, and that a world model should (i) infer a \dcode for them from the outcomes of its own actions and (ii) predict with that code.
Because the dynamics stay fixed while the agent remains in a world, the outcomes of every episode in that world inform the same code.
This matches the role of the world model in the blueprint for autonomous agents of~\cite{lecun2022path}, where it estimates the information about the world that perception does not provide, supported by a short-term memory.
We propose \lememoire, a JEPA world model that adapts to new dynamics without gradient steps (\cref{fig:overview}).
Concretely, a memory stores each executed transition as a row, a reader maps these rows to a \dcode, and the code conditions the predictor.
In training, every sequence comes from one of many worlds, and the memory holds earlier transitions from the same world.

At test time, in worlds both inside and beyond the training ranges, the dynamics are unknown and no weight changes.
The agent writes each new transition into its memory, which persists while it stays in one world~\cite{zhang2026jepattt,duan2016rl2}, and reads an updated code before every plan.

{Other world models also use memory: recurrent models carry the history of an episode in their state~\cite{hafner2019planet}, and the concurrent MemWM adds to a JEPA world model slots learned in training and fixed at test time~\cite{sun2026memwm}.%
}
On the same backbone, a history encoder limited to the current episode and test-time training both fall short of our memory (\cref{tab:ablations}).

We add \lememoire to two JEPA world models, LeWM~\cite{maes2026lewm} and temporal straightening (TS)~\cite{wang2026straightening}, and evaluate it on PushT, Reacher and OGBench-Cube~\cite{ogbench_park2025} under changes of friction, damping, inertia, actuator gain and delay, and under remapped commands.
On PushT, out-of-range success rises from 50\% for the same model trained on the same worlds without memory to 72\% with {\color{black}memory}.
\lememoire also outperforms test-time training~\cite{wang2026adajepa} applied to the same backbone out of range and under combined shifts, in all three environments and on both backbones (\cref{tab:main_results}).
Our main contributions are:
\begin{itemize}
\item We introduce \lememoire, a JEPA world model whose predictor is conditioned on a memory of its own transitions kept across episodes. From this memory, it infers a \dcode for its current world, with no weight updates and no dynamics parameters at test time.\looseness=-1
\item We build \bench, to our knowledge the broadest evaluation of JEPA world-model planning under unknown dynamics shifts: 82 test conditions in three standard benchmarks, grouped into shifts inside the training range, outside it, along factors never varied in training, and combinations.\looseness=-1
\item We show, on two JEPA backbones and three environments, that inference from memory outperforms test-time training out of range and under combined shifts.
\end{itemize}

%% file: sections/2_related.tex
\section{Related Work}
\label{sec:related}
\input{figures/overview}

\paragraph{JEPA world models.}
JEPA world models learn an encoder and an action-conditioned predictor from reward-free offline trajectories by predicting future latent states, and plan with MPC toward the latent state of a goal image~\cite{lecun2022path,zhou2025dinowm,sobal2025pldm,assran2025vjepa2,maes2026lewm,wang2026straightening,terver2026drives}.
They are frozen after training, so the dynamics they predict are fixed in their weights. Physical parameters can be added to the predictor's input when they are known~\cite{liu2026sgjepa}.
The concurrent MemWM adds a memory to a JEPA world model: a bank of learned slots, retrieved with the current latent state and fixed at test time, conditions the predictor and improves long-horizon planning on benchmarks whose dynamics do not change~\cite{sun2026memwm} (\cref{sec:supp_memories}). Remembered latent transitions have also modulated a JEPA predictor for prediction, without actions or planning~\cite{pandya2026epm}.
The memory of \lememoire instead holds the transitions that the agent executes in its current world, written as it acts, and the model reads a \dcode for that world from them.

\paragraph{Adapting world models at test time.}
A model that encounters new dynamics can also be updated on its recent transitions, as in online adaptation of dynamics models from a meta-learned initialization~\cite{nagabandi2019learning}.
For JEPA world models, AdaJEPA takes a gradient step on recent transitions at every replanning step and resets the model after each episode~\cite{wang2026adajepa}.
Concurrent work adds an optimistic planning term to this objective~\cite{mete2026optta}, keeps the updates across episodes~\cite{zhang2026jepattt}, learns small residual corrections around a frozen predictor~\cite{soni2026sandwich}, or keeps a video world model frozen and optimizes a low-dimensional environment latent on one or two exploratory trajectories~\cite{zhou2026t2z}.
All of these optimize at test time, which adds a learning rate and a number of steps to tune. \lememoire replaces the optimization with one forward pass of its reader and leaves the weights unchanged.

\input{figures/ph_shift_strip}
\paragraph{Inferring dynamics from experience.}
Domain and dynamics randomization train one model across many simulated variations~\cite{tobin2017dr,peng2018dr}.
Contextual and hidden-parameter Markov decision processes instead describe a family of environments whose dynamics depend on a hidden context~\cite{hallak2015cmdp,doshivelez2016hipmdp,killian2017hipmdp}, and a long line of work infers this context from recent experience.
Some methods estimate physical parameters online~\cite{yu2017uposi} or learn, from parameters available only in training, an encoding that an adaptation module recovers from history~\cite{kumar2021rma}. Others learn the context without labels, for dynamics models~\cite{nagabandi2019learning,lee2020cadm,seo2020tmcl,evans2022iida,palafox2026claw,dai2026sprii} or for meta-reinforcement learning~\cite{duan2016rl2,rakelly2019pearl,zintgraf2019cavia} (\cref{sec:supp_memories,sec:supp_efficiency}).
Recurrent latent world models carry the history in their state~\cite{hafner2019planet}, and recent Dreamer-style models infer an explicit context from pixels~\cite{roeder2025dali,wang2022protocad} or receive it as an input~\cite{prasanna2024crssm}. A vision-language-action policy infers its system configuration from probing actions before the task~\cite{wang2026icwm}.
None of these methods has been evaluated with a JEPA world model that plans toward goal images. \lememoire belongs to this family and differs in the span of experience its \dcode is read from and in what that code conditions.
History encoders read a short window of recent transitions~\cite{kumar2021rma,lee2020cadm}. Meta-reinforcement learning keeps earlier episodes for a policy~\cite{duan2016rl2,rakelly2019pearl}, and long-context world models keep them for prediction without planning~\cite{wang2026iclwm}. In-context reinforcement learning has also planned with dynamics learned in context from earlier trials~\cite{son2025dicp}. It learns from reward-labelled histories of low-dimensional states, whereas \lememoire plans from images without rewards under hidden dynamics.
\lememoire keeps {\color{black}earlier episodes for its planner's predictor}.\looseness=-1

Closest to our work is the concurrent PhysiCalWorld, which conditions DINO-WM or LeWM on a token inferred from interaction in the test scene, without dynamics parameters and without gradient steps at test time~\cite{luo2026physicalworld}.
Its token is computed once from a clip of exploratory actions collected before the task, and its test parameters are held out from inside the training ranges, with extrapolation measured for open-loop prediction along one parameter~\cite{luo2026physicalworld}.
\lememoire instead reads its \dcode from the transitions that the agent executes during its episodes, with no exploration phase, and updates it before every plan as the memory grows.
It can be trained without dynamics parameters or, where they are known, guided by a \tcode computed from them, and we evaluate planning outside the training range, along factors never varied in training, under remapped commands and under combined shifts.
We compare with PhysiCalWorld on the same backbone, data and planner (\cref{sec:experiments}), given either its exploratory clip or a clip of the agent's own earlier transitions (\cref{sec:supp_physicalworld}).

%% file: figures/overview.tex
\begin{figure*}[t]
\centering
\includegraphics[width=0.97\linewidth]{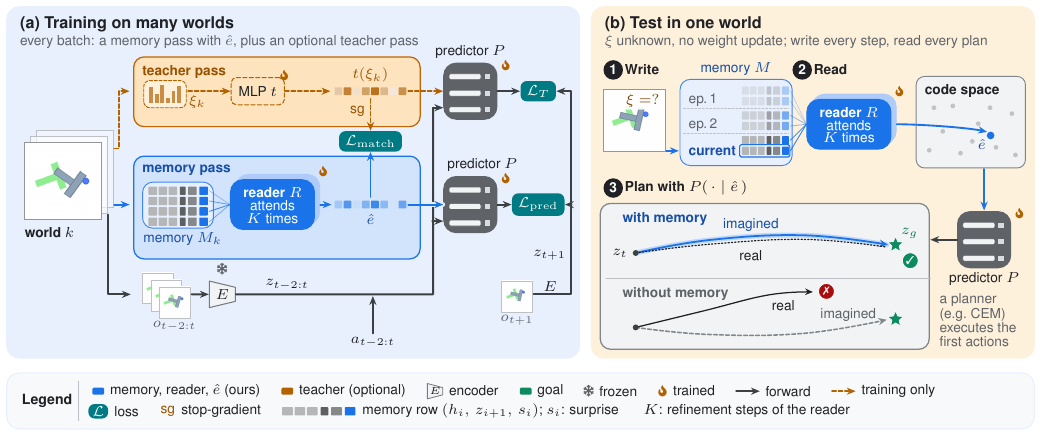}
\caption{\textbf{Overview of \lememoire.} \textbf{(a)}~Training on many worlds. The frozen encoder $E$ maps frames to latent states $z$. The memory $M$ holds earlier transitions from the same world. Each row stores three latent states, their actions, the next latent state {and its surprise $s_i$ (the error of the default-world predictor $P_0$)}. The reader $R$ maps $M$ to a \dcode $\hat e$, which conditions the predictor $P$ ($\mathcal{L}_{\mathrm{pred}}$). Optionally, where the training worlds expose their dynamics parameters $\xi$, a \tcode $t(\xi)$ also trains $P$ ($\mathcal{L}_T$) and is the target of $\hat e$ ($\mathcal{L}_{\mathrm{match}}$, sg: stop-gradient). \textbf{(b)}~Test in one world: $\xi$ is unknown and no weights change. The agent writes every executed transition into $M$, which persists across episodes, and reads an updated $\hat e$ before every plan with $P(\cdot \mid \hat e)$.}
\label{fig:\currfilebase}
\end{figure*}

%% file: figures/ph_shift_strip.tex
\begin{figure}[t]
\centering
\captionsetup{skip=3pt}%
\input{figures/lm_figstyle}
\def\fthreebadge{}
\definecolor{fthreeok}{HTML}{009E73}\definecolor{fthreebad}{HTML}{D55E00}
\definecolor{fthreeink}{HTML}{1F2933}\definecolor{fthreemute}{HTML}{6B7785}
\definecolor{fthreetagin}{HTML}{DCE9F2}\definecolor{fthreetagout}{HTML}{F6E3D3}\definecolor{fthreetagdef}{HTML}{E9ECEF}
\definecolor{fthreebar}{HTML}{7D9DB3}
\tikzset{fthreearrow/.style={-{Stealth[round, length=2.6mm, width=2.2mm]}, line width=1.1pt, line cap=round, fthreeink},
         fthreehalo/.style={-{Stealth[round, length=3.1mm, width=2.8mm]}, line width=2.6pt, line cap=round, white, opacity=0.9}}
\tikzset{fthreearrowS/.style={-{Stealth[round, length=1.6mm, width=1.35mm]}, line width=0.7pt, line cap=round, fthreeink},
         fthreehaloS/.style={-{Stealth[round, length=1.9mm, width=1.7mm]}, line width=1.6pt, line cap=round, white, opacity=0.9}}
\newlength\fthreeimg
\newcommand{\fthreepanel}[9]{\begin{tikzpicture}[x=\fthreeimg/600, y=\fthreeimg/600, font=\sffamily]
  \path (0,0) ++(0pt,-12.5pt) coordinate (bl);
  \begin{scope}
    \clip[rounded corners=3pt] (bl) rectangle (600,549);
    \fill[white] (bl) rectangle (600,549);
    \node[inner sep=0, anchor=south west] at (0,-1) {\includegraphics[width=\fthreeimg]{figures/fig3p/#1.png}};
    \node[inner sep=0, anchor=south west, opacity=0.45] at (0,-1) {\includegraphics[width=\fthreeimg]{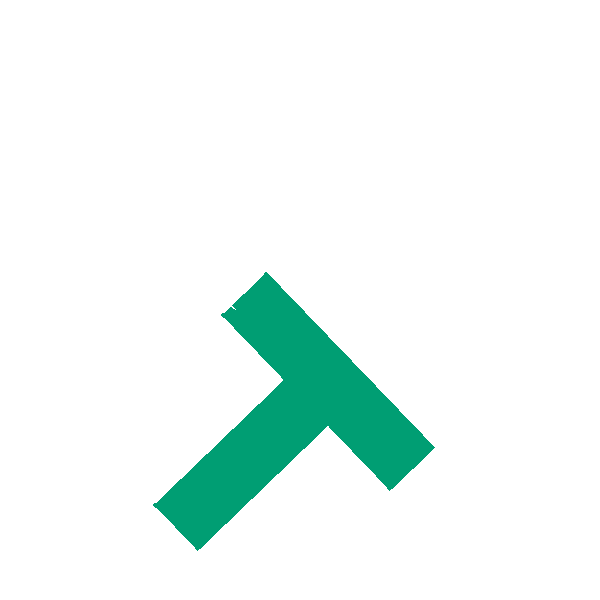}};
    \node[inner sep=0, anchor=south west, opacity=0.22] at (0,-1) {\includegraphics[width=\fthreeimg]{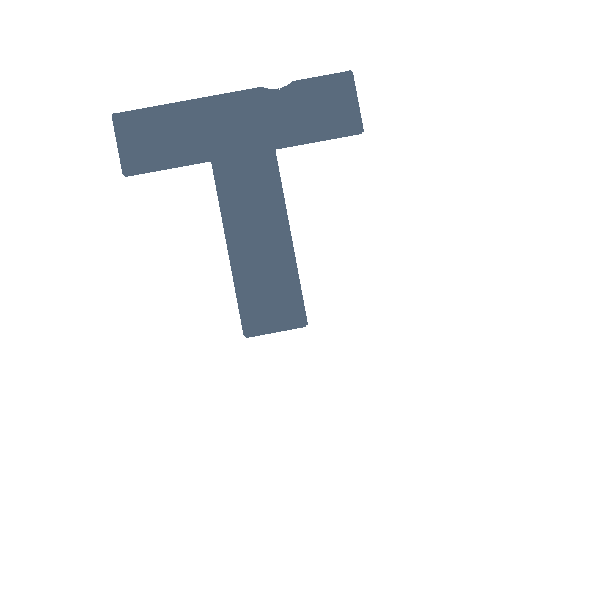}};
    \draw[fthreehaloS] (231,599-93) -- (#8,599-#9);
    \draw[fthreearrowS] (231,599-93) -- (#8,599-#9);
    \fill[white] (231,599-93) circle (16); \fill[fthreeink] (231,599-93) circle (10.5);
    \fill[white] (bl) rectangle (600,0);
    \draw[black!10, line width=0.4pt] (0,0) -- (600,0);
    \node[fill=#5, rounded corners=1.6pt, inner xsep=2pt, inner ysep=1.2pt, text=white, font=\fontsize{6}{7.2}\sffamily\bfseries] at ($(bl)!0.5!(600,0)$) {#6};
  \end{scope}
  \draw[black!14, line width=0.5pt, rounded corners=3pt] (bl) rectangle (600,549);
  \node[anchor=south, fill=#4, rounded corners=1.4pt, inner xsep=2.2pt, inner ysep=1.1pt, font=\fontsize{6.5}{7.8}\sffamily, text=fthreeink!85]
        (tag) at ([yshift=2.5pt]300,549) {\vphantom{gy}#3};
  \node[anchor=base, inner sep=0pt, font=\fontsize{8}{9.6}\sffamily\bfseries, text=fthreeink] (t) at ([yshift=4.8pt]tag.north) {#2};
  \ifx\fthreebadge\empty\else\node[anchor=base west, font=\lmtitle, text=lmink, inner sep=0pt] at (0,0 |- t.base) {(\fthreebadge)};\fi
  \path ([yshift=-17.5pt]0,0) coordinate (b0) ++(40pt,0) coordinate (b1); \path ($(b0)!#7/100!(b1)$) coordinate (bf);
  \fill[black!7, rounded corners=1.2pt] ([yshift=-2pt]b0) rectangle ([yshift=2pt]b1);
  \fill[fthreebar, rounded corners=1.2pt] ([yshift=-2pt]b0) rectangle ([yshift=2pt]bf);
  \node[anchor=west, inner sep=0pt, font=\fontsize{7}{8.4}\sffamily\bfseries, text=fthreeink] at ([xshift=2.5pt]b1) {#7\%};
\end{tikzpicture}}
\setlength\fthreeimg{0.2415\linewidth}%
\def\fthreebadge{a}\fthreepanel{default_end}{Default}{training}{fthreetagdef}{fthreeok}{\ding{51}\,on target}{79}{349}{360}\def\fthreebadge{}\hfill
\fthreepanel{cmd_rot30_end}{Rotated 30$^\circ$}{in range}{fthreetagin}{fthreebad}{\ding{55}\,misses target}{25}{252}{339}\hfill
\fthreepanel{kv_x25_end}{Pusher 2.5$\times$}{beyond range}{fthreetagout}{fthreebad}{\ding{55}\,misses target}{9}{261}{194}\hfill
\fthreepanel{inertia_x01_end}{Turns easily}{beyond range}{fthreetagout}{fthreebad}{\ding{55}\,misses target}{39}{219}{314}\\[3pt]
{\sffamily\scriptsize\color{fthreemute}\centering
\makebox[\linewidth]{\tikz[baseline=-0.6ex]{\fill[black!22, rounded corners=0.8pt] (0,-0.09) rectangle (0.36,0.09);}\,start\hfill \tikz[baseline=-0.6ex]{\fill[fthreeok!55, rounded corners=0.8pt] (0,-0.09) rectangle (0.36,0.09);}\,end in the default world\hfill \tikz[baseline=-0.6ex]{\fill[fthreeink] (0,0) circle (0.045); \draw[-{Stealth[round, length=1.6mm, width=1.4mm]}, line width=0.8pt, fthreeink] (0,0) -- (0.5,0);}\,block path\hfill \tikz[baseline=-0.6ex]{\fill[fthreebar, rounded corners=0.8pt] (0,-0.07) rectangle (0.36,0.07);}\,planning success}\par}\vspace{4pt}
\begin{tikzpicture}[x=1pt, y=1pt]
\lmhead{0,111.38}{b}{Novelty of each test condition}
\node[anchor=south east, inner sep=0pt, font=\fontsize{7}{8.4}\sffamily\bfseries, text=lmink] at (233.3126,99.28) {Most shifted worlds score above 75\%};
\begin{axis}[lm, anchor=south west, at={(0pt,85.94pt)}, width=8.20cm, height=10.84pt, xmin=-40, xmax=101.72, ymin=-3.35, ymax=7.49,
  ytick=\empty, xtick=\empty, axis y line=none, axis x line=none, ymajorgrids=false]
\fill[lmcard, draw=lmcardb, line width=0.4pt, rounded corners=2.5pt] (axis cs:-40,-3.35) rectangle (axis cs:101.72,7.49);
\draw[lmcardb, line width=0.5pt] (axis cs:-2.4,-3.35) -- (axis cs:-2.4,7.49);
\draw[lmgrid, line width=0.5pt] (axis cs:25,-3.35) -- (axis cs:25,7.49) (axis cs:50,-3.35) -- (axis cs:50,7.49) (axis cs:75,-3.35) -- (axis cs:75,7.49);
\node[font=\lmlab, text=lmink, anchor=west, inner sep=0pt] at (axis cs:-39.2,2.07) {In range};
\addplot[only marks, lmIn] coordinates {(0.27,0.00)(1.95,3.16)(4.57,0.00)(19.85,0.00)(26.33,0.00)(26.92,4.09)(29.08,0.00)(30.81,3.09)(34.42,0.00)(53.23,0.00)(53.66,4.14)(57.36,0.00)(68.26,0.00)(77.81,0.00)(82.98,0.00)(95.11,0.00)};
\end{axis}
\begin{axis}[lm, anchor=south west, at={(0pt,63.88pt)}, width=8.20cm, height=21.06pt, xmin=-40, xmax=101.72, ymin=-10.53, ymax=10.53,
  ytick=\empty, xtick=\empty, axis y line=none, axis x line=none, ymajorgrids=false]
\fill[lmcard, draw=lmcardb, line width=0.4pt, rounded corners=2.5pt] (axis cs:-40,-10.53) rectangle (axis cs:101.72,10.53);
\draw[lmcardb, line width=0.5pt] (axis cs:-2.4,-10.53) -- (axis cs:-2.4,10.53);
\draw[lmgrid, line width=0.5pt] (axis cs:25,-10.53) -- (axis cs:25,10.53) (axis cs:50,-10.53) -- (axis cs:50,10.53) (axis cs:75,-10.53) -- (axis cs:75,10.53);
\node[font=\lmlab, text=lmink, anchor=west, inner sep=0pt] at (axis cs:-39.2,0.00) {Out of range};
\addplot[only marks, lmOut] coordinates {(25.55,0.00)(28.46,0.00)(46.57,0.00)(47.94,3.71)(57.00,0.00)(58.82,3.71)(62.34,0.00)(68.45,0.00)(69.48,3.71)(75.05,0.00)(75.38,3.71)(76.03,-3.71)(77.75,0.00)(79.27,3.71)(81.06,0.00)(88.11,0.00)(90.44,0.00)(90.74,3.71)(91.91,-3.71)(92.23,7.42)(92.85,0.00)(93.46,3.71)(95.37,0.00)(95.86,3.71)(97.43,-3.71)(98.32,0.00)(99.38,3.71)(99.49,7.42)(99.66,-7.42)};
\end{axis}
\begin{axis}[lm, anchor=south west, at={(0pt,46.94pt)}, width=8.20cm, height=15.94pt, xmin=-40, xmax=101.72, ymin=-7.97, ymax=7.97,
  ytick=\empty, xtick=\empty, axis y line=none, axis x line=none, ymajorgrids=false]
\fill[lmcard, draw=lmcardb, line width=0.4pt, rounded corners=2.5pt] (axis cs:-40,-7.97) rectangle (axis cs:101.72,7.97);
\draw[lmcardb, line width=0.5pt] (axis cs:-2.4,-7.97) -- (axis cs:-2.4,7.97);
\draw[lmgrid, line width=0.5pt] (axis cs:25,-7.97) -- (axis cs:25,7.97) (axis cs:50,-7.97) -- (axis cs:50,7.97) (axis cs:75,-7.97) -- (axis cs:75,7.97);
\node[font=\lmlab, text=lmink, anchor=west, inner sep=0pt] at (axis cs:-39.2,0.00) {Unseen dynamics};
\addplot[only marks, lmUn] coordinates {(50.47,0.00)(51.71,4.03)(80.61,0.00)};
\addplot[only marks, lmUn, fill opacity=0.6, draw opacity=0.6] coordinates {(97.76,-5.34)(97.87,-1.32)(98.39,2.70)(98.55,-3.73)(98.66,0.29)(99.49,4.31)(99.52,-2.12)(99.52,1.89)(99.77,-4.54)(99.95,-0.52)(100.00,3.50)(100.00,-2.93)(100.00,1.09)};
\end{axis}
\begin{axis}[lm, anchor=south west, at={(0pt,30pt)}, width=8.20cm, height=15.94pt, xmin=-40, xmax=101.72, ymin=-7.40, ymax=8.54,
  ytick=\empty, xtick=\empty, axis y line=none, axis x line=none, ymajorgrids=false]
\fill[lmcard, draw=lmcardb, line width=0.4pt, rounded corners=2.5pt] (axis cs:-40,-7.40) rectangle (axis cs:101.72,8.54);
\draw[lmcardb, line width=0.5pt] (axis cs:-2.4,-7.40) -- (axis cs:-2.4,8.54);
\draw[lmgrid, line width=0.5pt] (axis cs:25,-7.40) -- (axis cs:25,8.54) (axis cs:50,-7.40) -- (axis cs:50,8.54) (axis cs:75,-7.40) -- (axis cs:75,8.54);
\node[font=\lmlab, text=lmink, anchor=west, inner sep=0pt] at (axis cs:-39.2,0.57) {Combinations};
\addplot[only marks, lmCo] coordinates {(66.54,0.00)(69.48,0.00)(69.70,4.51)(70.28,-3.28)(77.70,0.00)(77.75,4.87)(88.05,0.00)(90.44,0.00)(93.34,0.00)(95.11,1.21)(95.46,-3.02)(97.08,0.00)(97.25,4.62)(98.32,-2.34)(99.38,0.38)(100.00,-3.74)(100.00,4.04)};
\begin{scope}[every node/.style={font=\lmtick, text=lmmuted, anchor=north, inner sep=0pt, yshift=-2.5pt}]
\node at (axis cs:0,-7.40) {0}; \node at (axis cs:25,-7.40) {25}; \node at (axis cs:50,-7.40) {50}; \node at (axis cs:75,-7.40) {75}; \node at (axis cs:100,-7.40) {100};
\end{scope}
\node[font=\lmfont{6.5}{10.4}, text=lmink, anchor=north, align=center, inner sep=0pt] at (axis cs:50,-7.40) [yshift=-12.5pt]
  {Novelty (\%)};%
\end{axis}
\end{tikzpicture}\par
\caption{\textbf{Under hidden dynamics, the same actions end in a different state in each world.} (a)~PushT: one recorded action sequence replayed unchanged in the default world and in three changed worlds, with the planning success of the fine-tuned backbone, the same model trained without memory, over 150 episodes. (b) Novelty of each \bench condition in all three environments: how far its end states lie from the nearest training world, as a percentile of the distances between training worlds.}
\label{fig:shift_strip}
\end{figure}

%% file: figures/lm_figstyle.tex
\def\lmfont#1#2{\fontfamily{phv}\fontseries{m}\fontshape{n}\fontsize{#1}{#2}\selectfont}
\def\lmtick{\lmfont{6.5}{7.5}}
\def\lmlab{\lmfont{7}{8.2}}
\def\lmnote{\lmfont{6.5}{7.5}}
\def\lmmaplab{\fontfamily{phv}\fontseries{c}\fontshape{n}\fontsize{6.5}{7.2}\selectfont}
\def\lmtitle{\fontfamily{phv}\fontseries{b}\fontshape{n}\fontsize{8.2}{9.6}\selectfont}
\def\lmbadgefont{\fontfamily{phv}\fontseries{b}\fontshape{n}\fontsize{6.2}{7}\selectfont}
\definecolor{lmink}{HTML}{222222}\definecolor{lmmuted}{HTML}{666666}\definecolor{lmfaint}{HTML}{A6A6A6}
\definecolor{lmgrid}{HTML}{E8E8E8}\definecolor{lmcard}{HTML}{FFFFFF}\definecolor{lmcardb}{HTML}{D6D6D6}
\definecolor{lmours}{HTML}{1A73E8}\definecolor{lmctrl}{HTML}{8C8C8C}\definecolor{lmpanel}{HTML}{F2F2F2}
\definecolor{gIn}{HTML}{0072B2}\definecolor{gOut}{HTML}{D55E00}\definecolor{gUn}{HTML}{009E73}\definecolor{gCo}{HTML}{CC79A7}
\definecolor{lmband}{HTML}{E8F0FE}
\def\lmms{1.75}%
\def\lmlw{0.95pt}%
\def\lmciw{2.3pt}%
\tikzset{lmmk/.style={draw=white, line width=0.45pt, fill=#1},
         lmline/.style={lmfaint, line width=0.7pt},
         lmpill/.style={fill=white, draw=lmcardb, line width=0.4pt, rounded corners=1.5pt},
         lmci/.style={line width=\lmciw, line cap=round, #1!40},%
         lmIn/.style={mark=*, mark size=\lmms pt, lmmk=gIn},
         lmOut/.style={mark=square*, mark size=0.86*\lmms pt, lmmk=gOut},
         lmUn/.style={mark=triangle*, mark size=1.18*\lmms pt, lmmk=gUn},
         lmCo/.style={mark=diamond*, mark size=1.18*\lmms pt, lmmk=gCo},
         lmLo/.style={mark=o, mark size=0.92*\lmms pt, draw=lmmuted, line width=0.55pt, fill=white},
         lmOurs/.style={mark=*, mark size=\lmms pt, lmmk=lmours}}
\pgfplotsset{lm/.style={scale only axis, clip=false, tick align=outside, xtick pos=left, ytick pos=left,
  tick label style={font=\lmtick, text=lmmuted}, typeset ticklabels with strut, label style={font=\lmlab, text=lmink},
  xticklabel={\pgfmathprintnumber[assume math mode=true]{\tick}}, yticklabel={\pgfmathprintnumber[assume math mode=true]{\tick}},
  axis x line*=bottom, axis y line*=left, axis line style={lmfaint, line width=0.5pt},
  every tick/.style={lmfaint, line width=0.45pt}, major tick length=1.8pt, ymajorgrids, grid style={lmgrid, line width=0.45pt}}}
\def\lmcard#1#2{\fill[lmcard, draw=lmcardb, line width=0.5pt, rounded corners=4pt] (#1) rectangle (#2);}
\def\lmhead#1#2#3{\node[anchor=west, font=\lmtitle, text=lmink, inner sep=0pt] at (#1) {(#2)\,#3};}
\ifdefined\lmcardoverride\lmcardoverride\fi

%% file: sections/3_method.tex
\needspace{4\baselineskip}
\section{Method}
\label{sec:method}

\subsection{Problem setup}
\label{sec:setup_problem}
We consider an environment whose dynamics depend on hidden parameters $\xi$, such as friction, damping, inertia, actuator gains and delays, or remapped commands (\cref{fig:shift_strip}).
We define a \emph{world} as this environment under one fixed value of $\xi$: at each step $t$ it takes an action $a_t$ and returns an image observation $o_{t+1} \sim p_\xi(\cdot \mid o_{\le t}, a_{\le t})$, so $\xi$ indexes the worlds of a contextual Markov decision process~\cite{hallak2015cmdp} observed through images.
All worlds share the observation and action spaces, and the default world has the original dynamics of the benchmark.
We train on reward-free offline trajectories from many worlds, knowing which trajectories share a world.
At test time, the agent does not know $\xi$.
It runs a sequence of \emph{episodes} in one world, each given by a start observation $o_0$ and a goal observation $o_g$, and each test condition of \cref{sec:experiments} is one such world.

\paragraph{JEPA world model.}
We build on a JEPA world model pretrained in the default world~\cite{maes2026lewm,wang2026straightening}.
Its encoder $E$ maps observations to latent states $z_t = E(o_t)$, and its predictor $P$ predicts $\hat z_{t+1} = P(h_t)$ from a short history $h_t = (z_{t-2:t}, a_{t-2:t})$.
The agent plans by searching for the actions whose predicted rollout ends closest to the goal,
\begin{equation}
  a^\star_{t:t+H-1} = \operatorname*{arg\,min}_{a_{t:t+H-1}} \big\| \hat z_{t+H} - z_g \big\|_2^2, \qquad z_g = E(o_g),
  \label{eq:plan}
\end{equation}
and replans as it acts~\cite{zhou2025dinowm,maes2026lewm} (\cref{sec:supp_protocol}).
Under a squared loss on data pooled over worlds, the best predictor of this form averages over the worlds consistent with its inputs,
\begin{equation}
  P^\star(h_t) = \mathbb{E}_{\xi \sim p(\xi \mid h_t)} \big[\, \mathbb{E}[z_{t+1} \mid h_t, \xi] \,\big],
  \label{eq:mean}
\end{equation}
so its plans degrade wherever the current world departs from this average.
{The predictor's input $h_t$ holds only states of the current episode, observed or predicted.}
{The transitions the agent has executed in the world also span earlier episodes and carry the information about $\xi$ that $h_t$ lacks.
They reach the predictor only through a summary.}

\input{tables/main_results}
\subsection{Memory and dynamics code}
\label{sec:reader}
\lememoire summarizes these transitions in a \dcode $\hat e$ that conditions the predictor (\cref{fig:overview}),
\begin{equation}
  \hat z_{t+1} = P(h_t \mid \hat e), \qquad \hat e = R(M),
  \label{eq:cond}
\end{equation}
where $M$ is a memory of transitions from the current world and $R$ is a reader. The best predictor of this form averages over $p(\xi \mid h_t, \hat e)$ instead of $p(\xi \mid h_t)$ in \cref{eq:mean}, so the better the code distinguishes worlds, the closer its predictions are to those of the current world.

\paragraph{Memory.}
$\xi$ acts on the dynamics, so it is reflected in the outcomes of actions. Each row of $M$ therefore stores one executed transition and its surprise {$s_i$, the error of the fixed pretrained (default-world) predictor $P_0$},
\begin{equation}
  m_i = \big(h_i,\, z_{i+1},\, s_i\big), \qquad s_i = z_{i+1} - P_0(h_i),
  \label{eq:row}
\end{equation}
Rows hold no goal, no success signal and no $\xi$, so transitions recorded for one goal also inform later plans.\looseness=-1

\paragraph{Reader.}
Each row holds a whole transition and $\xi$ is fixed within a world, so $R$ treats $M$ as a set. It embeds the rows with a shared MLP, pools them with learned attention weights, and refines the pooled vector with $K$ steps of a weight-tied cross-attention to the rows ($K = 8$, or $0$ in our LeWM model on PushT). A 16-dimensional bottleneck maps the result to $\hat e$.
For an empty memory, $R$ outputs a learned code $\hat e_\varnothing${, which the predictor uses until the first transition is stored (\cref{tab:ablations})}. With $\hat e_\varnothing$ the predictor has only the evidence in $h_t$, like $P^\star$ in \cref{eq:mean}, and stored transitions supply the information that $h_t$ lacks (\cref{fig:onetask}).

\paragraph{Conditioning.}
{The \dcode $\hat e$ conditions the predictor (\cref{eq:cond}) through an entry point set per backbone and environment (\cref{sec:supp_architectures}). In LeWM, and in TS on Cube, $\hat e$ is added to the action embedding that the predictor reads. In TS on PushT and Reacher, gated cross-attention layers let the predictor's tokens attend to $\hat e$, chosen over adding $\hat e$ to the action embedding on validation episodes. Both entry points outperform the fine-tuned backbone on these two environments (App.~\cref{tab:supp_arch_variants}).
}

\subsection{Training on many worlds}
\label{sec:training}
We train on trajectories pooled over many training worlds, with the image encoder frozen.
Each training sample consists of a short sequence from world $k$ and a memory $M_k$ of a random number of transitions from the same world, taken from its other episodes and from the steps before the sequence, never after it.
Some samples have an empty memory, so that $\hat e_\varnothing$ is also trained.
Training minimizes
\begin{equation}
  \mathcal{L}_{\mathrm{pred}} = \big\| P\big(h_t \mid R(M_k)\big) - z_{t+1} \big\|_2^2.
  \label{eq:lpred}
\end{equation}
Only the memory distinguishes samples from worlds that are similar over a few frames, so minimizing $\mathcal{L}_{\mathrm{pred}}$ trains $R$ to extract the information that $h_t$ lacks.

\paragraph{Optional teacher code.}
When $\xi$ is known for the training worlds, a learned function of it can also supervise the code, as in RMA~\cite{kumar2021rma}.
A small MLP maps $\xi_k$ to a \tcode $t(\xi_k)$, which conditions the predictor in a teacher pass on the same batch, with loss $\mathcal{L}_T$ (\cref{eq:lpred} with $t(\xi_k)$ in place of $R(M_k)$), and the \dcode is regressed toward it,
\begin{equation}
  \mathcal{L}_{\mathrm{match}} = \big\| \hat e - \operatorname{sg}\!\big(t(\xi_k)\big) \big\|_2^2,
  \label{eq:teacher}
\end{equation}
where $\operatorname{sg}$ stops the gradient. Both terms are added to the training loss.
The predictor thus learns to use a code from the start of training, and the reader has a target code for each memory.
The teacher is optional: three of our six models are trained without it (\cref{sec:supp_implementation}), and without it our LeWM model on PushT keeps more than two thirds of its gain over the fine-tuned backbone (\cref{tab:ablations}).

\subsection{Planning with memory}
\label{sec:planning}
At test time, the agent enters a test world with an empty memory and runs its episodes sequentially (\cref{alg:test_time}).
The memory then holds the transitions of all earlier episodes in the world.
{Each executed action adds one row, and the step limit of each episode bounds the number of rows it adds.}
No weights change and $\xi$ is never used. Memory reads and writes are forward passes, and inference cost is comparable to that of the base model (App.~\cref{tab:cost}).\looseness=-1

\input{figures/alg_test_time}

%% file: tables/main_results.tex
\begin{table*}[t]
\centering
\footnotesize
\setlength{\tabcolsep}{3pt}
\resizebox{\ifdim\width>\textwidth\textwidth\else\width\fi}{!}{%
\begin{tabular}{@{}l cc@{\hspace{10pt}}cccc@{\hspace{10pt}}ccc@{}}
\toprule
& & & \multicolumn{4}{c}{\textit{LeWM backbone}} & \multicolumn{3}{c}{\textit{TS backbone}} \\
\cmidrule(lr){4-7} \cmidrule(l){8-10}
Test group & PLDM~\cite{sobal2025pldm} & GCBC~\cite{ghosh2019gcsl} & LeWM~\cite{maes2026lewm} & AdaJEPA~\cite{wang2026adajepa} & PhysiCalWorld~\cite{luo2026physicalworld} & Ours & TS~\cite{wang2026straightening} & AdaJEPA~\cite{wang2026adajepa} & Ours \\
\midrule
\multicolumn{10}{@{}l}{\textit{PushT}} \\
In range        & 61.8\pmstd{1.8} & 74.0\pmstd{7.7} & 64.9\pmstd{2.9} & 55.0\pmstd{1.9} & 82.2\pmstd{4.7} & \grpbest \textbf{86.3}\pmstd{1.3} & 80.7\pmstd{1.9} & 64.5\pmstd{4.2} & \grpbest \underline{83.0}\pmstd{2.7} \\
Out of range    & 51.7\pmstd{3.0} & 62.3\pmstd{6.2} & 54.3\pmstd{2.7} & 48.8\pmstd{0.4} & 66.4\pmstd{2.1} & \grpbest \textbf{72.3}\pmstd{3.2} & 66.2\pmstd{2.4} & 54.5\pmstd{1.6} & \grpbest \underline{66.9}\pmstd{3.3} \\
Unseen dynamics & 71.5\pmstd{1.0} & 74.3\pmstd{9.0} & 76.5\pmstd{2.3} & 65.0\pmstd{2.0} & \underline{80.0}\pmstd{6.9} & \grpbest \textbf{84.5}\pmstd{5.7} & 75.2\pmstd{6.8} & 63.5\pmstd{6.1} & \grpbest 78.7\pmstd{3.1} \\
Combinations    & 30.8\pmstd{2.0} & 27.7\pmstd{3.0} & 32.7\pmstd{2.3} & 30.0\pmstd{1.0} & 37.2\pmstd{0.8} & \grpbest \textbf{53.2}\pmstd{1.8} & \grpbest \underline{51.8}\pmstd{2.7} & 43.4\pmstd{1.7} & \grpbest \underline{51.8}\pmstd{3.3} \\
\midrule
\multicolumn{10}{@{}l}{\textit{Reacher}} \\
In range        & 56.4\pmstd{4.1} & ---$^\ddagger$ & 50.3\pmstd{3.8} & 51.9\pmstd{3.6} & 54.4\pmstd{3.2} & \grpbest \underline{80.5}\pmstd{4.4} & 66.4\pmstd{1.7} & 79.1\pmstd{3.2} & \grpbest \textbf{87.2}\pmstd{2.2} \\
Out of range    & 11.4\pmstd{2.2} & ---$^\ddagger$ & 11.9\pmstd{3.6} & 15.3\pmstd{1.4} & 15.0\pmstd{5.1} & \grpbest 28.9\pmstd{5.5} & \underline{35.0}\pmstd{4.2} & 22.7\pmstd{2.9} & \grpbest \textbf{62.3}\pmstd{1.2} \\
Unseen dynamics & 18.0\pmstd{2.6} & --- & 14.9\pmstd{1.1} & 14.7\pmstd{2.5} & 16.1\pmstd{3.6} & \grpbest \underline{35.7}\pmstd{1.9} & 18.3\pmstd{1.5} & 28.2\pmstd{2.4} & \grpbest \textbf{38.1}\pmstd{5.4} \\
Combinations    & 23.7\pmstd{4.4} & ---$^\ddagger$ & 27.7\pmstd{4.5} & 36.8\pmstd{6.3} & 33.2\pmstd{6.3} & \grpbest 39.7\pmstd{9.1} & \underline{56.8}\pmstd{3.7} & 22.0\pmstd{4.8} & \grpbest \textbf{65.3}\pmstd{1.6} \\
\midrule
\multicolumn{10}{@{}l}{\textit{Cube}} \\
In range        & 64.8\pmstd{4.3} & \textbf{93.9}\pmstd{1.9} & 71.4\pmstd{4.2} & \grpbest 87.4\pmstd{3.3} & 86.1\pmstd{2.1} & 86.4\pmstd{0.9} & 81.6\pmstd{4.0} & 81.8\pmstd{1.6} & \grpbest \underline{88.4}\pmstd{0.2} \\
Out of range    & 58.9\pmstd{2.6} & 75.9\pmstd{2.8} & 62.7\pmstd{3.2} & 71.3\pmstd{1.0} & \underline{80.0}\pmstd{2.5} & \grpbest \textbf{82.6}\pmstd{1.6} & 69.9\pmstd{4.9} & 69.9\pmstd{4.2} & \grpbest 77.7\pmstd{2.5} \\
Unseen dynamics & 51.7\pmstd{4.9} & 61.6\pmstd{3.0} & 56.2\pmstd{4.3} & \grpbest \textbf{66.2}\pmstd{0.5} & 63.4\pmstd{4.2} & 61.4\pmstd{2.6} & 60.4\pmstd{4.9} & 59.3\pmstd{2.9} & \grpbest \underline{63.8}\pmstd{3.2} \\
Combinations    & 58.4\pmstd{4.2} & \underline{80.3}\pmstd{1.0} & 61.6\pmstd{2.0} & 68.3\pmstd{2.2} & 74.1\pmstd{1.8} & \grpbest \textbf{80.4}\pmstd{2.0} & 67.1\pmstd{3.9} & 60.5\pmstd{3.9} & \grpbest 70.1\pmstd{0.2} \\
\bottomrule
\end{tabular}%
}
\caption{\textbf{Planning success (\%) in changed worlds.} LeWM and temporal straightening (TS) backbones. Mean $\pm$ std over three 50-episode blocks. Best in \textbf{bold}, second best \underline{underlined}, and the best within each backbone {\setlength{\fboxsep}{1pt}\colorbox{gray!15}{shaded}}. A dash marks a cell not measured, or ($^\ddagger$) a success below 3\%. Three training seeds of our LeWM model on PushT vary by a std of at most 0.7 points per group (\cref{sec:supp_protocol}). Ours is significantly better in over 80\% of paired comparisons (Holm-corrected McNemar).}
\label{tab:\currfilebase}
\end{table*}

%% file: figures/alg_test_time.tex
\begin{algorithm}[t]
\caption{\lememoire at test time in one world.}
\label{alg:test_time}
\begin{algorithmic}[1]
\Require encoder $E$, predictor $P$, reader $R$, default-world predictor $P_0$
\State $M \gets \emptyset$
\For{each episode with goal observation $o_g$}
  \State $z_g \gets E(o_g)$
  \While{the goal is not reached and within budget}
    \State $\hat e \gets R(M)$ \Comment{one forward pass}
    \State plan with $P(\cdot \mid \hat e)$ toward $z_g$ \Comment{\cref{eq:plan}}
    \For{each executed action $a_t$}
      \State $z_{t+1} \gets E(o_{t+1})$
      \State $M \gets M \cup \{(h_t,\, z_{t+1},\, z_{t+1} - P_0(h_t))\}$
    \EndFor
  \EndWhile
\EndFor \Comment{$M$ persists across episodes}
\end{algorithmic}
\end{algorithm}

%% file: sections/4_experiments.tex
\section{Experiments}
\label{sec:experiments}
We test whether memory improves planning under dynamics shifts (\cref{sec:main_results}), what the \dcode represents (\cref{sec:code_analysis}) and how much experience the memory needs (\cref{sec:memory_curve}), then ablate the memory (\cref{sec:ablations}) and test on real robot data (\cref{sec:realdata}).

\subsection{Experimental setup}
\label{sec:setup}
\paragraph{Environments and test conditions.}
We use PushT, Reacher and OGBench-Cube~\cite{ogbench_park2025}, whose released versions are the default worlds.
Training worlds change friction, damping, inertia, actuator gain and delay, and remap the commands, each within a training range (\cref{sec:supp_conditions}).
The 82 test conditions, which we call \bench, fall into four groups: in range (including the default world), out of range, unseen dynamics, such as table tilt or swapped commands, and combinations of several changes (\cref{fig:shift_strip}b).
Our suite has more conditions and kinds of change than the suites used to test related JEPA world models and tests both new kinds and combined shifts (\cref{tab:suites}).\looseness=1

\paragraph{Models and baselines.}
We add \lememoire to LeWM~\cite{maes2026lewm} and TS~\cite{wang2026straightening}.
The fine-tuned backbone (our control) is the same backbone fine-tuned on the training worlds without memory.
Following LeWM, we compare with PLDM~\cite{sobal2025pldm}, the offline goal-conditioned policy GCBC~\cite{ghosh2019gcsl} and the released backbones (\cref{sec:supp_baselines}).
Of the methods that adapt a JEPA planner at test time, we run those available on the same backbones. AdaJEPA~\cite{wang2026adajepa} adapts the fine-tuned backbone with its published and tuned settings (App.~\cref{tab:supp_adajepa_lr}), and the concurrent PhysiCalWorld~\cite{luo2026physicalworld} conditions LeWM on a token from an exploration clip collected before each episode (\cref{sec:supp_physicalworld}).\looseness=1

\paragraph{Protocol.}
Each condition has 150 test episodes, each with a start and a goal, solved in a fixed order with the memory kept across them.
For every method, we select the checkpoint and settings with the highest planning success on a fixed set of held-out validation episodes and test them on 150 additional episodes (\cref{sec:supp_protocol}).
We compare methods with paired exact McNemar tests, Holm-corrected. Every $\pm$ is the standard deviation over three blocks of 50 episodes.

\input{tables/suites}
\subsection{Main results}
\label{sec:main_results}
Out of range and under combined shifts, \lememoire outperforms test-time training applied to the same backbone in all three environments and on both backbones, significantly in 11 of these 12 comparisons (\cref{tab:main_results}).
Against the fine-tuned backbone, memory raises out-of-range success on PushT from 50.3\% to 72.3\% (\cref{tab:ablations}) and, with TS, on Reacher from 23.1\% to 62.3\% (\cref{sec:supp_control}).
On PushT, training on many worlds without memory lowers success below that of the released LeWM (App.~\cref{tab:supp_control}). With memory, our model exceeds the released LeWM in every group (\cref{tab:main_results}) and on every training seed (\cref{sec:supp_protocol}).
On Cube, memory adds 4.4 to 8.0 points over the fine-tuned backbone in every group with TS. With LeWM on Cube, it raises success out of range and under combinations.
On the TS backbone, \lememoire is the best method in every Reacher group. Memory thus improves planning out of range.

\needspace{4\baselineskip}
\subsection{\texorpdfstring{Interpreting the dynamics code}{Interpreting the dynamics code}}
\label{sec:code_analysis}
\input{figures/fig_dial}

The \dcode encodes several hidden factors of the current world, and the predictor uses it.
{\color{black}With all episodes in memory}, each kind of change moves its code in its own direction (App.~\cref{fig:code_map}).
For friction, damping, inertia and pusher gain, the projection of a world's code onto the first principal component of that factor's codes is monotonic in the factor, also beyond the training range (App.~\cref{fig:code_map}b).
To understand what the code encodes, we fit a linear probe.
A flexible probe can fit even random labels~\cite{hewitt2019control}, so what a linear probe recovers is attributable to the code itself.
Fit on the codes of all other worlds, the probe explains 84\%, 62\% and 57\% of the variation in a held-out world's damping, pusher gain and command lag, but not its friction (\cref{sec:supp_code_detail}).
It retains 80\% for damping and 51\% for pusher gain when no world in the fit shares the held-out value.
On Reacher, whose model trains without the teacher, the codes rank eight table tilts in their true order (\cref{sec:supp_code_detail}).

Decodability does not imply that the predictor uses a factor, so we intervene on the code alone and hold the weights, the initial state and the commands fixed.
We interpolate linearly from the default world's code ($w = 0$) to the test world's code ($w = 1$)~\cite{park2024linear}.
In each of three worlds (commands rotated by 30$^\circ$, commands scaled by 1.6, pusher gain doubled), the test world's code more than halves the median distance between the predicted and the real end of the push (\cref{fig:code_dial}).
A displacement of equal norm in a random direction leaves the predicted end farther from the real end than the test world's code does, in every draw.
Across 32 worlds, codes inferred from different five-episode memories of the same world are less consistent in worlds where planning succeeds less often (rank correlation $-0.63$, and $-0.47$ with behaviour distance from default held fixed, \cref{sec:supp_code_detail}). The code thus conveys hidden factors to the predictor.
\subsection{\texorpdfstring{Adapting from little experience}{Adapting from little experience}}
\label{sec:memory_curve}
\input{figures/fig_onetask}
\input{figures/fig_realdata}
\input{tables/ablations}
One episode of experience yields most of the gain (\cref{fig:onetask}).
In the four worlds that change the commands or the pusher gain, a single earlier episode reduces the distance between the predicted and the real end of the push to 12\,px, from 31\,px with an empty memory. All 150 episodes reduce it to 10\,px, and on average the first episode alone yields at least four fifths of the gain.

Out of range, the gain over the fine-tuned backbone is present from the second episode onward and remains near 20 points (\cref{fig:onetask}c).
Reading the last 50 transitions (at least five episodes) bounds the time per plan and keeps success within 1.3 points of the full memory and above every baseline on PushT (App.~\cref{fig:memory_cap}).
\subsection{Memory content and span}
\label{sec:ablations}
The content of the memory accounts for half to three quarters of the gain over the fine-tuned backbone.
The fine-tuned backbone loses 18.0 to 30.0 points, an empty memory loses 9.0 to 22.7 points, and the memory of another world loses up to 33.5 points, so the plans follow the world in which the memory was recorded.
Relative to an empty memory, another world's memory raises success when its code moves in the same direction as the test world's code and lowers it otherwise (Spearman $\rho = 0.71$ over 32 worlds).
The history encoder follows the recipe of history-based context inference~\cite{lee2020cadm,kumar2021rma}, with a memory of the current episode only, in training and at test. It matches ours in architecture, capacity, data, teacher and training budget.
Clearing the memory after every episode lowers success by 9.3 to 18.4 points, and the history encoder loses 9.5 to 17.0 points, so transitions from earlier episodes contribute to the gain. Kept in AdaJEPA's weights instead, this experience lowers its success (\cref{sec:supp_adajepa}).
Given all earlier episodes at test, the history encoder stays 3.8 to 14.1 points below our memory, so the gain also requires a reader trained on earlier episodes (App.~\cref{tab:supp_history_earlier}).

\subsection{\texorpdfstring{Real robot data}{Real robot data}}
\label{sec:realdata}
The memory also lowers prediction error on real robot data when added to the frozen pretrained RoboJEPA world model~\cite{zholus2026robojepa}, so without the memory its predictions are unchanged.
We test it on recorded DROID episodes~\cite{khazatsky2024droid}, where no dynamics parameters are known, and treat each scene as a world.
On 64 test episodes from a scene left out of RoboJEPA's pretraining and our training, same-scene memory lowers the wrist view's 9-step rollout error in 60 episodes and by 1.3\% on average ($p < 10^{-3}$).
This is about 60\% of the gain from fine-tuning the predictor.
Another scene's memory raises this error by 0.7\% (\cref{fig:realdata}).
As in simulation, the gain comes from the memory's content.

%% file: tables/suites.tex
\begin{table}[t]
\centering
\resizebox{\linewidth}{!}{%
\scriptsize
\setlength{\tabcolsep}{2.4pt}
\begin{tabular}{@{}lcccccc@{}}
\toprule
 & Cond. & Kinds & New & Comb. & Hidden & Ep. \\
\midrule
DINO-WM, PLDM, LeWM & 0 & \multicolumn{5}{c}{one dynamics per environment} \\
AdaJEPA~\cite{wang2026adajepa} & 2 & 2 & 2$^\dagger$ & 0 & \ding{51} & 150 \\
JEPA-TTT$^\ast$~\cite{zhang2026jepattt} & 8 & 5 & 8$^\dagger$ & 1 & \ding{51} & 100 \\
SG-JEPA~\cite{liu2026sgjepa} & 25$^\ddagger$ & 1 & 0 & 0 & \ding{55} & 200 \\
PhysiCalWorld$^\ast$~\cite{luo2026physicalworld} & 60 & 6 & 0 & 15 & \ding{51} & 5 \\
\midrule
\bench (ours) & 82 & 16 & 14 & 17 & \ding{51} & 150 \\
\bottomrule
\end{tabular}}
\caption{\textbf{Test suites for dynamics shifts in JEPA world models.} Conditions (Cond.), kinds of change, conditions with a kind absent from training (New) or several kinds at once (Comb.), hidden dynamics (Hidden) and episodes per condition (Ep.). $^\dagger$trained on one dynamics. $^\ddagger$per task. $^\ast$concurrent.}
\label{tab:suites}
\end{table}

%% file: figures/fig_dial.tex
\begin{figure*}[t]
\centering
\captionsetup{skip=3pt}%
\includegraphics[width=0.97\textwidth]{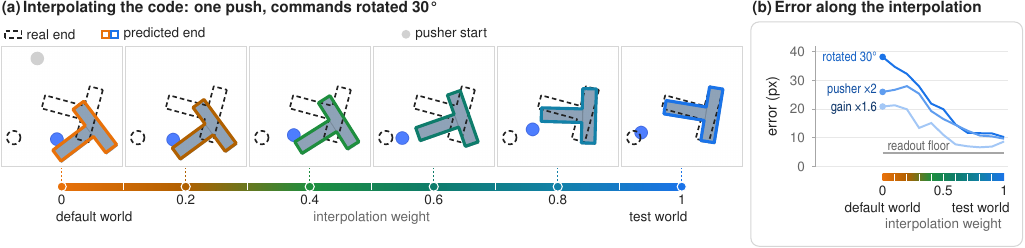}
\caption{\textbf{Push error falls as the \dcode is interpolated toward the test world's code.} (a) One push with commands rotated 30$^\circ$: the predicted end at each interpolation weight, decoded and drawn. The dashed marker is the real end, which coincides with the start because the rotated push misses the block. (b) Median distance between the predicted and the real block end against the interpolation weight. At weight 1, all ten random codes of equal magnitude end 18--57\,px from the real end, farther than the test world's code (9--10\,px) in each of the three worlds. LeWM, PushT.}
\label{fig:code_dial}
\end{figure*}

%% file: figures/fig_onetask.tex
\begin{figure*}[t]
\centering
\captionsetup{skip=3pt}%
\includegraphics[width=\textwidth]{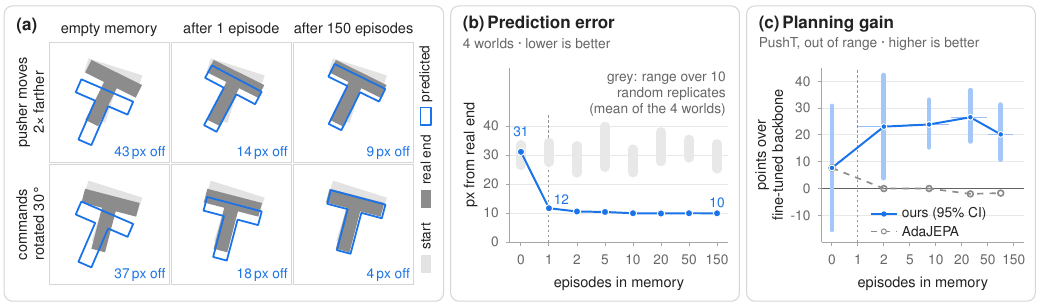}
\caption{\textbf{Prediction and planning improve within the first few episodes in memory.} (a) The block's end predicted for one push in two worlds, with an empty memory, after one episode and after 150 episodes, against its real end. (b) Median distance between the predicted and the real end, averaged over four worlds that rotate or scale the commands or strengthen the pusher. (c) Gain in planning success over the fine-tuned backbone on PushT out of range, for ours and AdaJEPA, against the number of episodes already in memory. AdaJEPA shows no gain after the first episode. LeWM, PushT.}
\label{fig:onetask}
\end{figure*}

%% file: figures/fig_realdata.tex
\begin{figure*}[t]
\centering
\captionsetup{skip=3pt}%
\input{figures/lm_figstyle}
\begin{tikzpicture}[x=1cm, y=1cm]
\lmcard{0,0}{8.95,5.22}\lmcard{9.1,0}{17.44,5.22}
\lmhead{0.22,4.96}{a}{Real DROID frames, held-out episodes}
\lmhead{9.32,4.96}{b}{}
\def\rdimg#1#2#3{\begin{scope}\clip[rounded corners=2.5pt] (#2,#3) rectangle ++(2.6,1.95);
  \node[inner sep=0, anchor=south west] at (#2,#3) {\includegraphics[width=2.6cm]{figures/droid/#1.png}};\end{scope}
  \draw[lmcardb, line width=0.5pt, rounded corners=2.5pt] (#2,#3) rectangle ++(2.6,1.95);}
\rdimg{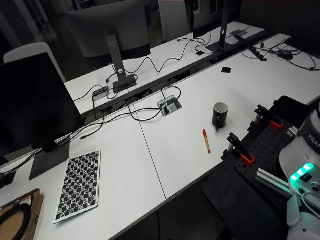}{0.62}{2.69}\rdimg{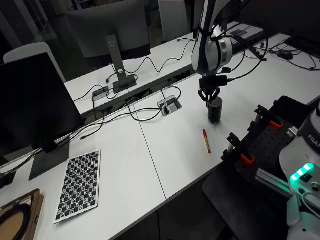}{3.34}{2.69}\rdimg{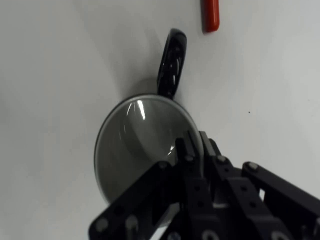}{6.06}{2.69}
\rdimg{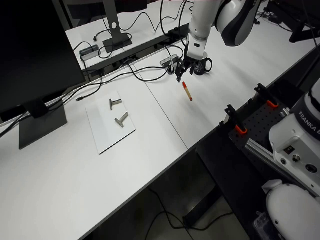}{0.62}{0.62}\rdimg{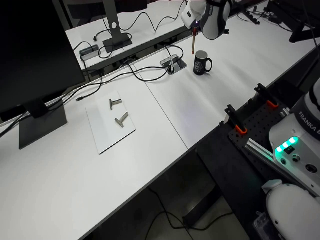}{3.34}{0.62}\rdimg{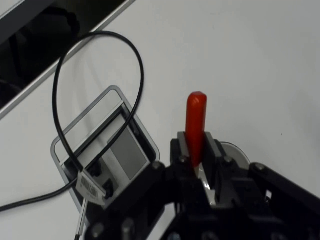}{6.06}{0.62}
\begin{scope}[every node/.style={font=\lmlab, text=lmink, inner sep=0pt}]
\node[rotate=90] at (0.38,3.665) {episode 1}; \node[rotate=90] at (0.38,1.595) {episode 2};
\node at (1.92,0.36) {start}; \node at (4.64,0.36) {goal (7.2\,s later)}; \node at (7.36,0.36) {goal (wrist view)};
\end{scope}
\begin{scope}[xshift=9.1cm]
\begin{axis}[lm, at={(1.4cm,0.62cm)}, anchor=south west, width=6.73cm, height=4.02cm, xmin=0.3, xmax=2.7, ymin=-4.6, ymax=4.4,
  xtick={0.8,2.2}, xticklabels={correct memory, wrong memory}, ytick={-4,-2,0,2,4}, xtick style={draw=none},
  yticklabels={\textminus4, \textminus2, 0, 2, 4}, ylabel={error reduction vs.\\released RoboJEPA (\%)}, ylabel style={yshift=-3pt, align=center}]
\fill[lmours!9] (axis cs:0.3,0) rectangle (axis cs:2.7,4.4);
\draw[lmmuted, dashed, line width=0.6pt] (axis cs:0.3,0) -- (axis cs:2.7,0);
\addplot[only marks, color=lmours, mark=*, mark size=1.15pt, mark options={draw opacity=0, fill=lmours, fill opacity=0.45}] coordinates {(0.65,0.616)(0.68,2.621)(0.71,0.529)(0.74,0.799)(0.77,2.095)(0.80,2.940)(0.83,1.535)(0.86,1.683)(0.89,1.090)(0.92,0.810)(0.95,2.307)(0.65,1.383)(0.68,2.428)(0.71,2.299)(0.74,1.950)(0.77,0.067)(0.80,-0.070)(0.83,0.340)(0.86,1.975)(0.89,3.414)(0.92,0.647)(0.95,1.165)(0.65,1.370)(0.68,3.030)(0.71,-0.181)(0.74,0.648)(0.77,0.580)(0.80,0.694)(0.83,1.992)(0.86,1.800)(0.89,1.928)(0.92,0.181)(0.95,1.363)(0.65,0.504)(0.68,1.818)(0.71,0.576)(0.74,0.797)(0.77,3.079)(0.80,0.952)(0.83,0.793)(0.86,0.966)(0.89,1.127)(0.92,0.979)(0.95,0.904)(0.65,0.527)(0.68,1.287)(0.71,3.451)(0.74,2.320)(0.77,1.085)(0.80,1.732)(0.83,-0.808)(0.86,2.036)(0.89,1.806)(0.92,1.362)(0.95,0.495)(0.65,1.934)(0.68,-0.367)(0.71,0.857)(0.74,2.730)(0.77,0.407)(0.80,0.398)(0.83,1.121)(0.86,0.106)(0.89,0.362)};
\addplot[only marks, color=gOut, mark=*, mark size=1.15pt, mark options={draw opacity=0, fill=gOut, fill opacity=0.45}] coordinates {(2.05,-1.796)(2.08,-0.839)(2.11,-0.751)(2.14,-0.425)(2.17,-1.037)(2.20,-0.685)(2.23,0.048)(2.26,-0.753)(2.29,0.578)(2.32,-0.409)(2.35,-0.806)(2.05,-0.905)(2.08,0.043)(2.11,-1.382)(2.14,0.428)(2.17,-0.651)(2.20,0.169)(2.23,-0.940)(2.26,-1.026)(2.29,0.082)(2.32,-1.021)(2.35,-1.092)(2.05,-0.350)(2.08,-0.433)(2.11,0.178)(2.14,-1.460)(2.17,-3.229)(2.20,-1.542)(2.23,-1.172)(2.26,-0.819)(2.29,0.978)(2.32,-1.861)(2.35,-1.224)(2.05,-0.549)(2.08,-0.438)(2.11,-2.543)(2.14,-0.614)(2.17,0.585)(2.20,-1.270)(2.23,-0.422)(2.26,-0.328)(2.29,-0.162)(2.32,-1.351)(2.35,-1.158)(2.05,-0.611)(2.08,-3.201)(2.11,0.442)(2.14,1.767)(2.17,-0.346)(2.20,-0.416)(2.23,-1.077)(2.26,0.166)(2.29,-2.425)(2.32,-0.904)(2.35,-0.815)(2.05,1.394)(2.08,-3.092)(2.11,-0.531)(2.14,0.452)(2.17,-0.992)(2.20,-0.456)(2.23,-0.930)(2.26,-0.531)(2.29,0.439)};
\draw[white, line width=4.6pt, line cap=round] (axis cs:0.8,1.035) -- (axis cs:0.8,1.508);
\draw[lmours!85!black, line width=2.8pt, line cap=round] (axis cs:0.8,1.035) -- (axis cs:0.8,1.508);
\draw[lmours!85!black, line width=1.4pt, line cap=round] (axis cs:0.54,1.271) -- (axis cs:1.06,1.271);
\draw[white, line width=4.6pt, line cap=round] (axis cs:2.2,-0.929) -- (axis cs:2.2,-0.447);
\draw[gOut!85!black, line width=2.8pt, line cap=round] (axis cs:2.2,-0.929) -- (axis cs:2.2,-0.447);
\draw[gOut!85!black, line width=1.4pt, line cap=round] (axis cs:1.94,-0.688) -- (axis cs:2.46,-0.688);
\node[font=\lmfont{6}{7}, text=lmmuted, anchor=north west, align=left, inner sep=0pt] at (axis cs:0.32,0) [yshift=-1.6pt] {released\\RoboJEPA};
\node[font=\lmnote, text=lmours!85!black, anchor=south, inner sep=0pt] at (axis cs:0.8,4.4) [yshift=2.5pt] {better in 60 of 64 episodes};
\node[font=\lmnote, text=gOut!85!black, anchor=south, inner sep=0pt] at (axis cs:2.2,4.4) [yshift=2.5pt] {worse in 49 of 64};
\end{axis}
\end{scope}
\end{tikzpicture}
\caption{\textbf{On real robot data (DROID), correct memory lowers prediction error and wrong memory raises it.} (a)~Start, goal 7.2\,s later, and wrist view at the goal, from two held-out episodes. (b)~Each dot is one of 64 held-out episodes and shows the percent reduction in the wrist camera's 9-step rollout error (L1) when RoboJEPA reads a memory, relative to the released RoboJEPA without memory (dashed line). Above the line, memory lowers the error. Correct memory holds transitions recorded in the same scene as the test episode, and wrong memory holds transitions from a different scene. Horizontal lines mark the mean, and vertical bars its 95\% CI.}
\label{fig:realdata}
\end{figure*}

%% file: tables/ablations.tex
\begin{table}[t]
\centering
\footnotesize
\setlength{\tabcolsep}{3pt}
\providecommand{\abdelta}[1]{\,{\scriptsize\textcolor{gray}{#1}}}
\resizebox{\ifdim\width>\linewidth\linewidth\else\width\fi}{!}{%
\begin{tabular}{@{}lcccc@{}}
\toprule
 & In range & Out of range & Unseen dyn. & Combinations \\
\midrule
\textbf{Ours} & \textbf{86.3} & \textbf{72.3} & \textbf{84.5} & \textbf{53.2} \\
\midrule
\multicolumn{5}{@{}l}{\textit{Content}} \\
\hspace*{0.9em}Empty memory (reads nothing) & 63.6\abdelta{$-$22.7} & 56.4\abdelta{$-$15.8} & 75.5\abdelta{$-$9.0} & 40.0\abdelta{$-$13.2} \\
\hspace*{0.9em}Mismatched memory (another world's) & 58.1\abdelta{$-$28.2} & 60.9\abdelta{$-$11.3} & 79.7\abdelta{$-$4.8} & 19.8\abdelta{$-$33.5} \\
\addlinespace[3pt]
\multicolumn{5}{@{}l}{\textit{Span}} \\
\hspace*{0.9em}Memory reset every episode & 67.9\abdelta{$-$18.4} & 59.8\abdelta{$-$12.5} & 75.2\abdelta{$-$9.3} & 42.0\abdelta{$-$11.2} \\
\addlinespace[3pt]
\multicolumn{5}{@{}l}{\textit{Alternative adaptation methods, same backbone}} \\
\hspace*{0.9em}History encoder, current episode only & 69.3\abdelta{$-$17.0} & 58.8\abdelta{$-$13.5} & 75.0\abdelta{$-$9.5} & 39.1\abdelta{$-$14.2} \\
\hspace*{0.9em}History encoder, given all earlier episodes & 74.8\abdelta{$-$11.5} & 58.2\abdelta{$-$14.1} & 80.7\abdelta{$-$3.8} & 43.3\abdelta{$-$9.9} \\
\hspace*{0.9em}Context collected before the task & 85.0\abdelta{$-$1.2} & 72.2\abdelta{$-$0.1} & 84.8\abdelta{$+$0.3} & 51.8\abdelta{$-$1.4} \\
\hspace*{0.9em}Test-time training, reset every episode & 55.0\abdelta{$-$31.3} & 48.8\abdelta{$-$23.5} & 65.0\abdelta{$-$19.5} & 30.0\abdelta{$-$23.2} \\
\addlinespace[3pt]
\multicolumn{5}{@{}l}{\textit{Training}} \\
\hspace*{0.9em}Trained without teacher & 82.6\abdelta{$-$3.7} & 65.7\abdelta{$-$6.5} & 80.3\abdelta{$-$4.2} & 47.3\abdelta{$-$5.9} \\
\addlinespace[3pt]
Fine-tuned backbone & 56.3\abdelta{$-$30.0} & 50.3\abdelta{$-$21.9} & 66.5\abdelta{$-$18.0} & 32.0\abdelta{$-$21.2} \\
\bottomrule
\end{tabular}%
}
\caption{\textbf{Success falls when the memory is empty, from another world or reset every episode.} PushT success (\%) with LeWM. The grey numbers give the change from ours in points. Mismatched memory comes from another world of the same group. The fine-tuned backbone is the same model trained without memory. The history encoder follows context inference~\cite{lee2020cadm,kumar2021rma}, test-time training is AdaJEPA~\cite{wang2026adajepa}, and the context collected before the task comes from a separate phase (\cref{sec:supp_efficiency}). Differences outside that row are significant (Holm-corrected McNemar, $p < 0.02$).}
\label{tab:ablations}
\end{table}

%% file: sections/5_conclusion.tex
\section{Conclusion}
\label{sec:conclusion}
With a memory of its own transitions, \lememoire outperforms test-time training out of range and under combined shifts, in three environments and on both backbones.
One episode of experience yields most of the gain.

\paragraph{Limitations and future work.}
\lememoire assumes that the dynamics of a world stay fixed while its memory grows. When the world switches every 25 episodes, a memory of its last 50 transitions re-adapts within about ten episodes and still outperforms the fine-tuned backbone and AdaJEPA (\cref{sec:supp_changing_world}).
Our next directions are larger models, real robots that plan under changed dynamics, and adapting the encoder with the memory to changes in visual appearance.\parfillskip=0pt\par\parfillskip=0pt plus 1fil

%% file: sections/X_suppl.tex
\clearpage
\appendix
\crefalias{section}{appendix}\crefalias{subsection}{subappendix}\crefalias{subsubsection}{subsubappendix}

\addtocontents{toc}{\protect\setcounter{tocdepth}{2}}
\onecolumn
\begin{center}
  {\large LeM\'emoire: Memory-Conditioned JEPA World Models for\\
   Planning in Changing Worlds}\\[0.8em]
  {\Large\bfseries Supplementary Material}\\[1.4em]
\end{center}

\begin{center}\textsc{\large Table of Contents}\end{center}
\noindent\rule{\linewidth}{0.4pt}\par
\setlength{\cftbeforesecskip}{13pt}
\setlength{\cftbeforesubsecskip}{5pt}
\begingroup\hypersetup{linkcolor=black}
\makeatletter
\let\lmTocNumberline\numberline
\def\lmTocEat{\@ifnextchar~\@gobble\@gobbletwo}%
\def\numberline#1\hskip -1em.{\lmTocNumberline{#1.}\lmTocEat}%
\setlength{\cftsecnumwidth}{1.9em}\setlength{\cftsubsecnumwidth}{2.6em}%
\@starttoc{toc}\makeatother
\endgroup
\noindent\rule{\linewidth}{0.4pt}
\twocolumn

\section{Implementation details}
\label{sec:supp_implementation}
\input{sections/supp/implementation}

\subsection{The teacher code}
\label{sec:supp_teacher}
The \tcode $t(\xi)$ is an optional training target: where the training worlds expose their parameters $\xi$, the \dcode is also trained to match $t(\xi)$. It is never computed at test time, where the agent sees only its memory. The models that use it are marked in \cref{tab:supp_teacher}.
\begin{table}[t]
\centering
\footnotesize
\setlength{\tabcolsep}{3.5pt}
\resizebox{\linewidth}{!}{%
\begin{tabular}{@{}llccc@{}}
\toprule
Backbone & Env & $K$ (train / test) & Code enters $P$ through & Teacher \\
\midrule
LeWM & PushT & 0 / 0 & action embedding & yes \\
LeWM & Reacher & 1--8 / 8 & action embedding & no \\
LeWM & Cube & 1--8 / 8 & action embedding & yes \\
TS & PushT & 1--8 / 8 & gated cross-attention & no \\
TS & Reacher & 1--8 / 8 & gated cross-attention & no \\
TS & Cube & 1--8 / 8 & action embedding & yes$^\ast$ \\
\bottomrule
\end{tabular}%
}
\caption{\textbf{Configuration of each \lememoire model in \cref{tab:main_results}.} $K$: refinement steps of the reader. No model uses the \tcode at test time. $^\ast$Starts from a model trained only with the teacher code.}
\label{tab:supp_teacher}
\end{table}
\input{sections/supp/teacher}

\section{Test conditions and world design}
\label{sec:supp_conditions}
Each environment is tested on a fixed set of conditions, grouped into the rows of \cref{tab:main_results}.
A condition changes one or two dynamics parameters of the agent (commands, actuators, joints), the object (the PushT block's friction, inertia and centre of mass) or the scene (table tilt). A few also change the image.
\emph{In range} conditions keep every parameter inside the range our training conditions cover, and include the default.
\emph{Out of range} conditions push a parameter that training varies beyond that range.
\emph{Unseen dynamics} conditions change the dynamics in a way training does not cover.
\emph{Combinations} apply two changes at once.
\Cref{tab:supp_conditions_pusht,tab:supp_conditions_reacher,tab:supp_conditions_cube} list every condition with its value and the training range.
Every condition runs the same 150 test episodes (start and goal pairs).

\input{tables/supp_conditions_pusht}
\input{tables/supp_conditions_reacher}
\input{tables/supp_conditions_cube}

\subsection{\texorpdfstring{Distinctness of test conditions}{Distinctness of test conditions}}
\label{sec:supp_distinctness}
\input{sections/supp/distinctness}

\section{Evaluation protocol}
\label{sec:supp_protocol}
\paragraph{Episode length.} Episodes run for 50 steps.
\paragraph{Statistics.}
We measure uncertainty over test episodes.
Every condition has 150 episodes, and all methods plan the same start and goal pairs from the same start states with planner seed 0. We compare methods with paired exact McNemar tests on the per-episode outcomes, Holm-corrected over the comparisons of each table. We also compare ours with the released backbone, the fine-tuned backbone and AdaJEPA over conditions, for each environment and backbone of \cref{tab:main_results}, with a two-sided exact sign test on the per-condition success rates (ties dropped, Holm-corrected over the 18 comparisons). On PushT with LeWM, ours is higher in 31 of the 32 conditions against the released LeWM and in all 32 against the other two ($p \le 2.3\times10^{-7}$). In all 18 comparisons, ours is higher in more conditions than it is lower, and 12 of the 18 are significant ($p < 0.05$). The $\pm$ in \cref{tab:main_results} is the standard deviation of the success rate over three blocks of 50 episodes.
We keep the checkpoint and settings that plan best on a fixed set of 50 validation episodes, with the same rule for every method, and evaluate them on 150 test episodes. On Cube, the fine-tuned backbone at steps 12k and 60k differs by 0.3 points over all test conditions, and seeds 2 and 3 of the teacher-free Cube model reuse seed 1's 32k checkpoint and stay within 0.6 points of it on the validation episodes.
Our PushT model on the LeWM backbone was also trained with two more seeds, each with its own fine-tuned backbone and with checkpoints chosen by the same rule, and evaluated on the episodes and groups of \cref{tab:main_results}.
On every seed and in every group, ours beats the same model with an empty memory, the fine-tuned backbone and the released LeWM (paired exact McNemar, $p \le 0.0016$). Over the three seeds, our success is $86.5 \pm 0.2$, $72.5 \pm 0.6$, $85.1 \pm 0.6$ and $53.3 \pm 0.7$ in the four groups (mean $\pm$ standard deviation), 20.2 to 29.9 points above the fine-tuned backbone.
On Cube, a looped-reader LeWM model trained without the teacher code was trained with three seeds and evaluated on all 28 Cube conditions of \cref{tab:main_results} (\cref{tab:supp_variants}). Over the three seeds, its success is $86.1 \pm 0.8$, $84.0 \pm 0.8$, $65.4 \pm 0.6$ and $78.9 \pm 0.8$ in the four groups (mean $\pm$ standard deviation). On every seed, it exceeds the released LeWM in every group (by 8.6 to 21.7 points, $p < 10^{-11}$), exceeds the fine-tuned backbone and its own empty memory out of range and under combinations (by 8.8 to 13.4 points, $p < 10^{-10}$), and stays within 1.7 points of the fine-tuned backbone in range and on unseen dynamics, with no significant difference on any seed.
\paragraph{Planning setup.}
Models on the LeWM backbone plan with the cross-entropy method (300 samples, 30 iterations, horizon 5) and execute all 5 planned actions before planning again.
Models on the TS backbone plan with the gradient-based model-predictive control of TS, execute one action per plan and receive the goal with proprioception.
The memory of one condition persists across its 150 episodes and holds transitions only, never goals or success.
Evaluation runs in deterministic mode, and all cells of one model run on one GPU type.

\paragraph{Release commitment.} {\color{black}We will release our code and checkpoints, the evaluation code, the \bench test suite (every condition with its start and goal seeds) and the training data, so that other methods can be compared on the same worlds.}

\section{Baselines}
\label{sec:supp_baselines}
\paragraph{Baseline settings.}
AdaJEPA runs on the fine-tuned backbone with its published test-time settings, taking one Adam step after every plan on the last predictor block (learning rate $5\times10^{-4}$) and on the projection head of the encoder ($10^{-5}$), and resetting after each episode. In its own setting, its released code raises success on the T-shape task by 30.0 points, from 50.7\% to 80.7\% (three test seeds, 150 episodes), where its paper reports 36.7 points, from 50.0\% to 86.7\%.
PhysiCalWorld uses our training recipe and checkpoint selection with its own conditioning module and loss, and no dynamics parameters.
The released LeWM reaches 87.3\% on the default PushT world in our harness, and 94.0\% on the episodes of its own evaluation, where its released implementation gives 96.0\%. On PushT, the In range row of \cref{tab:main_results} averages seven conditions, the default world among them.
\subsection{\texorpdfstring{Fine-tuned backbone}{Fine-tuned backbone}}
\label{sec:supp_control}
AdaJEPA adapts the fine-tuned backbone at test time, and \cref{tab:supp_control} gives its success in every cell of \cref{tab:main_results}.
With the same settings on the released LeWM, AdaJEPA reaches 61.6\%, 51.5\%, 73.7\% and 28.3\% on PushT in range, out of range, under unseen dynamics and under combinations, 2.8 to 4.4 points below the released model without updates and 10.8 to 24.9 points below ours.
\lememoire outperforms the stronger of the released and fine-tuned backbones in 21 of the 24 cells of \cref{tab:main_results}, equals it in one (PushT with TS, combinations) and is lower in two (Cube with LeWM, in range and unseen dynamics). Against the released backbone alone, ours is higher in 23 of the 24 cells and significantly higher in 19 of them, including all 12 on LeWM (paired exact McNemar, Holm-corrected over the 83 comparisons of \cref{tab:main_results}). Against the fine-tuned backbone alone, ours is higher in 22 of the 24 cells. The fine-tuned backbone falls below the released one in 14 of the 24 cells, consistent with \cref{eq:mean}. One predictor fit to all training worlds averages their outcomes, and the same actions end differently in each world (\cref{fig:shift_strip}a). On PushT with LeWM, the fine-tuned backbone uses the same worlds, training budget, settings and checkpoint-selection rule as ours (\cref{sec:supp_protocol}). With an empty memory, our model plans with its learned empty code $\hat e_\varnothing$, and the stored transitions add 9.0 to 22.7 points on top of it on PushT with LeWM (\cref{tab:ablations}).
\input{tables/supp_control}
\input{sections/supp/baseline_arms}

\subsection{\texorpdfstring{AdaJEPA protocols and variants}{AdaJEPA protocols and variants}}
\label{sec:supp_adajepa}
\input{sections/supp/adajepa_protocols}

\paragraph{Updates without reset.} Our memory keeps the experience of earlier episodes as stored transitions. To test whether weights can keep it instead, we carry AdaJEPA's adapted weights and batch-norm statistics across the episodes of each condition. AdaJEPA itself resets them before every episode, so this carrying variant is ours, built as a probe of persistence, not as our test-time-training baseline. The carried variant keeps AdaJEPA's update rule and does not replay earlier episodes. That baseline is AdaJEPA with its published test-time settings, also run at four other learning rates (\cref{tab:supp_adajepa_lr}). Carrying the weights lowers success in every PushT group, by 21 to 59 points, and in three of the five Reacher groups (\cref{tab:supp_adajepa_carry}). This diagnostic uses the released LeWM and AdaJEPA's default settings.
\begin{table}[h]
\centering
\footnotesize
\setlength{\tabcolsep}{3pt}
\resizebox{\ifdim\width>\linewidth\linewidth\else\width\fi}{!}{%
\begin{tabular}{@{}lcc@{}}
\toprule
Test group & PushT, $\Delta$ ($p$) & Reacher, $\Delta$ ($p$) \\
\midrule
In range & $-58.5$ ($3{\times}10^{-30}$) & $-10.8$ (0.004) \\
Out of range & $-35.4$ ($2{\times}10^{-53}$) & $-3.7$ (0.053) \\
Unseen dynamics & $-49.0$ ($9{\times}10^{-36}$) & $-1.3$ (0.86) \\
Unseen visual & $-40.4$ ($9{\times}10^{-24}$) & $-13.3$ (0.012) \\
Combinations & $-20.8$ ($2{\times}10^{-16}$) & $-20.5$ ($5{\times}10^{-8}$) \\
\bottomrule
\end{tabular}}
\caption{\textbf{AdaJEPA gets worse when its updates persist.} $\Delta$ is the paired change in success (points) when AdaJEPA's adapted weights are carried across the episodes of a condition instead of reset before each episode, with its exact McNemar $p$ (unadjusted). Released LeWM, 50 episodes per condition. These rows use a separate set of test conditions, so they are not comparable with \cref{tab:main_results}.}
\label{tab:supp_adajepa_carry}
\end{table}

\paragraph{Updates with replay.}
A second persistent variant trains on all stored experience. It starts from the fine-tuned backbone and keeps its weights and Adam state across the episodes of a condition. Both are reset between conditions. Before each plan once the condition holds a stored transition, it updates the last predictor block on a batch of 32 three-step windows drawn from all transitions stored so far in the condition. It uses AdaJEPA's one-step prediction loss, and the encoder stays frozen. Its learning rate and number of Adam steps per update are chosen over an eight-setting grid by the same rule as every method (\cref{sec:supp_protocol}), selecting a learning rate of $10^{-5}$ and one step. On the 32 PushT test conditions of \cref{tab:main_results} with LeWM and 150 episodes per condition, this variant reaches 57.4\%, 48.8\%, 64.3\% and 34.2\% in the four groups. Ours is above it by 19.1 to 28.9 points in every group (paired exact McNemar, Holm-corrected, $p \le 10^{-20}$). Four Adam steps per update score lower than one step at every learning rate, matching the learning-rate sweep in \cref{tab:supp_adajepa_lr}.

\paragraph{Learning-rate sweep.}
We also run AdaJEPA at four other learning rates of its test-time update, from $10^{-4}$ to $5\times10^{-3}$, on PushT with LeWM and the 150 episodes per condition (\cref{tab:supp_adajepa_lr}).
\begin{table}[h]
\centering
\footnotesize
\setlength{\tabcolsep}{3pt}
\resizebox{\linewidth}{!}{%
\begin{tabular}{@{}lcccccc@{}}
\toprule
 & \multicolumn{5}{c}{AdaJEPA, learning rate} & \\
\cmidrule{2-6}
Test group & $10^{-4}$ & $5{\times}10^{-4\,\dagger}$ & $10^{-3}$ & $2{\times}10^{-3}$ & $5{\times}10^{-3}$ & \lememoire \\
\midrule
In range & 56.2\pmstd{1.9} & 55.0\pmstd{1.9} & 53.4\pmstd{1.7} & 52.3\pmstd{1.8} & 50.2\pmstd{2.2} & \textbf{86.3}\pmstd{1.3} \\
Out of range & 49.4\pmstd{1.4} & 48.8\pmstd{0.4} & 46.8\pmstd{1.6} & 46.4\pmstd{1.0} & 45.1\pmstd{2.0} & \textbf{72.3}\pmstd{3.2} \\
Unseen dynamics & 65.5\pmstd{1.8} & 65.0\pmstd{2.0} & 64.5\pmstd{2.3} & 63.7\pmstd{2.5} & 64.2\pmstd{2.1} & \textbf{84.5}\pmstd{5.7} \\
Combinations & 32.2\pmstd{1.5} & 30.0\pmstd{1.0} & 29.1\pmstd{1.4} & 28.3\pmstd{0.9} & 28.3\pmstd{1.5} & \textbf{53.2}\pmstd{1.8} \\
\bottomrule
\end{tabular}}
\caption{\textbf{No learning rate closes the gap to \lememoire.} Success (\%) of AdaJEPA on PushT (LeWM) at five learning rates of its test-time update, the published one marked $^\dagger$. \lememoire is better in every cell, by 19 to 36 points (McNemar, Holm over 20 comparisons, largest $p = 6\times10^{-21}$). Mean $\pm$ std over three 50-episode blocks, 150 episodes per condition.}
\label{tab:supp_adajepa_lr}
\end{table}

\subsection{\texorpdfstring{Memories and contexts in related methods}{Memories and contexts in related methods}}
\label{sec:supp_memories}
Several recent and concurrent works equip a JEPA world model with a memory or adapt it at test time.
The memory of MemWM is a set of slots learned in training and is not written at test time.
EPM-JEPA conditions an action-free predictor, which does not plan, and PhysiCalWorld reads a clip of exploratory actions collected in the test scene before the task.
AdaJEPA and JEPA-TTT store recent and replayed transitions, respectively, and use them to update their weights.
OpTTA and Sandwich-Residuals also update parameters within each episode and reset them after it, with an optimistic term in the adaptation loss~\cite{mete2026optta} or with small residual corrections around a frozen predictor~\cite{soni2026sandwich}.
Ported to the same backbones, AdaJEPA reaches lower success than \lememoire in all 12 out-of-range and combination groups of the three environments and two backbones, significantly in 11. On the LeWM backbone, PhysiCalWorld reaches lower success than \lememoire in 11 of the 12 groups of the three environments, significantly in 8, and significantly higher in none (\cref{tab:main_results}, \cref{sec:supp_physicalworld}).

Outside JEPA world models, many methods condition a dynamics model or a policy on a context inferred from transitions.
UP-OSI, RMA and DALI read a window of recent transitions and act with a policy or an actor~\cite{yu2017uposi,kumar2021rma,roeder2025dali}. CaDM and TMCL plan with a dynamics model conditioned on the most recent transitions~\cite{lee2020cadm,seo2020tmcl}.
A history encoder that follows this recipe on our backbone, data, teacher and training budget, with a memory of the current episode only, reaches lower success than \lememoire on PushT by 17.0, 13.5, 9.5 and 14.2 points in the four groups (\cref{tab:ablations}).
In their evaluations, IIDA and CLAW plan with a frozen base model whose context is collected without goals before the agent acts, and the agent's later episodes are not added to it~\cite{evans2022iida,palafox2026claw}. IIDA's context can also grow with further interaction~\cite{evans2022iida}.
SPRII takes its context from another interaction of the same system, paired by a known relation~\cite{dai2026sprii}, and cRSSM receives the context values as an input~\cite{prasanna2024crssm}.
RL$^2$ and PEARL keep earlier episodes for a policy~\cite{duan2016rl2,rakelly2019pearl}, and in-context world models keep them for prediction without planning~\cite{wang2026iclwm}.
Among these methods, only \lememoire plans by conditioning its predictor on the transitions it executes toward its goals, kept across the episodes of a world, without weight updates.

\paragraph{History encoder given earlier episodes.}
{\color{black}The history encoder can fall below \lememoire because its memory starts empty in every episode or because its reader is trained on one episode only. To separate the two, we give the trained history encoder earlier episodes of the same world at test, without retraining, on the PushT conditions and episodes of \cref{tab:ablations}. Before its first plan, its memory holds either the previous episode of the world or every earlier one. This memory holds every earlier episode of the condition, matching our memory. Given every earlier episode, it recovers 5.4 of its 17.0-point gap in range, 5.7 of 9.5 points in unseen dynamics, 4.2 of 14.2 points under combinations and none out of range (\cref{tab:supp_history_earlier}). Both settings are 3.8 to 14.1 points below \lememoire in every group and differ from each other by at most 3.4 points in any group. The gain therefore also requires a reader trained on earlier episodes.}
\begin{table}[h]
\centering
\footnotesize
\setlength{\tabcolsep}{3pt}
\def\tS#1{\begin{tabular}[c]{@{}c@{}}#1\end{tabular}}\def\tL#1{\begin{tabular}[c]{@{}l@{}}#1\end{tabular}}%
\resizebox{\linewidth}{!}{%
\begin{tabular}{@{}lcccc@{}}
\toprule
Memory at test & In range & \tS{Out of\\range} & \tS{Unseen\\dynamics} & Combinations \\
\midrule
\multicolumn{5}{@{}l}{\textit{Success (\%)}} \\
\hspace*{0.9em}\tL{Current episode only\\(history encoder)} & 69.3 & 58.8 & 75.0 & 39.1 \\
\hspace*{0.9em}\tL{Previous episode given\\at the start} & 76.1 & 61.6 & 79.5 & 40.2 \\
\hspace*{0.9em}All earlier episodes given & 74.8 & 58.2 & 80.7 & 43.3 \\
\hspace*{0.9em}\lememoire & 86.3 & 72.3 & 84.5 & 53.2 \\
\addlinespace[3pt]
\multicolumn{5}{@{}l}{\textit{Difference in points ($p$)}} \\
\hspace*{0.9em}Previous $-$ \lememoire & \tS{$-10.2$\\($1{\times}10^{-15}$)} & \tS{$-10.7$\\($2{\times}10^{-25}$)} & \tS{$-5.0$\\(0.0009)} & \tS{$-13.1$\\($4{\times}10^{-20}$)} \\
\hspace*{0.9em}All earlier $-$ \lememoire & \tS{$-11.5$\\($5{\times}10^{-20}$)} & \tS{$-14.1$\\($1{\times}10^{-38}$)} & \tS{$-3.8$\\(0.016)} & \tS{$-9.9$\\($1{\times}10^{-12}$)} \\
\hspace*{0.9em}Previous $-$ current only & \tS{$+6.8$\\($1.2{\times}10^{-5}$)} & \tS{$+2.8$\\(0.018)} & \tS{$+4.5$\\(0.050)} & \tS{$+1.1$\\(0.85)} \\
\hspace*{0.9em}All earlier $-$ current only & \tS{$+5.4$\\(0.0015)} & \tS{$-0.6$\\(0.85)} & \tS{$+5.7$\\(0.018)} & \tS{$+4.2$\\(0.0099)} \\
\bottomrule
\end{tabular}}
\caption{\textbf{Earlier episodes at test do not lift the history encoder to \lememoire.} PushT success (\%) with LeWM, 150 episodes per condition. The history encoder is trained with a memory of the current episode only and is not retrained. At test, its memory also holds the previous episode of the world or all earlier ones. Differences are paired and in points, with exact McNemar $p$ Holm-corrected over the 16 comparisons.}
\label{tab:supp_history_earlier}
\end{table}

\paragraph{Cross-episode persistence under per-action replanning.}
To test whether cross-episode gains hold when the planner replans after every action (up to 10 plans per episode, mean 8.5, median 10), we evaluate our TS Reacher model of \cref{tab:main_results} on all 22 Reacher test conditions (150 episodes per condition, same checkpoint and GPU) with its memory cleared before every episode, so that within each episode the reader accumulates every executed transition step by step. Clearing the memory between episodes yields 86.3\%, 35.5\%, 35.8\% and 38.0\% in the four groups, compared with 87.2\%, 62.3\%, 38.1\% and 65.3\% with persistent memory (paired gains from persistence of +0.9, +26.8, +2.3 and +27.3 points, Holm-corrected exact McNemar $p$ = $0.49$, $3.0\times 10^{-44}$, $0.42$ and $8.6\times 10^{-22}$). Under mild in-range shifts, within-episode accumulation already nears persistent memory (within 0.9 points), whereas under larger shifts (out of range and combinations) cross-episode persistence adds 26.8 and 27.3 points even under per-action replanning.

\subsection{PhysiCalWorld}
\label{sec:supp_physicalworld}
PhysiCalWorld~\cite{luo2026physicalworld} is concurrent work that also conditions a JEPA world model on a clip recorded in the test world. \Cref{tab:supp_physcal} lists how the two differ. We port it onto our backbone, data and planner, keep its protocol, and also run a variant that reads the agent's own frames instead of a separate clip, so that neither method takes extra environment steps. Its clip then holds frames from the agent's previous episode in the same world, and is empty in the first episode.
\begin{table}[t]
\centering
\footnotesize
\setlength{\tabcolsep}{3pt}
\resizebox{\ifdim\width>\linewidth\linewidth\else\width\fi}{!}{%
\begin{tabular}{@{}lll@{}}
\toprule
 & PhysiCalWorld (our port) & Ours \\
\midrule
Conditioned on & a 12-frame clip recorded before each episode & the transitions the agent executes \\
Data-collection policy & needed (a trained policy on PushT) & none \\
Extra environment steps & 50 per episode & 0 \\
Updated & once per episode, then fixed & at every plan \\
Across episodes in a world & recomputed (its paper reuses one token per scene) & kept \\
Parameters in training & never & optional \tcode \\
Shifts tested in its paper & a few factors, within the training range & 82 conditions (\cref{sec:supp_conditions}) \\
\bottomrule
\end{tabular}%
}
\caption{\textbf{PhysiCalWorld and \lememoire at test time.}}
\label{tab:supp_physcal}
\end{table}
Without its clip, PhysiCalWorld is significantly below ours in 10 of 12 groups (\cref{tab:supp_physcal_own}). Removing the clip lowers its success significantly only on Cube, in three of its four groups.
\begin{table}[h]
\centering
\footnotesize
\setlength{\tabcolsep}{4pt}
\begin{tabular}{@{}llccc@{}}
\toprule
 & & \multicolumn{2}{c}{PhysiCalWorld} & \\
\cmidrule(lr){3-4}
Env. & Test group & clip & previous episode & Ours \\
\midrule
PushT & In range & 82.2 & 78.9\textsuperscript{*} & 86.3 \\
 & Out of range & 66.4 & 65.0\textsuperscript{*} & 72.3 \\
 & Unseen dyn. & 80.0 & 79.8\textsuperscript{*} & 84.5 \\
 & Combinations & 37.2 & 36.8\textsuperscript{*} & 53.2 \\
Reacher & In range & 54.4 & 55.7\textsuperscript{*} & 80.5 \\
 & Out of range & 15.0 & 13.6\textsuperscript{*} & 28.9 \\
 & Unseen dyn. & 16.1 & 16.1\textsuperscript{*} & 35.7 \\
 & Combinations & 33.2 & 33.5 & 39.7 \\
Cube & In range & 86.1 & 77.4\textsuperscript{*}\textsuperscript{+} & 86.4 \\
 & Out of range & 80.0 & 68.1\textsuperscript{*}\textsuperscript{+} & 82.6 \\
 & Unseen dyn. & 63.4 & 60.1 & 61.4 \\
 & Combinations & 74.1 & 64.5\textsuperscript{*}\textsuperscript{+} & 80.4 \\\bottomrule
\end{tabular}
\caption{\textbf{PhysiCalWorld without its pre-episode clip.} Success (\%) with the clip recorded before each episode, as in \cref{tab:main_results}, and with frames from the agent's previous episode in the same world instead, so that it takes no extra steps. LeWM backbone, 150 episodes per condition. \textsuperscript{*}\,below ours, \textsuperscript{+}\,below its clip version (Holm-corrected paired McNemar over 24 tests, $p<0.05$).}
\label{tab:supp_physcal_own}
\end{table}

\subsection{JEPA-TTT}
\label{sec:supp_jepattt}
\input{sections/supp/jepattt}

\needspace{5\baselineskip}
\section{Additional results}
\label{sec:supp_results}
\subsection{\texorpdfstring{Analysis of the dynamics code}{Analysis of the dynamics code}}\label{sec:supp_code_detail}
A linear probe fit on the codes of the other worlds decodes damping, pusher gain and command lag from the code of a left-out world ($R^2 = 0.84$, $0.62$ and $0.57$, and $0.40$ over all parameters). It does not decode friction ($R^2 = -0.43$). It still decodes damping and pusher gain when every world sharing the held-out world's parameter value is left out of the probe's fit.
On Reacher, whose model trains without the teacher and where some training worlds add a constant torque at the shoulder, the codes rank eight table tilts, which training never varies, in their true order (Spearman 1.0, \cref{fig:code_map}c). Across 32 worlds, codes inferred from disjoint five-episode memories of the same world are less consistent where planning succeeds less often (Spearman $\rho = -0.63$, 95\% CI $[-0.84, -0.29]$, and partial $\rho = -0.47$ given the behaviour distance from the default world).
\input{figures/fig_code_map}
\input{figures/fig_supp_predcurve}
\input{figures/fig_memory_cap}
\input{figures/fig_shift_size}
\input{sections/supp/variants}
\input{tables/supp_cube_default_versions}
\input{tables/supp_reacher_default_versions}

\subsection{\texorpdfstring{Switching dynamics during deployment}{Switching dynamics during deployment}}
\label{sec:supp_changing_world}
The test protocol of \cref{sec:experiments} keeps the world fixed within a condition.
In this experiment the world changes during deployment. The LeWM model plans 8 sequences of 150 PushT episodes, and every 25 episodes the world switches to another of the 32 test conditions, without clearing the memory.
A memory that reads only its last 50 transitions succeeds in 66.3\% of the episodes, 15.3 points above the fine-tuned backbone, 18.2 above AdaJEPA and 9.6 above the full memory (\cref{tab:supp_changing_world}, paired exact McNemar, Holm-corrected over 10 tests, $p \le 7\times10^{-4}$).
It reaches its plateau about 9 episodes after a switch (95\% CI 4 to 11).
The full memory keeps transitions of the previous world and gains less.
\input{tables/supp_changing_world}

\subsection{\texorpdfstring{Comparison with pre-collected contexts}{Comparison with pre-collected contexts}}
\label{sec:supp_efficiency}
Context encoders such as IIDA~\cite{evans2022iida} and PhysiCalWorld~\cite{luo2026physicalworld} read transitions collected before the agent pursues its goals.
We measure the interaction such a context needs on PushT with the LeWM model and change only where the stored transitions come from (\cref{tab:supp_efficiency}).
The weights, reader and planner stay the same.

\looseness=-1 In each fixed world, the agent explores before any goal, replaying training actions and smooth random actions in episodes of 50 steps.
At every episode, the collected context holds as many transitions as our memory holds at that point.
Collected at that size, it nearly doubles the interaction, adding 94\% to the steps of its own episodes (median over the 32 worlds, 80 to 103\%).
Our memory needs no steps beyond the episodes themselves.
At the 10\,Hz control rate of PushT, the extra steps, a median of 4{,}375 per world (3{,}500 to 7{,}400), would take about 7 minutes in each world (6 to 12).

In the changing world of \cref{sec:supp_changing_world}, our memory and the context each hold at most 50 transitions.
A context collected once at the start of a sequence and then frozen reaches 13.9 points lower success than our memory over all episodes (95\% CI 11.1 to 16.7).

Replacing the attention pooling of our reader with the average of IIDA's simplest context encoder~\cite{evans2022iida} lowers success under combined shifts by 3.1 points (95\% CI 1.2 to 5.0, Holm-corrected $p = 0.017$).

\begin{table*}[t]
\centering
\footnotesize
\setlength{\tabcolsep}{2.5pt}
\resizebox{\linewidth}{!}{%
\begin{tabular}{@{}lccccc@{}}
\toprule
\multicolumn{6}{@{}l}{\textit{Fixed worlds, success (\%)}} \\
Context & In range $\uparrow$ & Out of range $\uparrow$ & Unseen dyn. $\uparrow$ & Combinations $\uparrow$ & Exploration $\downarrow$ \\
\midrule
Ours, written while acting & 86.3 & 72.3 & 84.5 & 53.2 & \textbf{0 min} \\
Collected before the goals & 85.0 & 72.2 & 84.8 & 51.8 & 7.3 min / world \\
\midrule
\multicolumn{6}{@{}l}{\textit{Changing world, at most 50 transitions, success (\%) over all 1{,}200 episodes and difference to ours [95\% CI]}} \\
Ours, last 50 transitions & \multicolumn{4}{c}{\textbf{66.3}} & \textbf{0 min} \\
Collected once, then frozen & \multicolumn{4}{c}{52.4 ($-$13.9 [$-$16.7, $-$11.1])} & 0.5 min / sequence \\
\bottomrule
\end{tabular}}
\caption{\textbf{A context collected before the goals at our memory's size takes about 7 minutes of exploration per world before any episode.} PushT with the LeWM model, 150 episodes per condition. Exploration is in minutes at the 10\,Hz control rate of PushT, and bold marks its lowest entry. In fixed worlds, the collection holds only the transitions the context reads. Its median over the 32 worlds is 4{,}375 steps, and the collection adds a median of 94\% to the steps of the context's own episodes. In the changing world, the context is collected once at the start of each sequence of six worlds, in 300 steps. Differences carry unadjusted 95\% CIs. In the success columns, bold marks a paired difference with $p < 0.05$ in an exact McNemar test with Holm correction within each study, on the higher arm.}
\label{tab:supp_efficiency}
\end{table*}

\subsection{\texorpdfstring{Real-robot data}{Real-robot data}}
\label{sec:supp_droid}

Recorded episodes fix the actions in advance, so they test the prediction that planning relies on, not planning itself. Planning with \lememoire on a real robot is our next step (\cref{sec:conclusion}).
The memory read of RoboJEPA adds an attention hook in every predictor block with its own query and key projections and a zero-initialized output. It reads 16 stored transitions retrieved by end-effector pose (position plus 0.15 times rotation distance) from stored transitions of the same scene.
Read into the wrist-view tokens with every predictor weight unchanged, the memory lowers wrist-view 9-step error by $1.29\%$ while leaving the side view unchanged ($+0.07\%$, not significant).
Fine-tuning the predictor on our training scenes lowers the wrist error by $2.17\%$ without memory and by a further 0.3\% with it (paired, $p = 0.004$).

To our knowledge, no large robot video dataset records one robot under labelled changes of its own dynamics, such as a new payload or joint friction. Collecting such data is part of our future work.
We therefore treat the scenes of DROID as worlds.

\section{\texorpdfstring{Alternative memory architectures}{Alternative memory architectures}}
\label{sec:supp_architectures}
\input{sections/supp/architectures}

\section{{Memory size and test-time cost}}
\label{sec:supp_cost}
\input{sections/supp/cost_protocol}
\input{tables/cost}

\section{Additional qualitative results}
\label{sec:supp_qualitative}
\input{figures/supp_qual}
We replay the executed actions of ours, AdaJEPA and the fine-tuned backbone on one shifted test episode of PushT and of Reacher (\cref{fig:supp_qual}).
\input{figures/supp_visual}

\section{Limitations and future directions}
\label{sec:supp_limitations}
\paragraph{Non-stationary worlds and expressiveness.}
\lememoire assumes fixed dynamics within a memory window. When the world switches every 25 episodes, reading the last 50 transitions adapts within about nine episodes (\cref{sec:supp_changing_world}). Pairing this window with automatic change detection from prediction surprise, or with occasional weight updates on stored transitions for shifts outside the training space, is a natural extension.

\paragraph{Visual shifts and encoder adaptation.}
\lememoire targets hidden dynamics with the backbone's frozen encoder. Image corruptions (\cref{fig:supp_visual}) are left to the visual encoder, and adapting the encoder alongside the dynamics memory is future work.

\paragraph{Real-robot planning and scale.}
Training uses trajectories grouped by world, with optional dynamics parameters (omitted in three of six simulation models and on DROID). Extending closed-loop planning to physical robots under controlled payload or friction shifts, and to longer-horizon tasks, is our next step.

\needspace{5\baselineskip}
\section{Ethical considerations and broader impact}
\label{sec:supp_ethics}
\input{sections/supp/ethics}

\needspace{5\baselineskip}
\section{LLM usage statement}
\paragraph{Position.}
The authors regard large language models (LLMs) as a valuable means of accelerating research. But they also share the community's concern about ``AI Slop'' papers produced without meaningful human involvement, which consume reviewing effort without contributing verified work. The authors are committed to the transparent use of such tools. This statement therefore specifies how LLMs were and were not used in this work.

\begin{itemize}
\item \textbf{Literature search.} LLM-assisted search was used to identify related work beyond the papers known to the authors. Its output served only as a list of candidates: each proposed paper was read by an author before being cited, and every reference was verified against its original source.
\item \textbf{Writing.} The authors outlined each section and set every argument and claim. The tools turned these outlines into prose, edited it, checked it against the logged results and critiqued drafts, and the authors revised every sentence.
\item \textbf{Implementation and experiments.} The authors wrote the fundamental training and inference code. Coding agents implemented much of the remaining code from the authors' specifications and managed jobs on our clusters. An author audited all code before use.
\item \textbf{Discussion.} LLMs offered consultation and suggested alternatives, as a collaborator would. Adopting any suggestion remained the authors' decision. The authors observed that, even with frontier models, many suggestions were unfounded or leaned towards ideas already present in the literature, plausibly reflecting the models' training data.
\end{itemize}

\paragraph{Uses of LLMs.}
The authors used Claude (Anthropic), ChatGPT and Codex (OpenAI).

\paragraph{Exclusions.}
\begin{itemize}
\item \textbf{Research decisions.} The research question, method, experimental design and conclusions were determined by the authors. No decision of any kind was left to an LLM.
\item \textbf{Results.} No result was generated, modified or estimated by an LLM. All reported numbers derive from logged experimental runs.
\item \textbf{References.} No reference was included without verification by an author.
\item \textbf{Interpretation.} LLMs were not used to adapt evidence to a predetermined narrative. The conclusions follow from the measured results.
\end{itemize}

%% file: sections/supp/implementation.tex
\input{figures/supp_arch}
Across both backbones and all three environments, \lememoire uses the same memory-row format, permutation-invariant set reader and forward-only test-time loop (\cref{fig:supp_arch}, \cref{tab:supp_recipe}).
The code enters each predictor through an entry point set per backbone and environment, and on PushT and Reacher with TS both entry points beat the fine-tuned backbone (\cref{tab:supp_arch_variants}). Reader depth, entry point and optional teacher are chosen with the same validation rule as the checkpoints.

\paragraph{Memory rows.}
A row holds the history $h_i$ (three latent states and three actions), the next latent state and the surprise, $990$ numbers for LeWM on PushT and Reacher, $1035$ on Cube and $2000$ for TS.
Each action spans 5 simulator steps, and the memory gains one row per executed action.
The surprise is a vector, computed by the backbone as it was before memory training.
Rows carry no position or time index, so the reader treats them as a set, and at test time the reader sees every stored row. The current episode's own transitions enter the memory as soon as the episode has two frames.

\paragraph{Reader.}
A shared MLP embeds every row to 256 dimensions, and a learned score per row weights the rows into one pooled vector.
The looped reader then applies one weight-tied block $K$ times. The block cross-attends from the pooled vector to the rows (width 256, 4 heads) and passes the result through a feed-forward layer (width 512), with zero-initialized output projections.
$K$ is drawn uniformly from 1 to 8 for every training batch and set to 8 at test.
Our LeWM model on PushT uses the pooled vector directly ($K = 0$).
A linear map compresses the result to the 16-dimensional code. An MLP with a zero-initialized output layer maps it to $\hat e$, whose width matches the entry point of the backbone.

\paragraph{Training.}
Each sample pairs a short sequence from one training world with a memory drawn from that world in 80\% of samples, from another world in 10\% and empty in 10\%. The TS model on Cube draws 90\% from the same world and 10\% empty.
The number of rows is drawn between one and all the rows of the world, log-uniformly for LeWM on PushT and Cube and uniformly for the other models. The current sequence contributes only the steps before it.
LeWM trains its prediction loss on the next latent state and a five-step rollout. TS keeps its own prediction loss, with its straightening term on PushT and Reacher.
With the teacher, the predictor is also trained with $t(\xi)$ in place of $\hat e$, and the code is trained to match $t(\xi)$. For LeWM, this uses the normalized squared distance $\mathcal{L}_{\mathrm{match}} = \|\hat e - \mathrm{sg}(t)\|^2 / (\mathrm{sg}(\|t\|^2) + \epsilon)$ with weight $\lambda = 0.286$, where $\mathrm{sg}$ stops the gradient into the teacher, and for TS on Cube by the mean squared error with weight 1.
The image encoder stays frozen in every model.
The fine-tuned backbone follows the recipe of its model without the reader, the teacher and the code.

\begin{table}[t]
\centering
\footnotesize
\setlength{\tabcolsep}{2.2pt}
\resizebox{\linewidth}{!}{%
\begin{tabular}{@{}l*{6}{w{c}{1.1cm}}@{}}%
\toprule
& \multicolumn{3}{c}{\textit{LeWM backbone}} & \multicolumn{3}{c}{\textit{TS backbone}} \\
\cmidrule(lr){2-4} \cmidrule(l){5-7}
& PushT & Reacher & Cube & PushT & Reacher & Cube \\
\midrule
Width of $\hat e$      & 192 & 192 & 192 & 404 & 404 & 10 \\
Memory draws (\%)      & 80/10/10 & 80/10/10 & 80/10/10 & 80/10/10 & 80/10/10 & 90/0/10 \\
Training worlds        & 5{,}200 & 17{,}489 & 12{,}499 & 5{,}200 & 12{,}500 & 10{,}000 \\
Checkpoint (steps)     & 50k & 52k & 10k & 58k & 60k & 15k \\
Learning rate of $P$   & $5{\times}10^{-5}$ & $5{\times}10^{-5}$ & $5{\times}10^{-5}$ & $5{\times}10^{-4}$ & $5{\times}10^{-4}$ & $1.5{\times}10^{-4}$ \\
\midrule
Optimizer              & \multicolumn{6}{c}{\parbox{7.6cm}{\centering AdamW, batch 128, bf16. Reader learning rate $3{\times}10^{-4}$ with 500 warm-up steps and cosine decay}} \\
\bottomrule
\end{tabular}}
\caption{\textbf{Training recipe of each model in \cref{tab:main_results}.} Memory draws: from the same world, another world and empty. Reader, entry point and teacher of each model are in \cref{tab:supp_teacher}.}
\label{tab:supp_recipe}
\end{table}

%% file: figures/supp_arch.tex
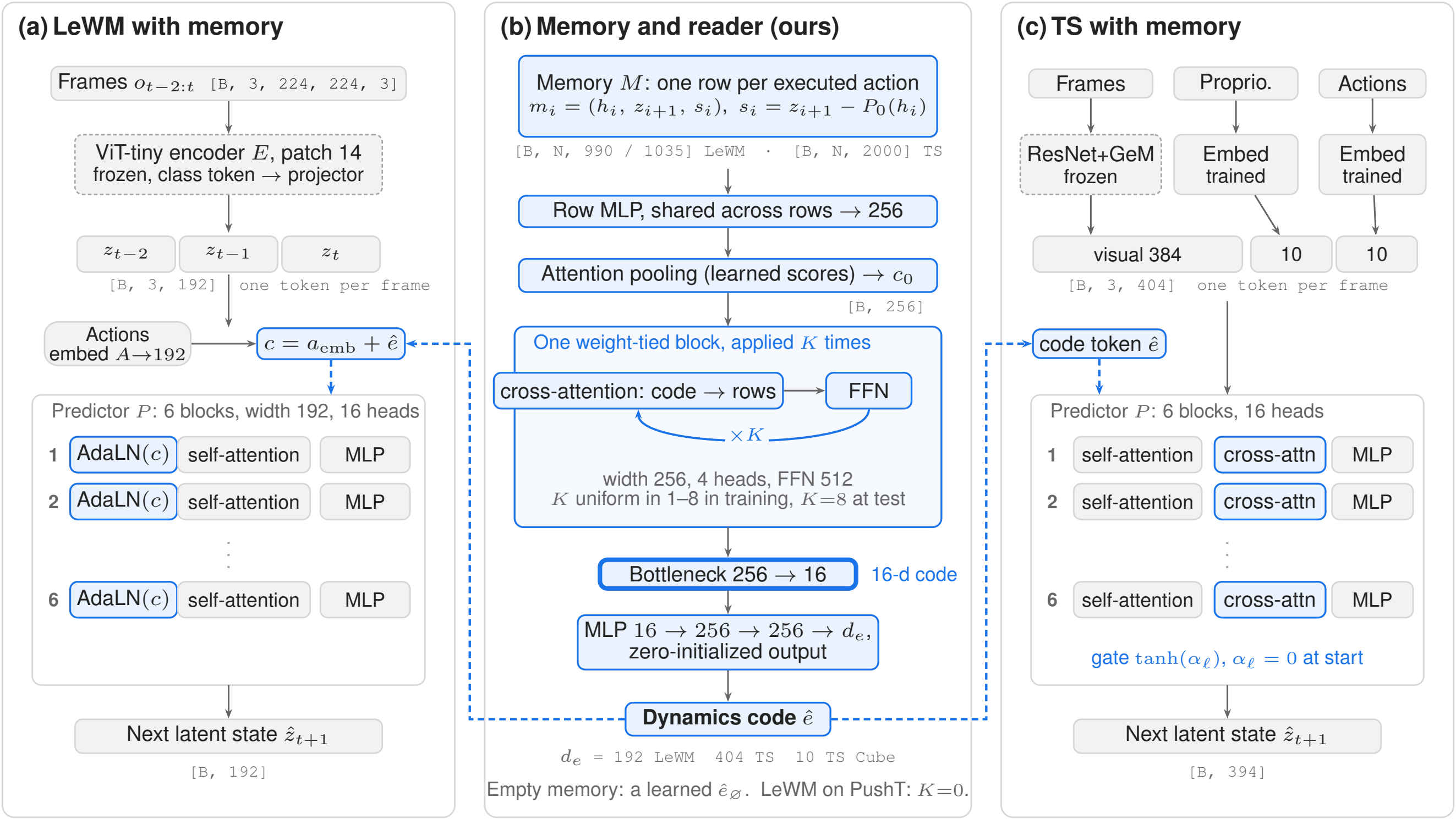
\begin{figure*}[t]
\centering
\input{figures/lm_figstyle}
\def\shp{\fontfamily{pcr}\fontseries{m}\fontsize{5.6}{6.4}\selectfont}%
\tikzset{
  bx/.style={draw=lmcardb, line width=0.5pt, fill=white, rounded corners=2.5pt, font=\lmlab, text=lmink, align=center, inner sep=2.2pt},
  rel/.style={bx, fill=lmpanel},
  frz/.style={bx, fill=lmpanel, draw=lmfaint, dash pattern=on 1.2pt off 0.8pt},
  our/.style={bx, draw=lmours, fill=lmours!9, line width=0.6pt},
  ourk/.style={bx, draw=lmours, fill=lmours!9, line width=1.5pt},
  sub/.style={bx, font=\lmnote, inner sep=1.4pt, minimum height=0.42cm},
  pill/.style={bx, rounded corners=4.5pt, font=\lmnote, inner sep=1.6pt},
  shp/.style={font=\shp, text=lmmuted, inner sep=0.5pt},
  arr/.style={-{Stealth[length=3.2pt]}, lmmuted, line width=0.55pt},
  oarr/.style={-{Stealth[length=3.4pt]}, lmours, line width=0.8pt, dashed, dash pattern=on 2.2pt off 1.4pt},
  tarr/.style={-{Stealth[length=3pt]}, lmfaint, line width=0.55pt, dash pattern=on 1.4pt off 1.2pt}}
\resizebox{\linewidth}{!}{%
\begin{tikzpicture}[x=1cm, y=1cm]
\lmcard{0,0}{5.3,9.55}
\lmhead{0.18,9.25}{a}{LeWM with memory}
\node[rel, minimum width=3.6cm] (la-in) at (2.65,8.6) {Frames $o_{t-2:t}$\ \ {\shp [B, 3, 224, 224, 3]}};
\node[frz, minimum width=3.6cm, minimum height=0.7cm] (la-enc) at (2.65,7.65) {ViT-tiny encoder $E$, patch 14 \\[-1pt] {\lmnote frozen, class token $\to$ projector}};
\draw[arr] (la-in) -- (la-enc);
\node[rel, minimum width=1.15cm, minimum height=0.42cm, font=\lmnote] (la-z2) at (1.45,6.6) {$z_{t-2}$};
\node[rel, minimum width=1.15cm, minimum height=0.42cm, font=\lmnote] (la-z1) at (2.65,6.6) {$z_{t-1}$};
\node[rel, minimum width=1.15cm, minimum height=0.42cm, font=\lmnote] (la-z0) at (3.85,6.6) {$z_{t}$};
\draw[arr] (la-enc) -- (2.65,6.85);
\node[shp, anchor=north east] at (2.55,6.33) {[B, 3, 192]};
\node[shp, anchor=north west] at (2.75,6.33) {one token per frame};
\node[pill, fill=lmpanel] (la-act) at (1.35,5.55) {Actions \\[-1pt] embed $A{\to}192$};
\node[pill, our] (la-c) at (3.85,5.55) {$c = a_{\mathrm{emb}} + \hat e$};
\draw[arr] (la-act) -- (la-c);
\fill[white, draw=lmcardb, line width=0.5pt, rounded corners=3pt] (0.35,1.55) rectangle (4.95,4.95);
\node[anchor=west, font=\lmnote, text=lmmuted] at (0.45,4.75) {Predictor $P$: 6 blocks, width 192, 16 heads};
\foreach \i/\y in {1/4.25, 2/3.7, 6/2.55} {
  \node[font=\lmbadgefont, text=lmmuted] at (0.6,\y) {\i};
  \node[sub, our, minimum width=1.05cm] at (1.42,\y) {AdaLN$(c)$};
  \node[sub, fill=lmpanel, minimum width=1.55cm] at (2.83,\y) {self-attention};
  \node[sub, fill=lmpanel, minimum width=1.05cm] at (4.25,\y) {MLP};}
\node[font=\lmlab, text=lmfaint] at (2.65,3.18) {$\vdots$};
\draw[oarr] (la-c.south) -- (3.85,4.95);
\draw[arr] (la-z1.south) ++(0,-0.02) -- ++(0,-0.62);
\node[rel, minimum width=3.6cm] (la-out) at (2.65,0.95) {Next latent state $\hat z_{t+1}$};
\node[shp, anchor=north] at (2.65,0.62) {[B, 192]};
\draw[arr] (2.65,1.55) -- (la-out);
\lmcard{5.65,0}{11.35,9.55}
\lmhead{5.83,9.25}{b}{Memory and reader (ours)}
\node[our, minimum width=4.9cm, minimum height=0.95cm] (rm) at (8.5,8.45) {Memory $M$: one row per executed action \\[-1pt] {\lmnote $m_i = (h_i,\, z_{i+1},\, s_i)$, \ $s_i = z_{i+1} - P_0(h_i)$}};
\node[shp, anchor=north] at (8.5,7.9) {[B, N, 990 / 1035] LeWM\ \ $\cdot$\ \ [B, N, 2000] TS};
\node[our, minimum width=4.9cm] (rmlp) at (8.5,7.1) {Row MLP, shared across rows $\to$ 256};
\draw[arr] (8.5,7.6) -- (rmlp);
\node[our, minimum width=4.9cm] (rpool) at (8.5,6.35) {Attention pooling (learned scores) $\to c_0$};
\draw[arr] (rmlp) -- (rpool);
\node[shp, anchor=west] at (9.85,5.98) {[B, 256]};
\draw[lmours, line width=0.6pt, rounded corners=3pt, fill=lmours!4] (6.0,3.4) rectangle (11.0,5.75);
\node[anchor=west, font=\lmnote, text=lmours] at (6.1,5.55) {One weight-tied block, applied $K$ times};
\node[sub, our, minimum width=2.5cm] (rx) at (7.45,5.0) {cross-attention: code $\to$ rows};
\node[sub, our, minimum width=1.0cm] (rf) at (10.15,5.0) {FFN};
\draw[arr] (rx) -- (rf);
\draw[lmours, line width=0.7pt, -{Stealth[length=3pt]}] (rf.south) .. controls (10.15,4.3) and (7.45,4.3) .. (rx.south);
\node[font=\lmbadgefont, text=lmours, fill=lmours!4, inner sep=1pt] at (8.72,4.47) {$\times K$};
\node[font=\lmnote, text=lmmuted, align=center] at (8.5,3.83) {width 256, 4 heads, FFN 512 \\[-1pt] $K$ uniform in 1--8 in training, $K{=}8$ at test};
\draw[arr] (rpool) -- (8.5,5.75);
\node[ourk, minimum width=3.0cm] (rb) at (8.5,2.85) {Bottleneck 256 $\to$ 16};
\draw[arr] (8.5,3.4) -- (rb);
\node[font=\lmnote, text=lmours, anchor=west] at (10.05,2.85) {16-d code};
\node[our, align=center] (rctx) at (8.5,2.05) {MLP $16 \to 256 \to 256 \to d_e$,\\[-1pt] zero-initialized output};
\draw[arr] (rb) -- (rctx);
\node[our, minimum width=2.4cm, font=\lmlab\bfseries] (re) at (8.5,1.15) {Dynamics code $\hat e$};
\draw[arr] (rctx) -- (re);
\node[shp, anchor=north] at (8.5,0.82) {$d_e$ = 192 LeWM\ \ 404 TS\ \ 10 TS Cube};
\node[font=\lmnote, text=lmmuted, align=center] at (8.5,0.3) {Empty memory: a learned $\hat e_\varnothing$.\ \ LeWM on PushT: $K{=}0$.};
\lmcard{11.7,0}{17.0,9.55}
\lmhead{11.88,9.25}{c}{TS with memory}
\node[rel, minimum width=1.45cm] (ta-in) at (12.75,8.6) {Frames};
\node[rel, minimum width=1.45cm] (ta-pr) at (14.45,8.6) {Proprio.};
\node[rel, minimum width=1.25cm] (ta-ac) at (16.05,8.6) {Actions};
\node[frz, minimum width=1.45cm, minimum height=0.7cm] (ta-enc) at (12.75,7.65) {ResNet+GeM \\[-1pt] {\lmnote frozen}};
\node[rel, minimum width=1.45cm, minimum height=0.7cm] (ta-penc) at (14.45,7.65) {Embed \\[-1pt] {\lmnote trained}};
\node[rel, minimum width=1.25cm, minimum height=0.7cm] (ta-aenc) at (16.05,7.65) {Embed \\[-1pt] {\lmnote trained}};
\draw[arr] (ta-in) -- (ta-enc); \draw[arr] (ta-pr) -- (ta-penc); \draw[arr] (ta-ac) -- (ta-aenc);
\node[rel, minimum width=2.45cm, minimum height=0.42cm, font=\lmnote] (ta-v) at (13.3,6.6) {visual 384};
\node[rel, minimum width=0.95cm, minimum height=0.42cm, font=\lmnote] (ta-p) at (15.1,6.6) {10};
\node[rel, minimum width=0.95cm, minimum height=0.42cm, font=\lmnote] (ta-a) at (16.1,6.6) {10};
\draw[arr] (ta-enc) -- (12.75,6.82); \draw[arr] (ta-penc) -- (ta-p); \draw[arr] (ta-aenc) -- (ta-a);
\node[shp, anchor=north] at (14.35,6.33) {[B, 3, 404]\ \ one token per frame};
\node[pill, our] (ta-c) at (12.85,5.55) {code token $\hat e$};
\fill[white, draw=lmcardb, line width=0.5pt, rounded corners=3pt] (12.05,1.55) rectangle (16.65,4.95);
\node[anchor=west, font=\lmnote, text=lmmuted] at (12.15,4.75) {Predictor $P$: 6 blocks, 16 heads};
\foreach \i/\y in {1/4.25, 2/3.7, 6/2.55} {
  \node[font=\lmbadgefont, text=lmmuted] at (12.3,\y) {\i};
  \node[sub, fill=lmpanel, minimum width=1.5cm] at (13.3,\y) {self-attention};
  \node[sub, our, minimum width=1.3cm] at (14.85,\y) {cross-attn};
  \node[sub, fill=lmpanel, minimum width=0.95cm] at (16.05,\y) {MLP};}
\node[font=\lmlab, text=lmfaint] at (14.35,3.18) {$\vdots$};
\draw[oarr] (ta-c.south) -- (12.85,4.95);
\draw[arr] (14.35,6.05) -- (14.35,4.95);
\node[font=\lmnote, text=lmours, align=center] at (14.35,1.85) {gate $\tanh(\alpha_\ell)$, $\alpha_\ell = 0$ at start};
\node[rel, minimum width=3.6cm] (ta-out) at (14.35,0.95) {Next latent state $\hat z_{t+1}$};
\node[shp, anchor=north] at (14.35,0.62) {[B, 394]};
\draw[arr] (14.35,1.55) -- (ta-out);
\draw[oarr] (re.west) -- (5.475,1.15) |- (la-c.east);
\draw[oarr] (re.east) -- (11.525,1.15) |- (ta-c.west);
\end{tikzpicture}}
\caption{\textbf{The memory and its reader added to each backbone.} \textbf{(a)}~In LeWM, the \dcode is added to the action embedding, which modulates every predictor block through AdaLN. \textbf{(b)}~The reader embeds each memory row, pools the rows with learned weights, refines the pooled vector with $K$ passes of one weight-tied cross-attention block, and compresses it to a 16-dimensional code before mapping it to $\hat e$. \textbf{(c)}~In TS, every predictor block cross-attends to a code token through a gate that starts closed. On Cube, the code is instead added to the action embedding. Grey: the backbone before memory training, blue what \lememoire adds.}
\label{fig:supp_arch}
\end{figure*}

%% file: sections/supp/teacher.tex
\paragraph{Teacher inputs.}
For LeWM, $\xi$ is a fixed-length encoding of the parameters of a world, 45 entries on PushT and 36 on Cube, each scaled to about $[-1, 1]$ with 0 for the default world.
The \tcode $t(\xi)$ is an MLP ($\xi \to 256 \to 256 \to 192$, zero-initialized output) whose output enters the predictor where $\hat e$ does.
For TS on Cube, $\xi$ has 11 entries (servo strength and damping, command rotation, three gains, delay, lag and three biases), and the teacher MLP has width 64 and output 10.

\paragraph{Training without the teacher.}
Without it, our LeWM model on PushT loses 3.7 to 6.5 points and keeps 70\% to 88\% of its gain over the fine-tuned backbone (\cref{tab:ablations}).
Three of our six models train without it (\cref{tab:supp_teacher}). Without dynamics parameters in training, our LeWM model on Reacher exceeds PhysiCalWorld in every group by 6.5 to 26.1 points (\cref{tab:main_results}), our PushT model trained without the teacher (\cref{tab:ablations}) is within 0.7 points of PhysiCalWorld in three groups and 10.1 points above it under combinations, and a looped-reader LeWM model on Cube reaches $86.1 \pm 0.8$, $84.0 \pm 0.8$, $65.4 \pm 0.6$ and $78.9 \pm 0.8$ over three training seeds (\cref{tab:supp_variants}), matching or exceeding PhysiCalWorld in every Cube group while keeping its out-of-range and combination gains over the fine-tuned backbone on every seed (11.7 to 13.1 and 9.6 to 11.1 points, $p < 10^{-11}$). On Cube, the teacher-trained model of \cref{tab:main_results} scores higher in range and under combinations.
As with the privileged information of RMA~\cite{kumar2021rma}, $\xi$ is used only in training: at test time, the code comes from the reader on the stored rows alone.

%% file: tables/supp_conditions_pusht.tex
\begin{table}[t]
\centering
\footnotesize
\setlength{\tabcolsep}{4pt}
\renewcommand{\arraystretch}{1.1}
\begin{tabular}{@{}L{0.26\linewidth}L{0.45\linewidth}L{0.23\linewidth}@{}}
\toprule
Condition & Change from the default & Training range \\
\midrule
\multicolumn{3}{@{}l}{\textit{In range (7)}} \\
Default & none & --- \\
Friction & block and pusher friction 0.5 & 0--0.75 \\
Damping & space damping 0.001 & 0--0.01 \\
Friction, damping & friction 0.5, damping 0.001 & as above \\
Command rotation & commands rotated by $30^\circ$ & any angle (only in mixed training worlds) \\
Command gain & commands $\times 1.6$ & $\times 0.6$--$\times 1.6$ per axis (only in mixed training worlds) \\
Uneven command gain & $x \times 1.25$, $y \times 0.8$ & $\times 0.6$--$\times 1.6$ per axis (only in mixed training worlds) \\
\midrule
\multicolumn{3}{@{}l}{\textit{Out of range (13)}} \\
Friction & friction 1.5 / 2.0 & 0--0.75 \\
Friction, damping & friction 2.0, damping 0.001 & as above \\
Damping & space damping 0.1 & 0--0.01 \\
Block inertia & moment of inertia $\times 0.1$ / $\times 0.2$ & $\times 0.3$--$\times 3$ \\
Pusher gain & controller velocity gain $\times 2$ / $\times 2.5$ & $\times 0.6$--$\times 1.6$ \\
Block centre of mass & at $(-25, 30)$ / $(20, 45)$ / $(0, 75)$ px & within 15 px of $(0, 45)$ \\
Command delay & 2 simulator steps & 0--1.5 steps \\
Command lag & first-order lag, time constant 4 steps & 0--2 steps \\
\midrule
\multicolumn{3}{@{}l}{\textit{Unseen dynamics (4)}} \\
Table tilt & block accelerated at 400 px/s$^2$ toward the top / left / bottom & not varied \\
Pusher size & collision radius $\times 2$, drawn at normal size & not varied \\
\midrule
\multicolumn{3}{@{}l}{\textit{Combinations (8)}} \\
Mirror, friction & left--right commands flipped + friction 0.5 & \\
Lag, inertia & lag 4 steps + inertia $\times 0.5$ & \\
Pusher gain, friction & gain $\times 2$ + friction 1.5 & \\
Mirror, pusher gain & up--down commands flipped + gain $\times 1.25$ & \\
Lag, centre of mass & lag 4 steps + centre of mass at $(0, 30)$ px & \\
Uneven gain, blur & $x \times 1.25$, $y \times 0.8$ + Gaussian blur, $\sigma = 2$ px & \\
Pusher gain, noise & gain $\times 2$ + salt-and-pepper noise on 1\% of pixels & \\
Lag, noise & lag 4 steps + salt-and-pepper noise on 1\% of pixels & \\
\bottomrule
\end{tabular}
\caption{\textbf{PushT test conditions.}
Lines with several values separated by slashes list one condition per value (32 conditions in total).
The default has zero friction and zero damping (quasi-static pushing).
Training ranges are those of our training conditions. A mirror (negative determinant) is never sampled in training.}
\label{tab:\currfilebase}
\end{table}

%% file: tables/supp_conditions_reacher.tex
\begin{table}[t]
\centering
\footnotesize
\setlength{\tabcolsep}{4pt}
\renewcommand{\arraystretch}{1.1}
\begin{tabular}{@{}L{0.26\linewidth}L{0.45\linewidth}L{0.23\linewidth}@{}}
\toprule
Condition & Change from the default & Training range \\
\midrule
\multicolumn{3}{@{}l}{\textit{In range (5)}} \\
Default & none & --- \\
Bias, lag & shoulder command bias 0.15 + command lag, time constant 1 step & $\pm 0.35$, 0--6 steps \\
Damping & joint damping $\times 1.5$ & $\times 0.3$--$\times 3.5$ \\
Friction, delay & joint friction 0.005 N\,m + action delay 1 step & 0--0.015 N\,m, 0--4 steps \\
Gear & actuator gear $\times 1.25$ (shoulder), $\times 0.8$ (wrist) & $\times 0.4$--$\times 2$ \\
\midrule
\multicolumn{3}{@{}l}{\textit{Out of range (7)}} \\
Damping & joint damping $\times 0.15$ / $\times 0.2$ & $\times 0.3$--$\times 3.5$ \\
Friction & joint friction 0.03 N\,m & 0--0.015 N\,m \\
Gear & actuator gear $\times 3$ / $\times 4$, both joints & $\times 0.4$--$\times 2$ \\
Command coupling & each joint also receives $1.2\times$ the other joint's command & up to $0.6\times$ \\
Brightness & image brightness $\times 0.4$ & $\times 0.6$--$\times 1.5$ \\
\midrule
\multicolumn{3}{@{}l}{\textit{Unseen dynamics (6)}} \\
Table tilt & in-plane gravity 0.5 / 2.5 / 3.5 m/s$^2$ & not varied \\
Swapped commands & shoulder and wrist commands swapped & own command $\times 0.5$--$\times 1.5$ or negated $\times 0.4$--$\times 0.8$, other's up to $0.6\times$ \\
Flipped command & shoulder command negated / wrist command negated & own command $\times 0.5$--$\times 1.5$ or negated $\times 0.4$--$\times 0.8$ \\
\midrule
\multicolumn{3}{@{}l}{\textit{Combinations (4)}} \\
Friction, gear & friction 0.03 N\,m + gear $\times 3$ & \\
Friction, gear & friction 0.03 N\,m + gear $\times 4$ & \\
Damping, blur & damping $\times 0.15$ + Gaussian blur, $\sigma = 2$ px & \\
Gear, noise & gear $\times 3$ + salt-and-pepper noise on 1\% of pixels & \\
\bottomrule
\end{tabular}
\caption{\textbf{Reacher test conditions.}
Lines with several values separated by slashes list one condition per value (22 conditions in total).
The default has joint damping 0.01, no joint friction, gear 0.05, no command coupling, no delay, lag or bias, and vertical gravity.
Training ranges are those of our training conditions.}
\label{tab:\currfilebase}
\end{table}

%% file: tables/supp_conditions_cube.tex
\begin{table}[t]
\centering
\footnotesize
\setlength{\tabcolsep}{4pt}
\renewcommand{\arraystretch}{1.1}
\begin{tabular}{@{}L{0.26\linewidth}L{0.45\linewidth}L{0.23\linewidth}@{}}
\toprule
Condition & Change from the default & Training range \\
\midrule
\multicolumn{3}{@{}l}{\textit{In range (7)}} \\
Default & none & --- \\
Servo stiffness & $\times 0.6$ & $\times 0.4$--$\times 2.5$ \\
Servo damping & $\times 1.8$ & $\times 0.4$--$\times 2.5$ \\
Command rotation & $30^\circ$ & $\pm 45^\circ$ \\
Command gain & $x \times 1.5$, $y \times 0.7$ & $\times 0.5$--$\times 2$, negated $\times 0.4$--$\times 0.6$ \\
Command lag & time constant 3 steps & 0--6 steps \\
Command bias & $+0.1$ on $x$ & 0--0.15 \\
\midrule
\multicolumn{3}{@{}l}{\textit{Out of range (10)}} \\
Servo stiffness & $\times 3.0$ & $\times 0.4$--$\times 2.5$ \\
Servo damping & $\times 0.33$ & $\times 0.4$--$\times 2.5$ \\
Command rotation & $56.25^\circ$ & $\pm 45^\circ$ \\
Command gain & $\times 2.38$ on $x, y, z$ / $\times 0.32$ on $x, y$ & $\times 0.5$--$\times 2$, negated $\times 0.4$--$\times 0.6$ \\
Mirrored axis & $x$ / $y$ / $z$ command negated (gain $-1$) & $\times 0.5$--$\times 2$, negated $\times 0.4$--$\times 0.6$ \\
Command lag & time constant 8.25 steps & 0--6 steps \\
Command bias & $+0.19$ on $x$ & 0--0.15 \\
\midrule
\multicolumn{3}{@{}l}{\textit{Unseen dynamics (6)}} \\
Table tilt & in-plane gravity 4.3 m/s$^2$ along $x$ & not varied \\
Swapped axes & $x \leftrightarrow z$ / $y \leftrightarrow z$ & not varied \\
Cycled axes & $(x, y, z)$ receive the $(z, x, y)$ / $(y, z, x)$ commands & not varied \\
Shear & $x$ command $+\,0.225 \times$ the $y$ command & not varied \\
\midrule
\multicolumn{3}{@{}l}{\textit{Combinations (5)}} \\
Rotation, mirror & $x \leftrightarrow y$ swap: rotation $90^\circ$ with one axis negated & \\
Damping, rotation & servo damping $\times 0.33$ + rotation $56.25^\circ$ & \\
Gain, rotation & gain $\times 0.32$ on $x, y$ + rotation $56.25^\circ$ & \\
Lag, stiffness & lag 8.25 steps + servo stiffness $\times 3.0$ & \\
Rotation, stiffness & rotation $56.25^\circ$ + servo stiffness $\times 3.0$ & \\
\bottomrule
\end{tabular}
\caption{\textbf{Cube test conditions.}
Lines with several values separated by slashes list one condition per value (28 conditions).
Rotation, gain and bias act on the commanded end-effector displacement $(\Delta x, \Delta y, \Delta z)$. Servo damping is the velocity gain of the arm's position servo.
Each out-of-range value is the first one beyond the training edge that moves the cube's trajectory by at least 2.4 cm.
}
\label{tab:\currfilebase}
\end{table}

%% file: sections/supp/distinctness.tex
\begin{figure*}[t]
\centering
\captionsetup{skip=3pt}
\input{figures/fig_supp_distinct}
\caption{\textbf{Every test world that changes the dynamics behaves differently from the default world.} Each mark is one test condition, placed by how far the end states of 32 replayed action sequences lie from those of the default world (0), measured in the simulator. The bands span the distances of the training worlds.}
\label{fig:supp_distinct}
\end{figure*}
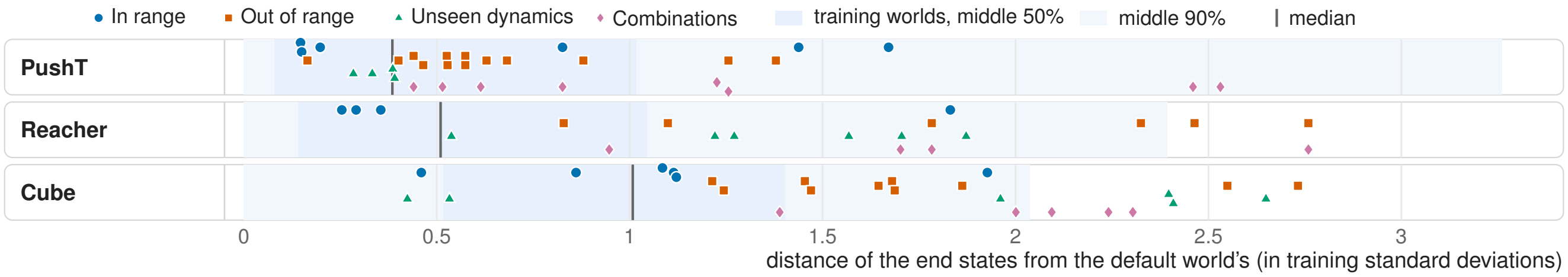

We measure how differently each test world behaves in the simulator itself, with no model.
From 32 start states recorded in the default world, we replay the same 50 simulator steps of recorded actions in every world. We keep the end state: the positions of the pusher and the block and the block's angle on PushT, the fingertip position and both joint angles on Reacher, and the positions of the cube and the gripper on Cube.
The distance between two worlds is the mean, over the 32 replays, of the Euclidean distance between their end states, with each coordinate divided by its standard deviation over the training worlds.
Every test condition that changes the dynamics is separated from the default world: its distance is at least three times the typical gap between neighbouring training worlds (\cref{fig:supp_distinct}).
\cref{fig:shift_strip}b ranks each condition by its distance to the nearest training world.
The two flipped commands on Reacher are partly covered by training, which also reverses command axes. Their behaviour is less novel than that of the other unseen-dynamics conditions.
We exclude factors that barely change the dynamics. On PushT, where pushing is quasi-static, scaling the block's mass or its angular damping leaves the replays unchanged. On Cube, the cube's mass, centre of mass, grip and friction move the final positions of the cube and the gripper by less than 0.024\,m on average (mass by at most 0.014\,m, even at 18.7 times its default).

%% file: figures/fig_supp_distinct.tex
\input{figures/lm_figstyle}
\pgfplotsset{lmdist/.style={lm, anchor=south west, width=17.40cm, height=0.62cm, xmin=-0.62, xmax=3.42, ymin=0.4, ymax=4.6,
  ytick=\empty, xtick=\empty, axis y line=none, axis x line=none, ymajorgrids=false}}
\def\lmdtrack#1#2#3#4#5#6{\fill[lmcard, draw=lmcardb, line width=0.4pt, rounded corners=2.5pt] (axis cs:-0.62,0.4) rectangle (axis cs:3.42,4.6);
  \fill[lmband!55] (axis cs:#2,0.4) rectangle (axis cs:#6,4.6);
  \fill[lmband!100!lmcardb] (axis cs:#3,0.4) rectangle (axis cs:#5,4.6);
  \draw[lmcardb, line width=0.5pt] (axis cs:-0.05,0.4) -- (axis cs:-0.05,4.6);
  \draw[lmgrid, line width=0.5pt] (axis cs:0.5,0.4) -- (axis cs:0.5,4.6) (axis cs:1,0.4) -- (axis cs:1,4.6) (axis cs:1.5,0.4) -- (axis cs:1.5,4.6)
    (axis cs:2,0.4) -- (axis cs:2,4.6) (axis cs:2.5,0.4) -- (axis cs:2.5,4.6) (axis cs:3,0.4) -- (axis cs:3,4.6);
  \draw[lmmuted, line width=0.8pt] (axis cs:#4,0.4) -- (axis cs:#4,4.6);
  \node[font=\lmlab, text=lmink, anchor=west, inner sep=0pt] at (axis cs:-0.58,2.5) {\textbf{#1}};}
\begin{tikzpicture}[x=1cm, y=1cm]
\begin{scope}[every node/.style={font=\lmtick, text=lmink, anchor=west, inner sep=0pt}]
\path plot[mark=*, mark size=1.5pt, mark options={draw=white, line width=0.45pt, fill=gIn}] coordinates {(1.05,2.66)}; \node at (1.18,2.66) {In range};
\path plot[mark=square*, mark size=1.3pt, mark options={draw=white, line width=0.45pt, fill=gOut}] coordinates {(2.50,2.66)}; \node at (2.63,2.66) {Out of range};
\path plot[mark=triangle*, mark size=1.75pt, mark options={draw=white, line width=0.45pt, fill=gUn}] coordinates {(4.40,2.66)}; \node at (4.53,2.66) {Unseen dynamics};
\path plot[mark=diamond*, mark size=1.75pt, mark options={draw=white, line width=0.45pt, fill=gCo}] coordinates {(6.65,2.66)}; \node at (6.78,2.66) {Combinations};
\fill[lmband!100!lmcardb] (8.60,2.58) rectangle (8.90,2.74); \node at (9.03,2.66) {training worlds, middle 50\%};
\fill[lmband!55] (12.00,2.58) rectangle (12.30,2.74); \node at (12.43,2.66) {middle 90\%};
\draw[lmmuted, line width=0.8pt] (14.20,2.56) -- (14.20,2.76); \node at (14.33,2.66) {median};
\end{scope}
\begin{axis}[lmdist, at={(0cm,1.80cm)}]
\lmdtrack{PushT}{0}{0.079}{0.385}{1.017}{3.262}
\addplot[only marks, lmIn] coordinates {(0.147,4.35) (0.150,3.65) (0.198,4) (0.826,4) (1.438,4) (1.671,4)};
\addplot[only marks, lmOut] coordinates {(0.165,3) (0.401,3) (0.440,3.35) (0.465,2.65) (0.526,3.35) (0.528,2.65) (0.574,3.35) (0.574,2.65) (0.629,3) (0.682,3) (0.880,3) (1.256,3) (1.379,3)};
\addplot[only marks, lmUn] coordinates {(0.284,2) (0.333,2) (0.386,2.35) (0.391,1.65)};
\addplot[only marks, lmCo] coordinates {(0.440,1) (0.515,1) (0.614,1) (0.826,1) (1.226,1.35) (1.256,0.65) (2.460,1) (2.531,1)};
\end{axis}
\begin{axis}[lmdist, at={(0cm,1.10cm)}]
\lmdtrack{Reacher}{0}{0.141}{0.510}{1.045}{2.393}
\addplot[only marks, lmIn] coordinates {(0.254,4) (0.291,4) (0.355,4) (1.831,4)};
\addplot[only marks, lmOut] coordinates {(0.829,3) (1.099,3) (1.783,3) (2.325,3) (2.464,3) (2.759,3)};
\addplot[only marks, lmUn] coordinates {(0.538,2) (1.221,2) (1.271,2) (1.568,2) (1.705,2) (1.872,2)};
\addplot[only marks, lmCo] coordinates {(0.947,1) (1.702,1) (1.783,1) (2.759,1)};
\end{axis}
\begin{axis}[lmdist, at={(0cm,0.40cm)}]
\lmdtrack{Cube}{0}{0.517}{1.008}{1.403}{2.037}
\addplot[only marks, lmIn] coordinates {(0.460,4) (0.861,4) (1.085,4.35) (1.114,4) (1.121,3.65) (1.927,4)};
\addplot[only marks, lmOut] coordinates {(1.214,3.35) (1.244,2.65) (1.454,3.35) (1.470,2.65) (1.645,3) (1.680,3.35) (1.687,2.65) (1.862,3) (2.549,3) (2.732,3)};
\addplot[only marks, lmUn] coordinates {(0.424,2) (0.533,2) (1.961,2) (2.397,2.35) (2.409,1.65) (2.649,2)};
\addplot[only marks, lmCo] coordinates {(1.389,1) (2.001,1) (2.094,1) (2.241,1) (2.304,1)};
\begin{scope}[every node/.style={font=\lmtick, text=lmmuted, anchor=north, inner sep=0pt, yshift=-2.5pt}]
\node at (axis cs:0,0.4) {0}; \node at (axis cs:0.5,0.4) {0.5}; \node at (axis cs:1,0.4) {1}; \node at (axis cs:1.5,0.4) {1.5};
\node at (axis cs:2,0.4) {2}; \node at (axis cs:2.5,0.4) {2.5}; \node at (axis cs:3,0.4) {3};
\end{scope}
\end{axis}
\node[font=\lmlab, text=lmink, anchor=north east, inner sep=0pt] at (17.40,0.06) {distance of the end states from the default world's (in training standard deviations)};
\end{tikzpicture}\par

%% file: tables/supp_control.tex
\begin{table}[t]
\centering
\footnotesize
\setlength{\tabcolsep}{2.5pt}
\resizebox{\linewidth}{!}{%
\begin{tabular}{@{}lcccccc@{}}
\toprule
& \multicolumn{3}{c}{\textit{LeWM backbone}} & \multicolumn{3}{c}{\textit{TS backbone}} \\
\cmidrule(lr){2-4} \cmidrule(l){5-7}
Test group & PushT & Reacher & Cube & PushT & Reacher & Cube \\
\midrule
In range        & 56.3\pmstd{3.5} & 49.9\pmstd{3.5} & 87.3\pmstd{3.6} & 62.8\pmstd{6.3} & 70.8\pmstd{0.4} & 84.0\pmstd{2.2} \\
Out of range    & 50.3\pmstd{0.9} & 18.3\pmstd{0.8} & 71.4\pmstd{1.2} & 54.7\pmstd{3.8} & 23.1\pmstd{3.3} & 70.7\pmstd{2.9} \\
Unseen dynamics & 66.5\pmstd{0.5} & 14.1\pmstd{2.8} & 66.4\pmstd{0.4} & 58.7\pmstd{9.3} & 22.8\pmstd{1.4} & 59.0\pmstd{2.9} \\
Combinations    & 32.0\pmstd{0.3} & 34.3\pmstd{4.2} & 68.4\pmstd{2.4} & 44.1\pmstd{2.3} & 25.8\pmstd{4.0} & 62.1\pmstd{4.0} \\
\bottomrule
\end{tabular}}
\caption{\textbf{The fine-tuned backbone in every cell of \cref{tab:main_results}.} Planning success (\%) of the same backbone trained on the same worlds without memory, mean $\pm$ std over three 50-episode blocks, 150 episodes per condition.}
\label{tab:supp_control}
\end{table}

%% file: sections/supp/baseline_arms.tex
We train PLDM and GCBC ourselves, each in two arms (\cref{tab:supp_arms}).
Arm (a) trains the method with its own recipe on the data of LeWM, which come from the default world only.
In the default world, PLDM (a) reaches 82.7\% on PushT and 79.3\% on Reacher, where LeWM reports 78\% for both, and GCBC (a) reaches 87.3\% on PushT, where LeWM reports 75\%.
Arm (b) starts from the final weights of (a) and fine-tunes them on our training worlds, as our fine-tuned backbone does, for 60{,}000 steps at batch size 128.
A checkpoint is saved every 5{,}000 steps, and we keep the one with the highest planning success on the validation episodes, with ties going to the later save. Validation loss never selects.
For PLDM and GCBC in each environment, \cref{tab:main_results} shows the arm with the higher success on the validation episodes, with ties going to (b). The other arm is reported in \cref{tab:supp_pusht_finetuned_versions,tab:supp_reacher_default_versions,tab:supp_cube_default_versions}.
For TS, arm (a) on PushT is the authors' released model. On Reacher and Cube, which the release does not cover, we train TS with its released PushT recipe on the data of LeWM.
\Cref{tab:main_results} shows TS (a) in every environment. On PushT, a TS model fine-tuned on our worlds from (a) for 60{,}000 steps, with its encoder frozen, scores below the release on the validation episodes.
\begin{table}[h]
\centering
\footnotesize
\setlength{\tabcolsep}{2.5pt}
\resizebox{\ifdim\width>\linewidth\linewidth\else\width\fi}{!}{%
\begin{tabular}{@{}llllll@{}}
\toprule
Method & Arm & Starts from & Data & Training & In \cref{tab:main_results} \\
\midrule
PLDM & (a) & scratch & LeWM's & own recipe, 10 epochs & PushT \\
 & (b) & PLDM (a) & ours & 60k steps & Reacher, Cube \\
GCBC & (a) & scratch$^\dagger$ & LeWM's & own recipe, 10 epochs & PushT \\
 & (b) & GCBC (a) & ours & 60k steps & Reacher, Cube \\
TS & (a) & released$^\S$ & & & all three \\
 & (b) & TS (a) & ours & 60k steps$^\ddagger$ & --- \\
\bottomrule
\end{tabular}%
}
\caption{\textbf{Two arms per outside baseline.} LeWM's data come from the default world, and ours are our training worlds. Arm (b) keeps the checkpoint with the highest success on the validation episodes. $^\dagger$Frozen DINOv2 encoder. $^\S$PushT. On Reacher and Cube, its released recipe on LeWM's data (2 epochs). $^\ddagger$PushT, encoder frozen.}
\label{tab:supp_arms}
\end{table}
\begin{table}[h]
\centering
\footnotesize
\setlength{\tabcolsep}{5pt}
\begin{tabular}{@{}lccc@{}}
\toprule
Test group & PLDM (b) & GCBC (b) & Ours \\
\midrule
In range        & 49.8\pmstd{2.7} & 42.4\pmstd{2.4} & \textbf{86.3}\pmstd{1.3} \\
Out of range    & 45.3\pmstd{2.4} & 31.9\pmstd{2.2} & \textbf{72.3}\pmstd{3.2} \\
Unseen dynamics & 58.3\pmstd{1.3} & 34.0\pmstd{2.6} & \textbf{84.5}\pmstd{5.7} \\
Combinations    & 29.4\pmstd{0.6} & 18.2\pmstd{0.9} & \textbf{53.2}\pmstd{1.8} \\
\bottomrule
\end{tabular}
\caption{\textbf{PushT, fine-tuned arms of PLDM and GCBC.} Success (\%) of arm (b), while \cref{tab:main_results} shows arm (a), which scores higher on the validation episodes. LeWM backbone for ours, 150 episodes per condition, $\pm$ over three blocks of 50 episodes.}
\label{tab:supp_pusht_finetuned_versions}
\end{table}

%% file: sections/supp/adajepa_protocols.tex
On the TS backbone, AdaJEPA raises Reacher in-range success by 8.3 points (70.8\% to 79.1\%) and unseen dynamics by 5.4 points (22.8\% to 28.2\%) over the fine-tuned backbone, yet \lememoire remains above it in every group of every environment, by 7.8 to 43.3 points out of range and under combinations (\cref{tab:main_results}).

%% file: sections/supp/jepattt.tex
JEPA-TTT~\cite{zhang2026jepattt} is concurrent work that also adapts a JEPA world model to changed dynamics and keeps what it learns across episodes, by training its predictor and action encoder at test time (\cref{tab:supp_jepattt}).
Its code was not public at the time of writing, so a comparison on our benchmark was not feasible. We cite and discuss it, and set out the differences below.
\begin{table}[h]
\centering
\footnotesize
\setlength{\tabcolsep}{3pt}
\resizebox{\ifdim\width>\linewidth\linewidth\else\width\fi}{!}{%
\begin{tabular}{@{}lll@{}}
\toprule
 & JEPA-TTT & Ours \\
\midrule
Adapts & predictor and action encoder & dynamics code \\
Update at test time & at most one gradient step per plan & stores each executed transition \\
Weights at test time & updated & fixed \\
Kept across episodes & weights, optimizer, replay buffer & memory of transitions \\
Task given by & reward head fitted offline & goal image \\
Adaptation budget & 500 episodes per world & the 150 test episodes \\
Shifts tested & 8 action or contact remappings & \cref{sec:supp_conditions} \\
\bottomrule
\end{tabular}%
}
\caption{\textbf{JEPA-TTT and \lememoire at test time.} JEPA-TTT is scored on held-out episodes after its adaptation episodes. Ours is scored on the episodes whose transitions it stores.}
\label{tab:supp_jepattt}
\end{table}

%% file: figures/fig_code_map.tex
\begin{figure*}[t]
\centering
\captionsetup{skip=3pt}%
\input{figures/lm_figstyle}
\colorlet{lmring}{lmcardb}
\begin{tikzpicture}[x=1cm, y=1cm]
\lmcard{0,0}{5.35,4.78}\lmcard{5.5,0}{12.65,4.78}\lmcard{12.8,0}{17.4,4.78}
\lmhead{0.22,4.50}{a}{Map of PushT codes}
\lmhead{5.68,4.50}{b}{Code vs.\ factor setting}
\lmhead{12.98,4.50}{c}{Code vs.\ tilt (Reacher)}
\begin{axis}[lm, name=a, at={(0.2cm,0.2cm)}, anchor=south west, width=5.01cm, height=4.0cm,
  xmin=-4.25, xmax=8.03, ymin=-4.5, ymax=5.3, hide axis,
  declare function={dX(\x)=\x+0.794616; dY(\y)=\y+1.313165; rR(\x,\y)=sqrt(dX(\x)^2+dY(\y)^2); lR(\x,\y)=ln(10*rR(\x,\y)+sqrt(100*rR(\x,\y)^2+1));},
  legend style={font=\lmtick, text=lmink, draw=none, fill=lmcard, rounded corners=2pt, at={(0.0,0.0)}, anchor=south west, row sep=-1.2pt, inner sep=1.2pt,
    nodes={inner sep=0.9pt}}, legend cell align=left]
\begin{scope}\clip (rel axis cs:0,0) rectangle (rel axis cs:1,1);
\draw[lmring, line width=0.5pt] (axis cs:-4.25,0) -- (axis cs:8.03,0) (axis cs:0,-4.5) -- (axis cs:0,5.3);
\draw[lmring, line width=0.6pt] (axis cs:0,0) circle [radius=0.8814] (axis cs:0,0) circle [radius=2.9982] (axis cs:0,0) circle [radius=5.2983];
\end{scope}
\node[font=\lmnote, text=lmmuted, anchor=north west, inner sep=1pt] at (axis cs:0.88,0) {0.1};
\node[font=\lmnote, text=lmmuted, anchor=north west, inner sep=1pt] at (axis cs:3.0,0) {1};
\node[font=\lmnote, text=lmmuted, anchor=north west, inner sep=1pt] at (axis cs:5.3,0) {10};
\node[font=\lmlab, anchor=east, inner sep=1.2pt, fill=lmcard] at (axis cs:8.03,0) {PC 1};
\node[font=\lmlab, anchor=north west, inner sep=1.2pt] at (axis cs:0.1,5.3) {PC 2};
\addplot[scatter, only marks, scatter src=explicit symbolic, scatter/classes={
    in_range={lmIn}, out_of_range={lmOut}, unseen_dynamics={lmUn}, combinations={lmCo}, unseen_visual={lmLo}}]
  table[x expr={lR(\thisrow{pc1},\thisrow{pc2})*dX(\thisrow{pc1})/rR(\thisrow{pc1},\thisrow{pc2})},
        y expr={lR(\thisrow{pc1},\thisrow{pc2})*dY(\thisrow{pc2})/rR(\thisrow{pc1},\thisrow{pc2})}, meta=group]
  {figures/fig56data/codemap_coords.tsv};
\legend{In range, Out of range, Unseen dyn., Combinations, Looks only}
\addplot[only marks, mark=*, mark size=1.15*\lmms pt, draw=lmink, line width=0.7pt, fill=gIn, forget plot] coordinates {(0,0)};
\begin{scope}[every node/.style={font=\lmmaplab, text=lmink, inner sep=1pt}]
\node[anchor=north east] at (axis cs:-0.45,-0.05) {default};
\node[anchor=west] at (axis cs:0.5,0.45) {tilt};
\node[anchor=east] at (axis cs:-0.47,1.3) {CoM};
\node[anchor=north west, inner sep=1pt] at (axis cs:0.85,1.9) {friction};
\node[anchor=east, inner sep=1pt] at (axis cs:-1.25,1.98) {damping};
\node[anchor=east, inner sep=1.5pt] at (axis cs:-1.6,2.6) {pusher size};
\node[anchor=west, inner sep=1pt] at (axis cs:0.1,3.72) {inertia};
\node[anchor=east, inner sep=1.5pt] at (axis cs:-1.5,3.5) {damping};
\node[anchor=west, text=lmmuted, inner sep=1.8pt, yshift=1.9pt] at (axis cs:-0.55,4.24) {red block};
\node[anchor=east, inner sep=1.5pt] at (axis cs:-1.7,4.46) {gain \texttimes1.6};
\node[anchor=south west] at (axis cs:3.0,2.95) {lag, delay};
\node[anchor=west, inner sep=1.5pt] at (axis cs:5.0,1.45) {rotated 30\textdegree};
\node[anchor=west, inner sep=1.5pt] at (axis cs:3.1,-3.85) {pusher \texttimes2, \texttimes2.5};
\node[anchor=west, inner sep=1.5pt] at (axis cs:5.75,-1.85) {mirrored};
\end{scope}
\end{axis}
\node[font=\lmlab, text=lmink, rotate=90, align=center, inner sep=0pt] at (5.79,2.47) {code position on\\the factor's PC 1};
\node[font=\lmlab, text=lmink] at (9.35,0.12) {factor setting (evenly spaced, shaded: in training)};
\pgfplotsset{lmfac/.style={lm, anchor=south west, width=3.04cm, height=1.49cm, ytick=\empty, axis y line=none, xmajorgrids=false}}
\begin{axis}[lmfac, at={(6.18cm,2.71cm)}, xmin=0.55, xmax=4.45, ymin=-1.56, ymax=2.07, xtick={1,2,3,4}, xticklabels={0,0.5,1.5,2}]
  \fill[lmband] (axis cs:0.55,-1.56) rectangle (axis cs:2.5,2.07);
  \node[anchor=north west, font=\lmlab, text=lmink, inner sep=1pt] at (rel axis cs:0.02,1) {friction};
  \addplot[lmline] coordinates {(1,-1.121) (2,-1.061) (3,0.851) (4,1.331)};
  \addplot[only marks, lmIn] coordinates {(1,-1.121) (2,-1.061)};
  \addplot[only marks, lmOut] coordinates {(3,0.851) (4,1.331)};
\end{axis}
\begin{axis}[lmfac, at={(9.48cm,2.71cm)}, xmin=0.55, xmax=3.45, ymin=-3.68, ymax=5.05, xtick={1,2,3}, xticklabels={0,10\textsuperscript{\textminus3},0.1}]
  \fill[lmband] (axis cs:0.55,-3.68) rectangle (axis cs:2.5,5.05);
  \node[anchor=north west, font=\lmlab, text=lmink, inner sep=1pt] at (rel axis cs:0.02,1) {damping};
  \addplot[lmline] coordinates {(1,-2.621) (2,-0.656) (3,3.276)};
  \addplot[only marks, lmIn] coordinates {(1,-2.621) (2,-0.656)};
  \addplot[only marks, lmOut] coordinates {(3,3.276)};
\end{axis}
\begin{axis}[lmfac, at={(6.18cm,0.74cm)}, xmin=0.55, xmax=3.45, ymin=-3.79, ymax=6.44, xtick={1,2,3}, xticklabels={0.1,0.2,1}]
  \fill[lmband] (axis cs:2.5,-3.79) rectangle (axis cs:3.45,6.44);
  \node[anchor=north west, font=\lmlab, text=lmink, inner sep=1pt] at (rel axis cs:0.02,1) {block inertia (\texttimes)};
  \addplot[lmline] coordinates {(1,-2.544) (2,-1.823) (3,4.367)};
  \addplot[only marks, lmIn] coordinates {(3,4.367)};
  \addplot[only marks, lmOut] coordinates {(1,-2.544) (2,-1.823)};
\end{axis}
\begin{axis}[lmfac, at={(9.48cm,0.74cm)}, xmin=0.55, xmax=3.45, ymin=-6.65, ymax=5.80, xtick={1,2,3}, xticklabels={1,2,2.5}]
  \fill[lmband] (axis cs:0.55,-6.65) rectangle (axis cs:1.5,5.80);
  \node[anchor=north west, font=\lmlab, text=lmink, inner sep=1pt] at (rel axis cs:0.02,1) {pusher gain (\texttimes)};
  \addplot[lmline] coordinates {(1,-5.133) (2,1.856) (3,3.277)};
  \addplot[only marks, lmIn] coordinates {(1,-5.133)};
  \addplot[only marks, lmOut] coordinates {(2,1.856) (3,3.277)};
\end{axis}
\begin{axis}[lm, name=c, at={(14.02cm,0.74cm)}, anchor=south west, width=3.22cm, height=3.46cm, xmin=-0.25, xmax=3.75, ymin=-8, ymax=140,
  xlabel={tilt (in-plane \textit{g}, m/s\textsuperscript{2})}, ylabel={distance from the default\\world's code ($\sigma$)}, xlabel style={yshift=4.5pt}, ylabel style={yshift=-7pt, align=center},
  xtick={0,1,2,3}, ytick={0,40,80,120}]
\fill[lmband] (axis cs:-0.25,-8) rectangle (axis cs:0.25,140);
\addplot[draw=none, fill=gUn, fill opacity=0.12, forget plot] coordinates {(0,0) (0.5,14.105) (0.75,33.266) (1,48.567) (1.5,72.666) (2,98.366) (2.5,107.373) (3.5,126.256)} \closedcycle;
\addplot[gUn, line width=\lmlw] coordinates {(0,0) (0.5,14.105) (0.75,33.266) (1,48.567) (1.5,72.666) (2,98.366) (2.5,107.373) (3.5,126.256)};
\addplot[only marks, lmIn] coordinates {(0,0)};
\addplot[only marks, lmUn] coordinates {(0.5,14.105) (0.75,33.266) (1,48.567) (1.5,72.666) (2,98.366) (2.5,107.373) (3.5,126.256)};
\node[font=\lmnote, text=lmmuted, anchor=north west, align=left, inner sep=0pt] at (axis cs:0.42,134) {order exact\\(Spearman 1.0)};
\end{axis}
\end{tikzpicture}
\caption{\textbf{Each kind of change moves the \dcode in its own direction.} (a)~PushT codes, placed by their direction on the first two principal components of the in-range codes and by their log distance from the default world's code (rings). Each kind of change has its own direction. (b)~Per factor: the code's position on the first principal component of that factor's codes (y-axis) against its setting (x-axis, evenly spaced). Shaded settings occur in training, and the code keeps moving beyond them. (c)~Reacher, trained without the teacher: each tilt's code distance from the default world's code, in units of that code's episode-to-episode spread ($\sigma$), orders a tilt that training never varies. LeWM backbone.}
\label{fig:code_map}
\end{figure*}
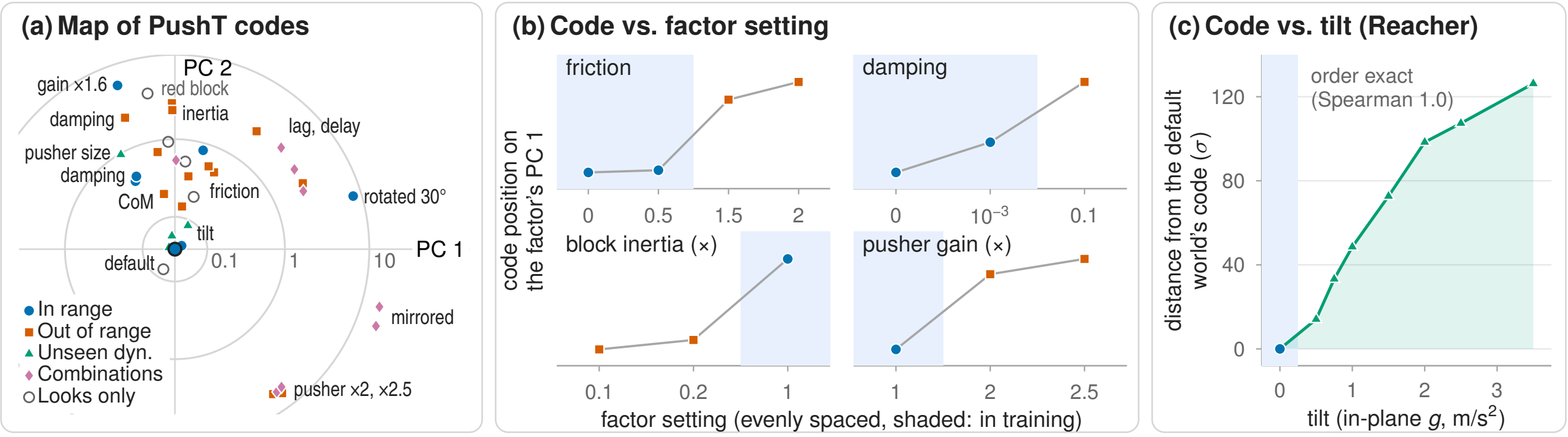

%% file: figures/fig_supp_predcurve.tex
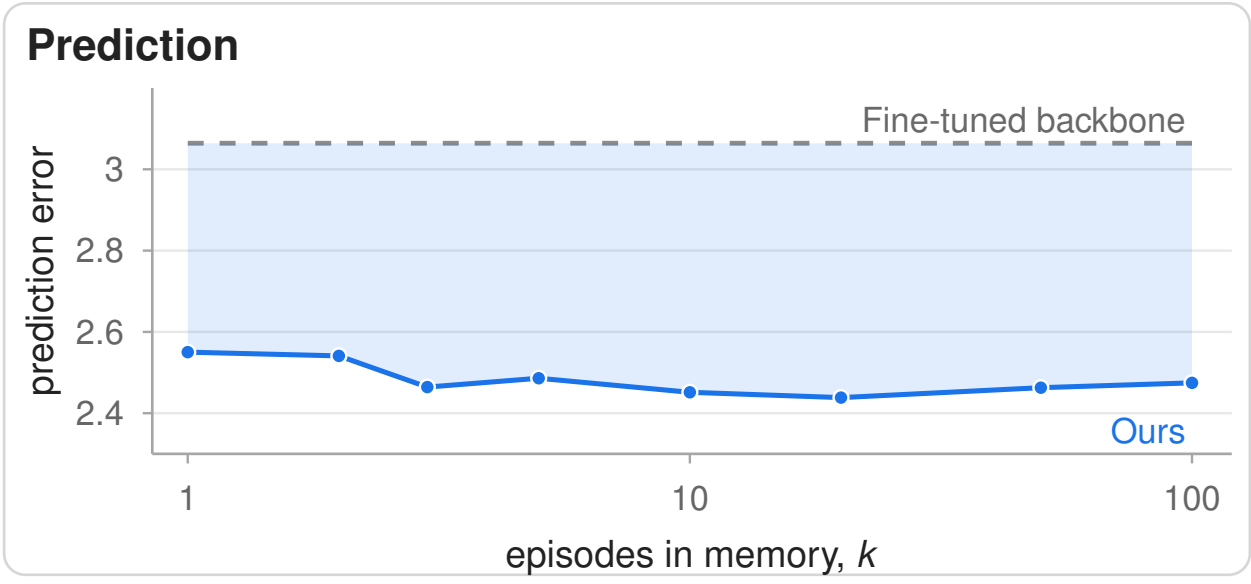
\begin{figure}[t]
\centering
\captionsetup{skip=3pt}%
\input{figures/lm_figstyle}
\pgfplotsset{fev/.style={lm, anchor=south west, width=7.14cm, height=2.42cm, xmajorgrids=false, xlabel style={yshift=3.5pt}, ylabel style={yshift=-4pt}}}
\begin{tikzpicture}[x=1cm, y=1cm]
\lmcard{0,0}{8.24,3.78}
\node[anchor=west, font=\lmtitle, text=lmink, inner sep=0pt] at (0.14,3.51) {Prediction};
\begin{axis}[fev, at={(0.98cm,0.80cm)}, xmode=log, xmin=0.85, xmax=120, xminorticks=false, ymin=2.3, ymax=3.2, xtick={1,10,100}, xticklabels={1,10,100},
  ytick={2.4,2.6,2.8,3.0}, xlabel={episodes in memory, \textit{k}}, ylabel={prediction error}]
\addplot[lmctrl, dashed, line width=0.9pt] table[x=k, y=control_median] {figures/fig56data/fig6c_summary.tsv};
\fill[lmours, opacity=0.13] (axis cs:1,2.550) -- (axis cs:2,2.541) -- (axis cs:3,2.464) -- (axis cs:5,2.486) -- (axis cs:10,2.451) -- (axis cs:20,2.438)
  -- (axis cs:50,2.463) -- (axis cs:100,2.475) -- (axis cs:100,3.064) -- (axis cs:1,3.064) -- cycle;
\addplot[lmours, line width=\lmlw, mark=*, mark size=0.8*\lmms pt, mark options={lmmk=lmours}] table[x=k, y=ours_median] {figures/fig56data/fig6c_summary.tsv};
\node[font=\lmnote, text=lmctrl!75!black, anchor=south east, inner sep=1pt] at (axis cs:100,3.075) {Fine-tuned backbone};
\node[font=\lmnote, text=lmours, anchor=north east, inner sep=1pt] at (axis cs:100,2.405) {Ours};
\end{axis}
\end{tikzpicture}
\caption{\textbf{Prediction error drops most with the first episode in memory.} One-step prediction error on held-out episodes against the number of episodes in memory, median over 31 PushT worlds, with the fine-tuned backbone as the reference. LeWM backbone.}
\label{fig:supp_predcurve}
\end{figure}

%% file: figures/fig_memory_cap.tex
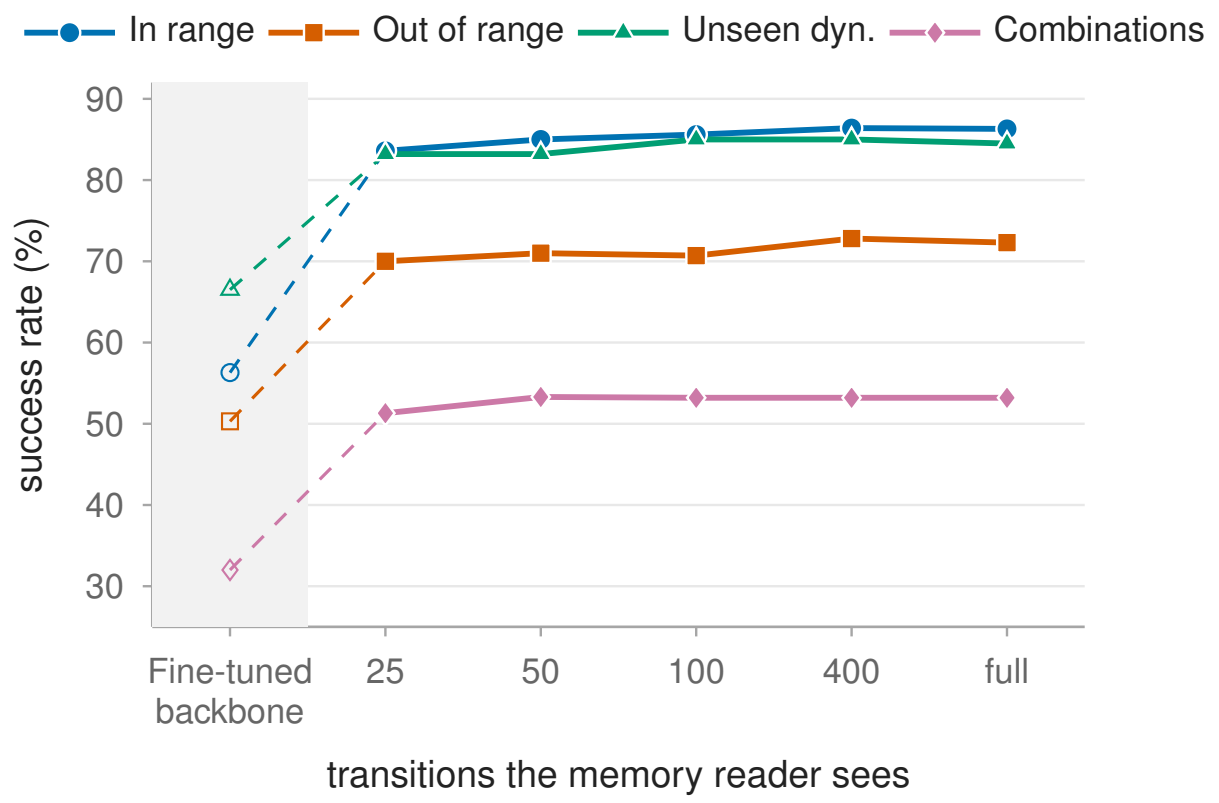
\begin{figure}[t]
\centering
\input{figures/lm_figstyle}%
\definecolor{gIn}{HTML}{0072B2}\definecolor{gOut}{HTML}{D55E00}\definecolor{gUn}{HTML}{009E73}\definecolor{gCo}{HTML}{CC79A7}
\begin{tikzpicture}
\begin{axis}[lm, width=0.74\linewidth, height=3.6cm, xmin=0.5, xmax=6.5, ymin=25, ymax=92,
  xtick={1,2,3,4,5,6}, xticklabels={{Fine-tuned\\backbone},25,50,100,400,full}, xticklabel style={align=center}, ytick={30,40,50,60,70,80,90},
  xlabel={transitions the memory reader sees}, ylabel={success rate (\%)},
  legend style={font=\lmnote, text=lmink, draw=none, fill=none, at={(0.5,1.02)}, anchor=south, legend columns=4, column sep=0pt}]
\fill[lmpanel] (axis cs:0.5,25) rectangle (axis cs:1.5,92);
\addplot[gIn, dashed, line width=0.6pt, forget plot] coordinates {(1,56.3) (2,83.6)};
\addplot[gOut, dashed, line width=0.6pt, forget plot] coordinates {(1,50.3) (2,70.0)};
\addplot[gUn, dashed, line width=0.6pt, forget plot] coordinates {(1,66.5) (2,83.2)};
\addplot[gCo, dashed, line width=0.6pt, forget plot] coordinates {(1,32.0) (2,51.3)};
\addplot[gIn, only marks, mark=o, mark size=1.6pt, forget plot] coordinates {(1,56.3)};
\addplot[gOut, only marks, mark=square, mark size=1.5pt, forget plot] coordinates {(1,50.3)};
\addplot[gUn, only marks, mark=triangle, mark size=1.9pt, forget plot] coordinates {(1,66.5)};
\addplot[gCo, only marks, mark=diamond, mark size=1.9pt, forget plot] coordinates {(1,32.0)};
\addplot[gIn, line width=1.1pt, mark=*, mark size=2.0pt, mark options={lmmk=gIn}] coordinates {(2,83.6) (3,85.0) (4,85.6) (5,86.4) (6,86.3)};
\addlegendentry{In range}
\addplot[gOut, line width=1.1pt, mark=square*, mark size=1.85pt, mark options={lmmk=gOut}] coordinates {(2,70.0) (3,71.0) (4,70.7) (5,72.8) (6,72.3)};
\addlegendentry{Out of range}
\addplot[gUn, line width=1.1pt, mark=triangle*, mark size=2.3pt, mark options={lmmk=gUn}] coordinates {(2,83.2) (3,83.2) (4,85.0) (5,85.0) (6,84.5)};
\addlegendentry{Unseen dyn.}
\addplot[gCo, line width=1.1pt, mark=diamond*, mark size=2.3pt, mark options={lmmk=gCo}] coordinates {(2,51.3) (3,53.3) (4,53.2) (5,53.2) (6,53.2)};
\addlegendentry{Combinations}
\end{axis}
\end{tikzpicture}
\caption{\textbf{Reading the last 50 transitions keeps the gain of the full memory.} PushT success as the memory reader sees only the most recent stored transitions, from the fine-tuned backbone (the same model trained without memory) to the full memory of \cref{tab:main_results}. From 50 transitions on, success stays within 5 points of the full memory in every group (non-inferiority, 90\% CI, margin fixed in advance). LeWM, 150 episodes per condition.}
\label{fig:memory_cap}
\end{figure}

%% file: figures/fig_shift_size.tex
\begin{figure}[t]
\centering
\input{figures/lm_figstyle}%
\definecolor{sigC}{HTML}{1A73E8}
\begin{tikzpicture}
\begin{axis}[lm, width=0.80\linewidth, height=3.8cm, xmin=-0.12, xmax=3.15, ymin=-20, ymax=75,
  xtick={0,1,2,3}, xticklabels={0,1,2,10}, ytick={-20,0,20,40,60},
  xlabel={shift, in units of the training range}, ylabel={gain over fine-tuned\\backbone (points)}, ylabel style={align=center},
  clip=false,
  error bars/error bar style={line width=0.35pt, opacity=0.45}, error bars/error mark=none]
\fill[lmband] (axis cs:-0.12,-20) rectangle (axis cs:1,75);
\node[font=\lmnote, text=lmmuted, anchor=center] at (axis cs:0.44,-12) {training range};
\addplot[lmink!70, line width=0.5pt, forget plot] coordinates {(-0.12,0) (3.15,0)};
\draw[white, line width=2.2pt] (axis cs:2.83,-20) -- (axis cs:2.89,-20);
\node[font=\lmnote, text=lmmuted, inner sep=0pt] at (axis cs:2.86,-20) {//};
\addplot[only marks, mark=*, lmctrl, mark options={fill=lmctrl}, mark size=1.5pt, error bars/.cd, y dir=both, y explicit] coordinates { (3.0000,9.33) += (0,9.34) -= (0,9.33) (1.6667,6.00) += (0,5.33) -= (0,5.33)};
\addplot[only marks, mark=o, lmctrl, mark size=1.5pt, error bars/.cd, y dir=both, y explicit] coordinates { (0.6667,0.67) += (0,3.33) -= (0,3.34) (0.4748,0.67) += (0,4.00) -= (0,4.00)};
\addplot[only marks, mark=square*, lmctrl, mark options={fill=lmctrl}, mark size=1.4pt, error bars/.cd, y dir=both, y explicit] coordinates { (0.0000,-6.00) += (0,6.67) -= (0,6.67) (1.5757,2.00) += (0,8.67) -= (0,8.00) (1.3368,0.67) += (0,8.00) -= (0,8.00) (1.5850,6.67) += (0,10.00) -= (0,10.67) (2.0000,12.00) += (0,11.33) -= (0,11.33) (2.0000,6.00) += (0,11.33) -= (0,10.67) (2.0000,-5.33) += (0,10.66) -= (0,10.67)};
\addplot[only marks, mark=square, lmctrl, mark size=1.4pt, error bars/.cd, y dir=both, y explicit] coordinates { (1.5850,10.67) += (0,10.66) -= (0,10.67)};
\addplot[only marks, mark=triangle*, lmctrl, mark options={fill=lmctrl}, mark size=1.8pt, error bars/.cd, y dir=both, y explicit] coordinates { (0.5575,2.00) += (0,6.00) -= (0,5.33) (0.6415,-2.00) += (0,5.33) -= (0,5.33) (0.6667,1.33) += (0,5.34) -= (0,5.33) (0.5850,-4.00) += (0,4.67) -= (0,4.67) (0.5000,-2.00) += (0,3.33) -= (0,3.33) (0.6667,4.00) += (0,6.00) -= (0,6.00) (1.2500,6.67) += (0,7.33) -= (0,6.67) (1.6500,4.00) += (0,5.33) -= (0,5.33) (1.3750,1.33) += (0,4.00) -= (0,4.00)};
\addplot[only marks, mark=*, sigC, mark options={fill=sigC}, mark size=1.5pt, error bars/.cd, y dir=both, y explicit] coordinates { (0.0000,13.33) += (0,6.67) -= (0,6.00) (0.6667,14.67) += (0,6.66) -= (0,6.67) (0.1000,17.33) += (0,7.34) -= (0,7.33) (0.6667,20.00) += (0,8.00) -= (0,8.00) (0.1667,64.67) += (0,8.00) -= (0,8.00) (1.0000,58.67) += (0,8.00) -= (0,8.00) (0.4748,21.33) += (0,8.67) -= (0,8.66) (2.0000,13.33) += (0,8.67) -= (0,8.00) (2.6667,16.00) += (0,8.67) -= (0,9.33) (2.6667,20.67) += (0,9.33) -= (0,9.34) (1.9125,22.67) += (0,8.00) -= (0,8.00) (1.3368,23.33) += (0,7.35) -= (0,7.33) (1.4748,59.33) += (0,8.00) -= (0,8.00) (1.9495,58.00) += (0,8.00) -= (0,8.00) (1.3333,19.33) += (0,7.34) -= (0,6.66) (2.0000,8.00) += (0,6.67) -= (0,6.67) (1.3333,8.67) += (0,6.66) -= (0,6.67) (2.0000,20.67) += (0,7.33) -= (0,7.34) (2.0000,24.00) += (0,8.67) -= (0,8.67) (2.0000,43.33) += (0,8.67) -= (0,8.66) (2.0000,17.33) += (0,8.00) -= (0,8.00)};
\addplot[only marks, mark=o, sigC, mark size=1.5pt, error bars/.cd, y dir=both, y explicit] coordinates { (0.4748,13.33) += (0,8.67) -= (0,9.33) (1.4748,56.67) += (0,8.00) -= (0,8.00) (2.0000,14.00) += (0,8.67) -= (0,8.67)};
\addplot[only marks, mark=square*, sigC, mark options={fill=sigC}, mark size=1.4pt, error bars/.cd, y dir=both, y explicit] coordinates { (0.4286,55.33) += (0,8.00) -= (0,8.00) (0.3237,42.00) += (0,8.67) -= (0,8.67) (0.3333,45.33) += (0,10.00) -= (0,10.00) (0.3219,16.67) += (0,9.33) -= (0,8.67) (2.0000,47.33) += (0,8.00) -= (0,8.00)};
\addplot[only marks, mark=square, sigC, mark size=1.4pt, error bars/.cd, y dir=both, y explicit] coordinates { (1.5757,10.00) += (0,7.33) -= (0,7.33)};
\addplot[only marks, mark=triangle*, sigC, mark options={fill=sigC}, mark size=1.8pt, error bars/.cd, y dir=both, y explicit] coordinates { (0.0000,-6.00) += (0,4.00) -= (0,4.67) (1.2000,10.67) += (0,6.66) -= (0,6.67) (1.2000,13.33) += (0,6.67) -= (0,6.00) (1.2500,18.67) += (0,6.66) -= (0,6.67) (1.2750,25.33) += (0,7.34) -= (0,7.33) (1.2500,20.00) += (0,7.33) -= (0,7.33) (1.6500,13.33) += (0,6.00) -= (0,5.33) (1.3750,9.33) += (0,6.67) -= (0,6.66) (1.2500,20.00) += (0,6.67) -= (0,6.67)};
\end{axis}
\end{tikzpicture}\\[1pt]
{\lmtick\color{lmink}$\bullet$ PushT \enspace $\blacksquare$ Reacher \enspace $\blacktriangle$ Cube \enspace \textcolor{sigC}{blue}: $p<0.05$ \enspace open: visual part}
\caption{\textbf{Gains hold outside the training range.} Each point is one test condition: the size of its shift, in units of the training range (shaded, with combinations at their larger shift), against the paired gain of ours over the fine-tuned backbone in success rate ($\pm$95\% CI, 150 episodes). Blue: $p<0.05$ (exact McNemar). Conditions without a training range (tilts, axis swaps and flips) are listed in \cref{sec:supp_conditions}. LeWM backbone.}
\label{fig:shift_size_supp}
\end{figure}
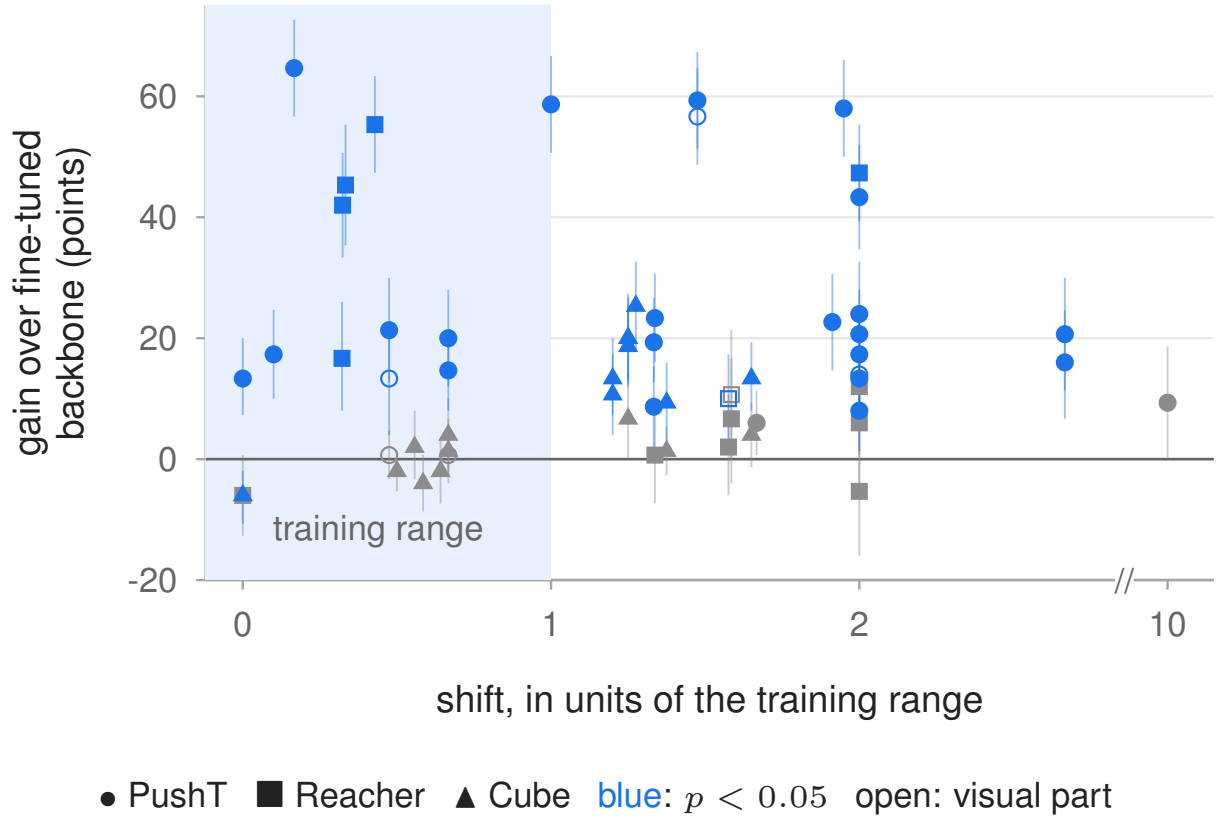

%% file: sections/supp/variants.tex
\paragraph{Training seeds and teacher ablation.}
\Cref{tab:supp_variants} summarizes the three-seed replications on PushT and teacher-free Cube (all test conditions of \cref{tab:main_results}) together with the PushT teacher ablation. Both three-seed models vary by at most 0.8 points standard deviation across seeds, and the teacher improves PushT in every group while teacher-free Cube matches or exceeds PhysiCalWorld in every group.
\begin{table}[t]
\centering
\footnotesize
\setlength{\tabcolsep}{2.6pt}
\resizebox{\linewidth}{!}{%
\begin{tabular}{@{}llcccc@{}}
\toprule
Model or comparison & Env & In range & Out of range & Unseen dyn. & Comb. \\
\midrule
\multicolumn{6}{@{}l}{\textit{Three training seeds, success (\%, mean $\pm$ std across seeds)}} \\
\lememoire (3 seeds) & PushT & $86.5 \pm 0.2$ & $72.5 \pm 0.6$ & $85.1 \pm 0.6$ & $53.3 \pm 0.7$ \\
\lememoire, no teacher (3 seeds) & Cube & $86.1 \pm 0.8$ & $84.0 \pm 0.8$ & $65.4 \pm 0.6$ & $78.9 \pm 0.8$ \\
\midrule
\multicolumn{6}{@{}l}{\textit{Teacher ablation, paired difference in points}} \\
Teacher $-$ none & PushT & $+3.7^\ast$ & $+6.5^\ast$ & $+4.2^\ast$ & $+5.9^\ast$ \\
\bottomrule
\end{tabular}}
\caption{\textbf{Three training seeds on PushT and teacher-free Cube, and the PushT teacher ablation.} LeWM backbone, 150 episodes per condition on the exact test sets of \cref{tab:main_results}. Top: mean $\pm$ std over three training seeds. Bottom: paired difference between teacher and no-teacher models on PushT ($^\ast$\,$p < 0.05$, exact McNemar).}
\label{tab:supp_variants}
\end{table}

%% file: tables/supp_cube_default_versions.tex
\begin{table}[t]
\centering
\small
\setlength{\tabcolsep}{6pt}
\begin{tabular}{@{}lccc@{}}
\toprule
Test group & PLDM~\cite{sobal2025pldm} & GCBC~\cite{ghosh2019gcsl} & Ours \\
\midrule
In range        & 59.6\pmstd{4.6} & \textbf{91.9}\pmstd{0.2} & 86.4\pmstd{0.9} \\
Out of range    & 54.2\pmstd{4.5} & 71.9\pmstd{2.1} & \textbf{82.6}\pmstd{1.6} \\
Unseen dynamics & 49.2\pmstd{5.2} & 60.3\pmstd{2.6} & \textbf{61.4}\pmstd{2.6} \\
Combinations    & 53.3\pmstd{4.4} & 68.7\pmstd{1.7} & \textbf{80.4}\pmstd{2.0} \\
\bottomrule
\end{tabular}
\caption{\textbf{Cube, default versions of PLDM and GCBC} (success rate in \%, 150 episodes per condition, $\pm$ over three blocks of 50 episodes).
These versions are trained on LeWM's default data. \Cref{tab:main_results} shows the versions fine-tuned on our training conditions, which score higher on the validation episodes.
}
\label{tab:\currfilebase}
\end{table}

%% file: tables/supp_reacher_default_versions.tex
\begin{table}[t]
\centering
\small
\setlength{\tabcolsep}{6pt}
\begin{tabular}{@{}lccc@{}}
\toprule
Test group & PLDM~\cite{sobal2025pldm} (a) & GCBC~\cite{ghosh2019gcsl} (a) & Ours \\
\midrule
In range        & 48.9\pmstd{3.2} & 1.3\pmstd{1.2} & \textbf{80.5}\pmstd{4.4} \\
Out of range    & 14.1\pmstd{2.2} & 1.4\pmstd{1.0} & \textbf{28.9}\pmstd{5.5} \\
Unseen dynamics & ---             & ---            & \textbf{35.7}\pmstd{1.9} \\
Combinations    & 27.5\pmstd{2.8} & 1.2\pmstd{0.8} & \textbf{39.7}\pmstd{9.1} \\
\bottomrule
\end{tabular}
\caption{\textbf{Reacher, default versions of PLDM and GCBC.} Success (\%) of the versions trained on LeWM's default data. \cref{tab:main_results} shows the versions fine-tuned on our training worlds. The default versions were not evaluated on the Unseen dynamics conditions. LeWM backbone for ours, 150 episodes per condition.}
\label{tab:\currfilebase}
\end{table}

%% file: tables/supp_changing_world.tex
\begin{table}[t]
\centering
\footnotesize
\setlength{\tabcolsep}{4pt}
\begin{tabular}{@{}lccc@{}}
\toprule
& & \multicolumn{2}{c}{Episodes since the switch} \\
\cmidrule(l){3-4}
Model & All episodes & 0--4 & 15--24 \\
\midrule
Ours, last 50 transitions      & \textbf{66.3} & \textbf{52.5} & \textbf{71.5} \\
Ours, full memory              & 56.8          & 48.5          & 57.0 \\
Fine-tuned backbone            & 51.0          & 49.5          & 55.8 \\
AdaJEPA                        & 48.2          & 48.5          & 52.5 \\
\bottomrule
\end{tabular}
\caption{\textbf{A short memory keeps adapting when the world changes.} Planning success (\%) on PushT when the world switches every 25 episodes, over all episodes and over the first and last episodes after each switch. Best per column in \textbf{bold}. LeWM, 8 sequences of 150 episodes.}
\label{tab:supp_changing_world}
\vspace{-4pt}%
\end{table}

%% file: sections/supp/architectures.tex
\input{figures/supp_arch_variants}
In LeWM, and in TS on Cube, the code uses the backbone's own action path. In LeWM it is added to the action embedding, which modulates every predictor block through adaptive layer normalization (AdaLN)~\cite{peebles2023dit,maes2026lewm}, and in TS to the input action embedding.
The memory can enter the predictor in more than one place.
On the TS backbone, we trained the reader with the code entering every predictor block through cross-attention, as in \cref{tab:main_results} on PushT and Reacher, and with the code added to the action part of each frame token, which leaves the predictor unchanged (\cref{fig:supp_arch_variants}).
On PushT and Reacher, both variants succeed more often than the fine-tuned backbone in every group we measured (\cref{tab:supp_arch_variants}), so there the gain does not hinge on one way of conditioning the predictor.
On Cube, the TS model of \cref{tab:main_results} adds the code to the action, trained with the teacher code. It also succeeds more often than the fine-tuned backbone in every group (\cref{tab:supp_control}).
\begin{table}[t]
\centering
\footnotesize
\setlength{\tabcolsep}{2.4pt}
\resizebox{\linewidth}{!}{%
\begin{tabular}{@{}lcccccc@{}}
\toprule
& \multicolumn{3}{c}{PushT} & \multicolumn{3}{c}{Reacher} \\
\cmidrule(lr){2-4} \cmidrule(l){5-7}
Test group & cross-attn. & action & \begin{tabular}[c]{@{}c@{}}Fine-tuned\\backbone\end{tabular} & cross-attn. & action & \begin{tabular}[c]{@{}c@{}}Fine-tuned\\backbone\end{tabular} \\
\midrule
In range        & \textbf{83.0} & 79.6 & 62.8 & 87.2 & \textbf{90.7} & 70.8 \\
Out of range    & \textbf{66.9} & 64.6 & 54.7 & \textbf{62.3} & 60.5 & 23.1 \\
Unseen dynamics & \textbf{78.7} & 71.2 & 58.7 & \textbf{38.1} & ---  & 22.8 \\
Combinations    & \textbf{51.8} & 49.0 & 44.1 & 65.3 & \textbf{67.2} & 25.8 \\
\bottomrule
\end{tabular}}
\caption{\textbf{Both entry points beat the fine-tuned backbone on PushT and Reacher.} Success (\%) of the TS model with the code entering through cross-attention (\cref{tab:main_results}) or added to the action, and of the fine-tuned backbone. Every difference from the fine-tuned backbone has $p \le 3\times10^{-4}$ (paired exact McNemar, Holm-corrected per environment). ---: not evaluated on the Unseen dynamics conditions. 150 episodes per condition.}
\label{tab:supp_arch_variants}
\end{table}

%% file: figures/supp_arch_variants.tex
\par\medskip\noindent\begin{minipage}{\columnwidth}
\centering
\input{figures/lm_figstyle}
\tikzset{
  bx/.style={draw=lmcardb, line width=0.5pt, fill=white, rounded corners=2.5pt, font=\lmlab, text=lmink, align=center, inner sep=2.2pt},
  sub/.style={bx, font=\lmnote, inner sep=1.4pt, minimum height=0.42cm},
  our/.style={bx, draw=lmours, fill=lmours!9, line width=0.6pt},
  pill/.style={bx, rounded corners=4.5pt, font=\lmnote, inner sep=1.6pt},
  arr/.style={-{Stealth[length=3.2pt]}, lmmuted, line width=0.55pt},
  oarr/.style={-{Stealth[length=3.4pt]}, lmours, line width=0.8pt, dash pattern=on 2.2pt off 1.4pt}}
\resizebox{\linewidth}{!}{%
\begin{tikzpicture}[x=1cm, y=1cm]
\lmcard{0,0}{4.6,5.6}
\lmhead{0.15,5.3}{a}{Code token in every block}
\node[sub, fill=lmpanel, minimum width=1.75cm] at (1.25,4.6) {visual 384};
\node[sub, fill=lmpanel, minimum width=0.6cm] at (2.5,4.6) {10};
\node[sub, fill=lmpanel, minimum width=0.6cm] at (3.15,4.6) {10};
\node[font=\lmnote, text=lmmuted] at (2.2,4.97) {frame token: image $|$ proprio. $|$ action};
\node[pill, our] (va) at (4.05,3.45) {$\hat e$ (404)};
\fill[white, draw=lmcardb, line width=0.5pt, rounded corners=3pt] (0.2,0.55) rectangle (3.5,3.85);
\node[anchor=west, font=\lmnote, text=lmmuted] at (0.27,3.65) {6 blocks};
\foreach \y in {3.15, 2.55, 1.35} {
  \node[sub, fill=lmpanel, minimum width=0.9cm] at (0.78,\y) {self-attn.};
  \node[sub, our, minimum width=0.95cm] at (1.9,\y) {cross-attn.};
  \node[sub, fill=lmpanel, minimum width=0.6cm] at (2.95,\y) {MLP};}
\node[font=\lmlab, text=lmfaint] at (1.85,1.98) {$\vdots$};
\draw[arr] (2.2,4.38) -- (2.2,3.85);
\draw[oarr] (va.south) |- (3.5,2.25);
\node[font=\lmnote, text=lmours, anchor=west, align=left] at (3.55,1.85) {to every\\[-2pt]block};
\node[font=\lmnote, text=lmours, align=center] at (2.3,0.27) {gated by $\tanh(\alpha_\ell)$, $\alpha_\ell=0$ at start};
\lmcard{4.9,0}{9.5,5.6}
\lmhead{5.05,5.3}{b}{Code added to the action}
\node[sub, fill=lmpanel, minimum width=1.75cm] at (6.15,4.6) {visual 384};
\node[sub, fill=lmpanel, minimum width=0.6cm] at (7.4,4.6) {10};
\node[sub, our, minimum width=0.9cm] (vb3) at (8.25,4.6) {10${}+\hat e$};
\node[font=\lmnote, text=lmmuted] at (7.1,4.97) {frame token: image $|$ proprio. $|$ action};
\node[pill, our] (vb) at (8.85,3.45) {$\hat e$ (10)};
\draw[oarr] (vb.north) -- (8.85,4.0) -| ([xshift=-0.2cm]vb3.south east);
\fill[white, draw=lmcardb, line width=0.5pt, rounded corners=3pt] (5.15,0.55) rectangle (8.3,3.85);
\node[anchor=west, font=\lmnote, text=lmmuted] at (5.22,3.65) {6 blocks, unchanged};
\foreach \y in {3.15, 2.55, 1.35} {
  \node[sub, fill=lmpanel, minimum width=1.55cm] at (6.15,\y) {self-attention};
  \node[sub, fill=lmpanel, minimum width=1.0cm] at (7.6,\y) {MLP};}
\node[font=\lmlab, text=lmfaint] at (6.75,1.98) {$\vdots$};
\draw[arr] (7.1,4.38) -- (7.1,3.85);
\node[font=\lmnote, text=lmmuted, align=center] at (7.2,0.27) {no new layers in the predictor};
\end{tikzpicture}}

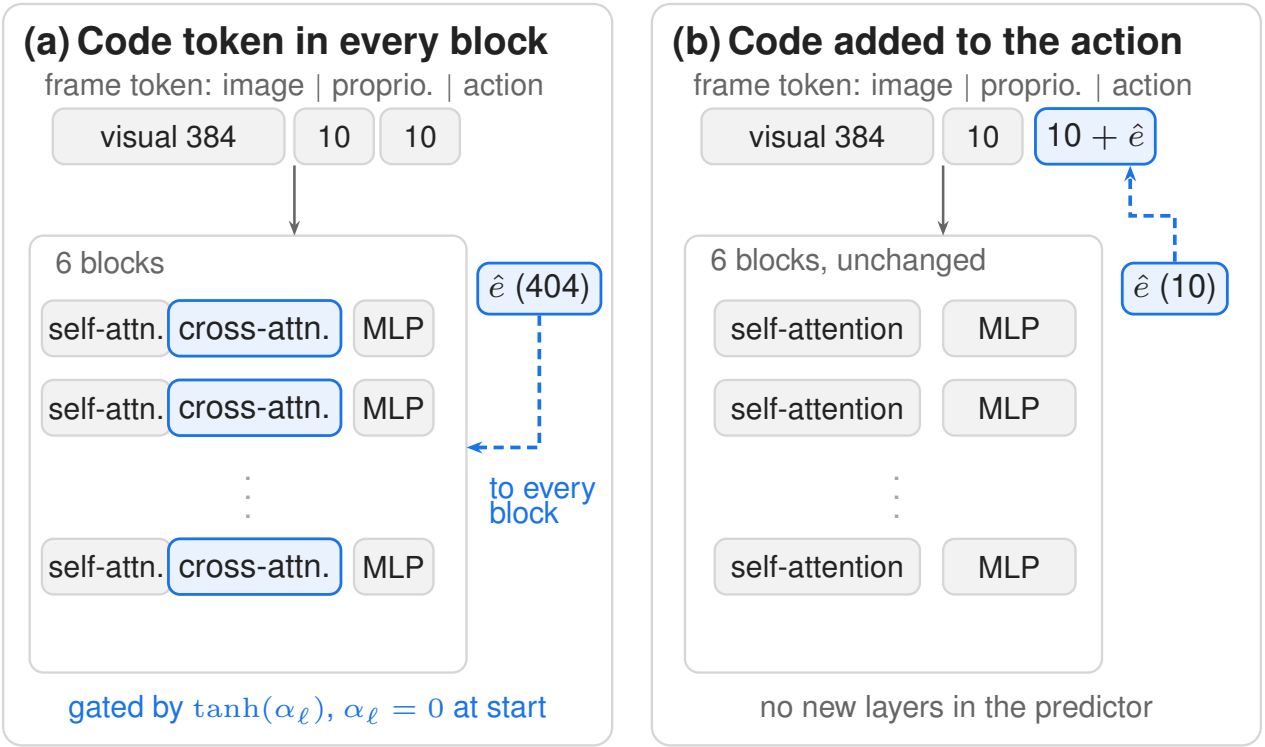
\captionof{figure}{\textbf{Two entry points for the dynamics code in TS.} \textbf{(a)}~Every predictor block cross-attends to a per-frame code token through a gate that starts closed. \textbf{(b)}~The code is added to the action part of each frame token, and the predictor is unchanged. Both read the memory with the same reader (\cref{fig:supp_arch}b). Blue: added by \lememoire.}
\label{fig:supp_arch_variants}
\end{minipage}\par\medskip

%% file: sections/supp/cost_protocol.tex
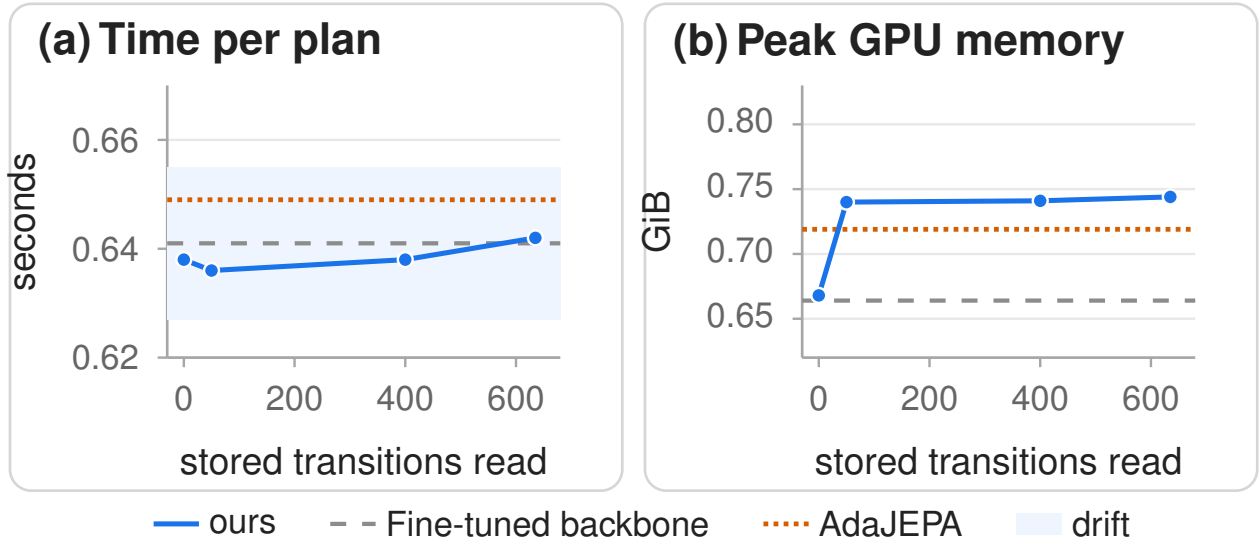
\begin{figure}[t]
\centering
\captionsetup{skip=3pt}
\input{figures/fig_supp_cost}
\caption{\textbf{Reading the memory adds no time beyond run-to-run variation and 0.08\,GiB of GPU memory.} Median time per plan (a) and peak GPU memory (b) of ours against the number of stored transitions the reader sees, next to the fine-tuned backbone and AdaJEPA. The band around the fine-tuned backbone spans the largest change in any untouched method's time between two runs (0.014\,s). PushT, one H100.}
\label{fig:supp_cost}
\end{figure}

We time the fine-tuned backbone, AdaJEPA, PhysiCalWorld and ours on PushT in one run on one H100 GPU, running each method as its own process with no other process on the GPU.
Each method plans episodes 100--129 of the default world and two shifted worlds in our harness, in deterministic mode. A warm-up on five episodes and the first episode of each world are not timed.
The time per plan covers the planning call, including AdaJEPA's gradient step, and the memory reads made until the next plan, synchronized with the GPU. The peak GPU memory is the largest memory allocated during a plan.
Our memory starts with the transitions stored during episodes 0--99 of the run in \cref{tab:main_results} and grows during the timed episodes, and the reader sees none of them, the last 50, the last 400 or all stored transitions.
The reader recomputes the \dcode after every executed action, but a plan uses only the code computed immediately before it. We skip the other reads, which leaves every planned action unchanged bit for bit. We also encode each stored transition once and cache it.
With both changes, the time per plan stays within the run-to-run variation of 0.014\,s at every memory size, and the peak memory exceeds that of the fine-tuned backbone by 0.08\,GiB (\cref{fig:supp_cost}).

%% file: figures/fig_supp_cost.tex
\input{figures/lm_figstyle}
\pgfplotsset{fcost/.style={lm, width=2.60cm, height=1.80cm, xmin=-30, xmax=680, xtick={0,200,400,600}, xlabel={stored transitions read}}}
\begin{tikzpicture}[x=1cm, y=1cm]
\lmcard{0,0}{4.05,3.22}
\lmcard{4.2,0}{8.25,3.22}
\lmhead{0.18,2.96}{a}{Time per plan}
\lmhead{4.38,2.96}{b}{Peak GPU memory}
\begin{axis}[fcost, at={(1.04cm,0.88cm)}, ymin=0.62, ymax=0.67, ytick={0.62,0.64,0.66}, yticklabel style={/pgf/number format/fixed, /pgf/number format/precision=2},
  ylabel={seconds}]
\fill[lmband!70] (axis cs:-30,0.627) rectangle (axis cs:680,0.655);
\draw[lmctrl, dashed, line width=0.8pt] (axis cs:-30,0.641) -- (axis cs:680,0.641);
\draw[gOut, densely dotted, line width=0.9pt] (axis cs:-30,0.649) -- (axis cs:680,0.649);

\addplot[lmours, line width=\lmlw, mark=*, mark size=0.8*\lmms pt, mark options={lmmk=lmours}] coordinates {(0,0.638) (50,0.636) (400,0.638) (635,0.642)};
\end{axis}
\begin{axis}[fcost, at={(5.24cm,0.88cm)}, ymin=0.62, ymax=0.83, ytick={0.65,0.70,0.75,0.80}, yticklabels={0.65,0.70,0.75,0.80},
  ylabel={GiB}]
\draw[lmctrl, dashed, line width=0.8pt] (axis cs:-30,0.664) -- (axis cs:680,0.664);
\draw[gOut, densely dotted, line width=0.9pt] (axis cs:-30,0.719) -- (axis cs:680,0.719);

\addplot[lmours, line width=\lmlw, mark=*, mark size=0.8*\lmms pt, mark options={lmmk=lmours}] coordinates {(0,0.668) (50,0.740) (400,0.741) (635,0.744)};
\end{axis}
\begin{scope}[every node/.style={font=\lmnote, text=lmink, anchor=west, inner sep=0pt}]
\draw[lmours, line width=\lmlw] (0.95,-0.22) -- (1.25,-0.22); \node (lgA) at (1.31,-0.22) {ours};
\draw[lmctrl, dashed, line width=0.8pt] ([xshift=0.35cm]lgA.east) -- ++(0.30,0); \node (lgB) at ([xshift=0.71cm]lgA.east) {Fine-tuned backbone};
\draw[gOut, densely dotted, line width=0.9pt] ([xshift=0.35cm]lgB.east) -- ++(0.30,0); \node (lgC) at ([xshift=0.71cm]lgB.east) {AdaJEPA};
\fill[lmband!70] ([xshift=0.35cm,yshift=-0.07cm]lgC.east) rectangle ++(0.30,0.14); \node (lgD) at ([xshift=0.71cm]lgC.east) {drift};
\end{scope}
\end{tikzpicture}\par

%% file: tables/cost.tex
\begin{table}[t]
\centering
\scriptsize
\setlength{\tabcolsep}{2.5pt}
\resizebox{\ifdim\width>\linewidth\linewidth\else\width\fi}{!}{%
\begin{tabular}{@{}lccccc@{}}
\toprule
 & Weights & Test-world data & Memory & & Peak \\
Method & at test time & before an episode & across episodes & Time/plan & VRAM \\
\midrule
Fine-tuned backbone & fixed & none & --- & 0.64\,s & 0.66\,GiB \\
AdaJEPA & 1 grad.\ step per replan & none & no & 0.65\,s & 0.72\,GiB \\
PhysiCalWorld & fixed & 50-step clip & no & 0.64\,s & 0.74\,GiB \\
Ours & fixed & none & yes & 0.64\,s & 0.74\,GiB \\
\bottomrule
\end{tabular}%
}
\caption{\textbf{What each method needs at test time.} Median time and peak GPU memory per plan on PushT (one H100). Time differences are below run-to-run variation (0.014\,s). Ours changes no weights and needs no test-world data before an episode. For timing, ours reads its memory only when planning, with stored transitions cached.}
\label{tab:cost}
\end{table}

%% file: figures/supp_qual.tex
\begin{figure*}[p]
\centering
\input{figures/lm_figstyle}
\newcommand{\qf}[2]{\tikz[baseline=(q.center)]{\node[inner sep=0pt] (q) {\includegraphics[width=0.121\linewidth]{figures/qual/#1}}; \node[anchor=south east, inner sep=0.8pt, fill=white, fill opacity=0.8, text opacity=1, font=\lmnote, text=lmink] at (q.south east) {#2};}}
\newcommand{\ql}[2]{\parbox[c]{1.25cm}{\raggedleft{\lmlab\textbf{#1}}\\[-1pt]{\lmnote #2}}}
\setlength{\tabcolsep}{0.8pt}
\renewcommand{\arraystretch}{0.9}
\begin{tabular}{@{}r@{\hspace{3pt}}cccccc@{\hspace{8pt}}c@{}}
\multicolumn{8}{@{}l}{\lmtitle (a)\,PushT, tilted table (unseen dynamics)} \\[1pt]
\ql{\textcolor{lmours}{Ours}}{succeeds} & \qf{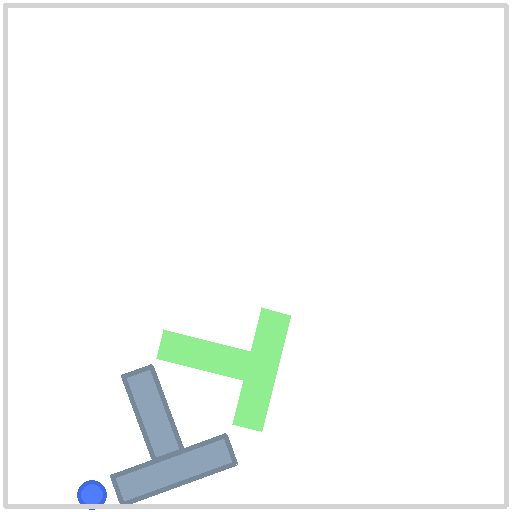}{0} & \qf{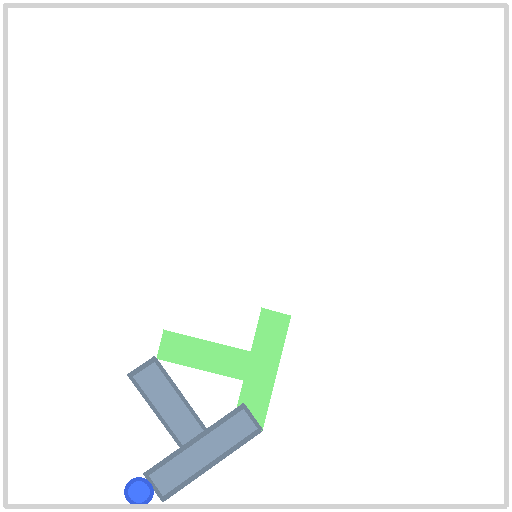}{5} & \qf{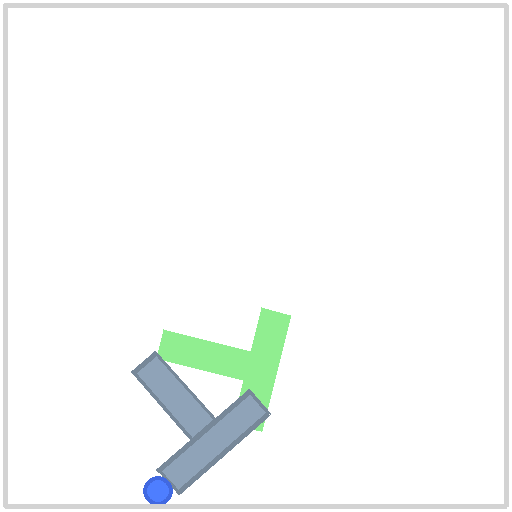}{10} & \qf{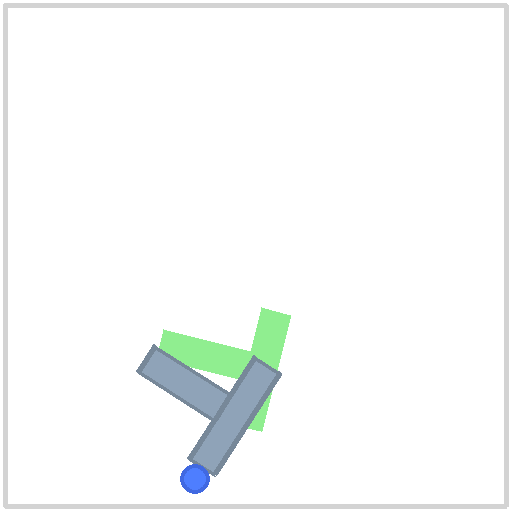}{15} & \qf{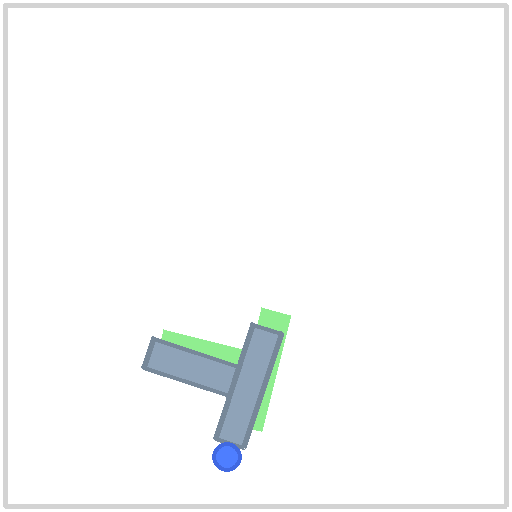}{20} & \qf{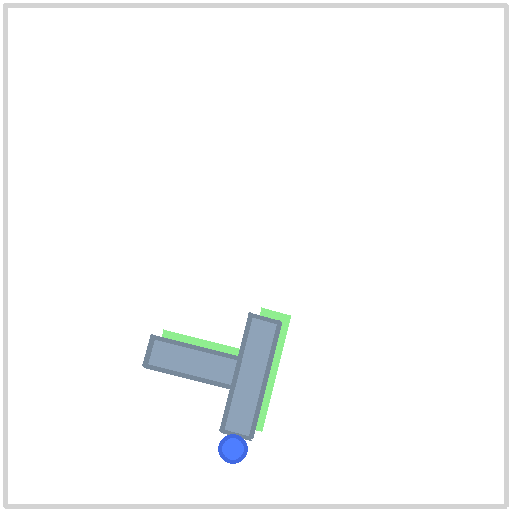}{24} & \qf{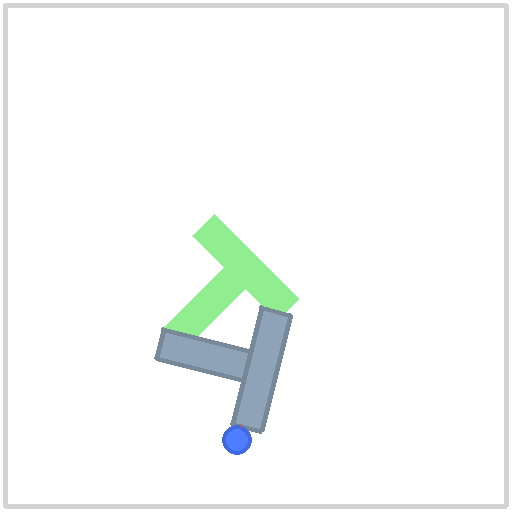}{goal} \\ \noalign{\vspace{3pt}}
\ql{AdaJEPA}{fails} & \qf{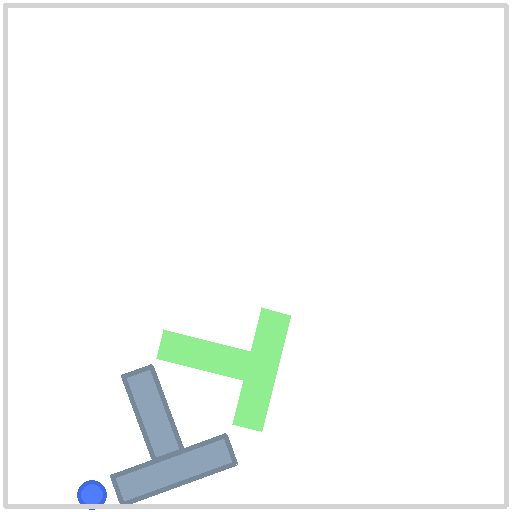}{0} & \qf{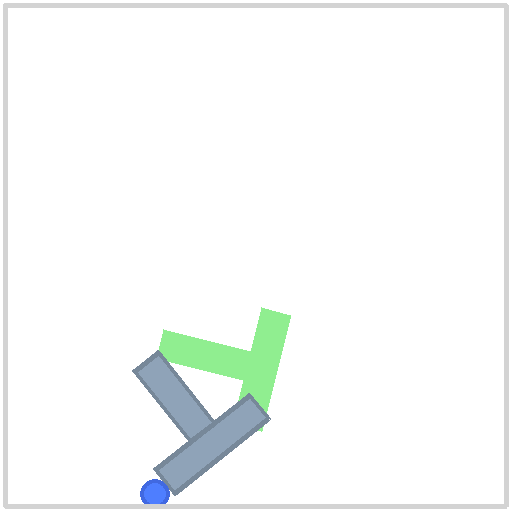}{10} & \qf{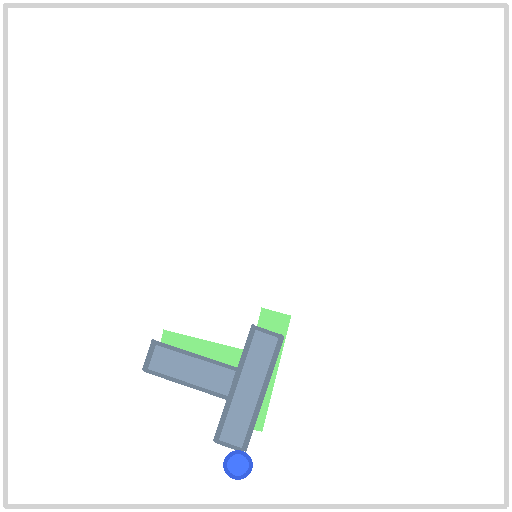}{20} & \qf{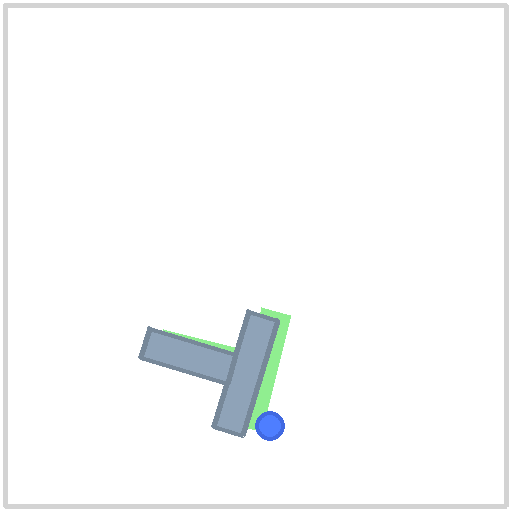}{30} & \qf{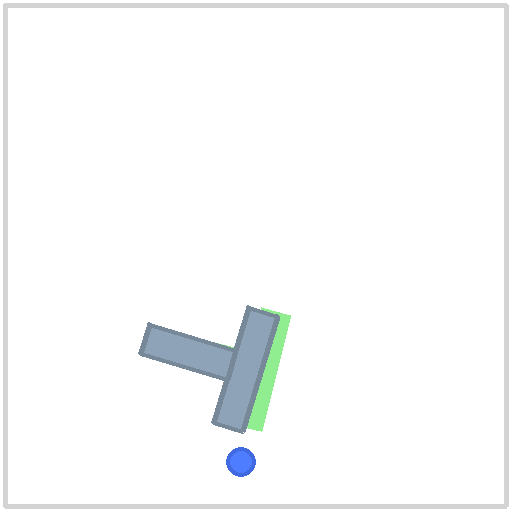}{40} & \qf{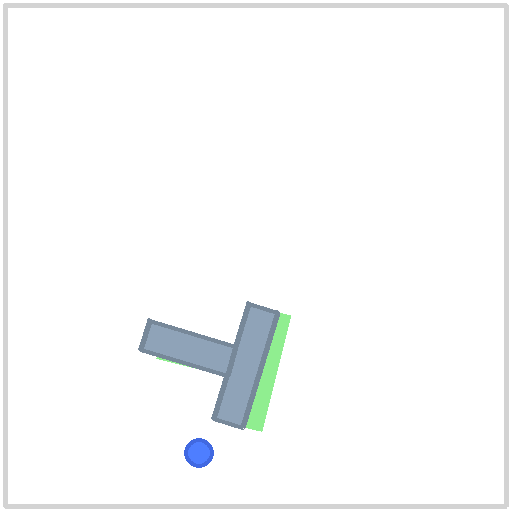}{50} & \\ \noalign{\vspace{3pt}}
\ql{Fine-tuned backbone}{fails} & \qf{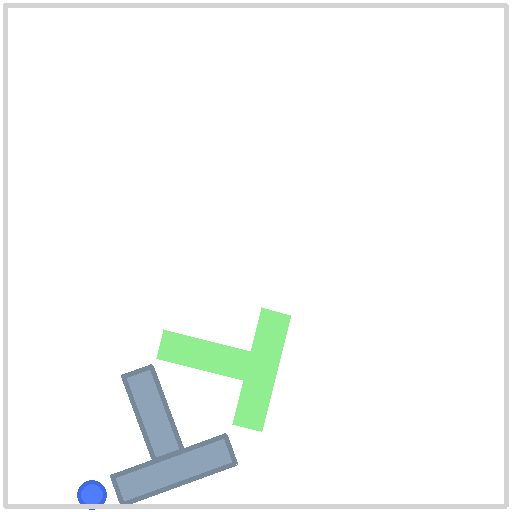}{0} & \qf{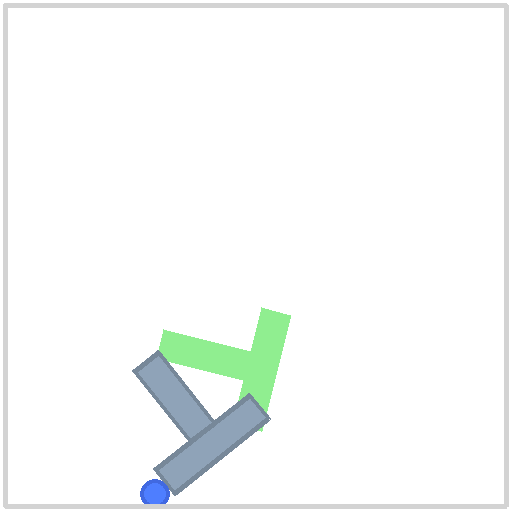}{10} & \qf{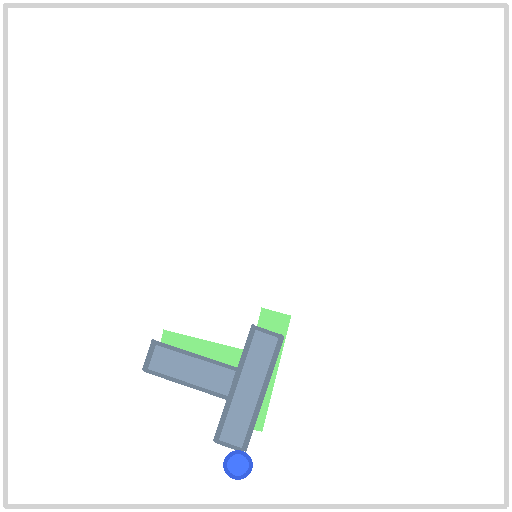}{20} & \qf{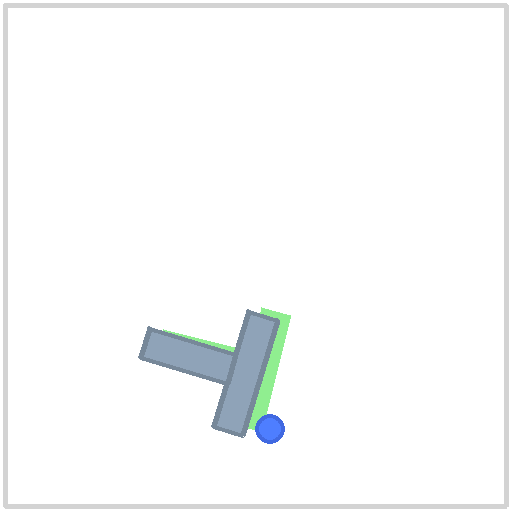}{30} & \qf{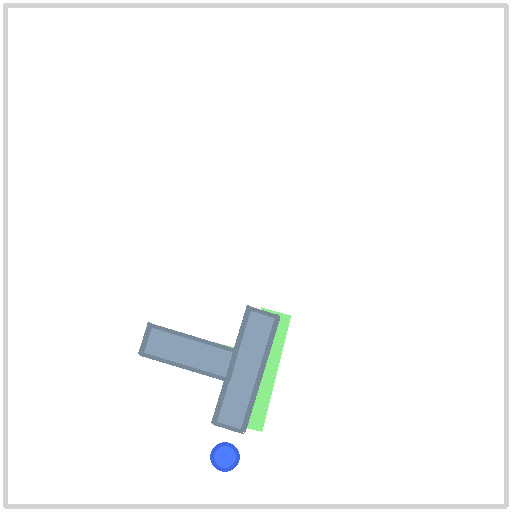}{40} & \qf{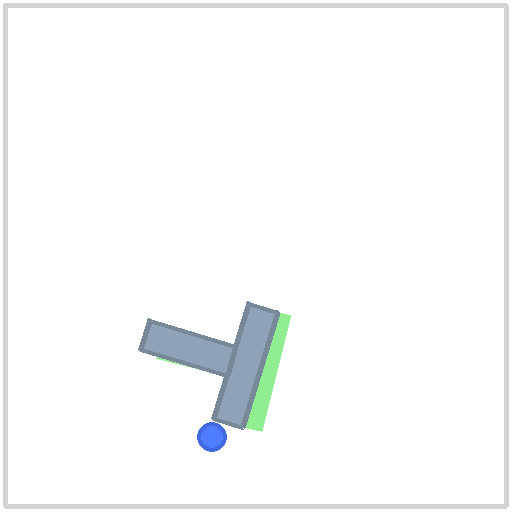}{50} & \\ \noalign{\vspace{10pt}}
\multicolumn{8}{@{}l}{\lmtitle (b)\,Reacher, out of range} \\[1pt]
\ql{\textcolor{lmours}{Ours}}{succeeds} & \qf{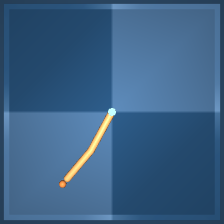}{0} & \qf{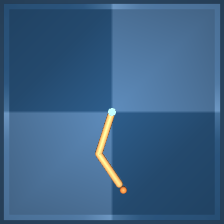}{7} & \qf{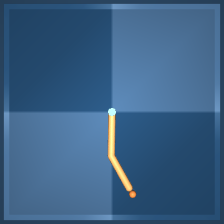}{14} & \qf{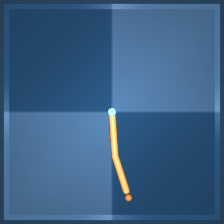}{20} & \qf{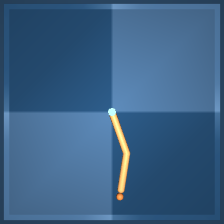}{27} & \qf{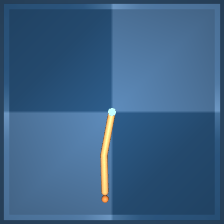}{34} & \qf{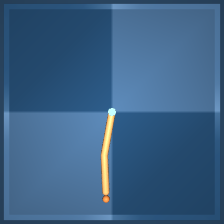}{goal} \\ \noalign{\vspace{3pt}}
\ql{AdaJEPA}{fails} & \qf{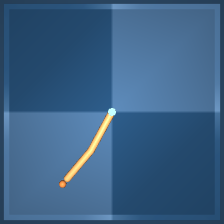}{0} & \qf{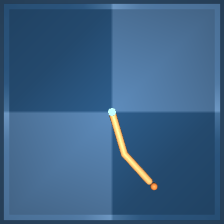}{10} & \qf{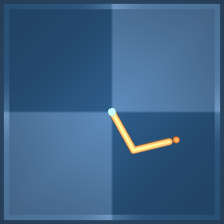}{20} & \qf{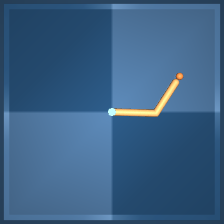}{30} & \qf{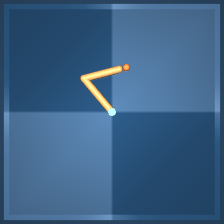}{40} & \qf{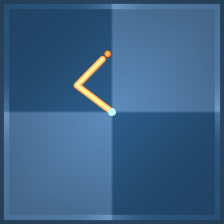}{50} & \\ \noalign{\vspace{3pt}}
\ql{Fine-tuned backbone}{fails} & \qf{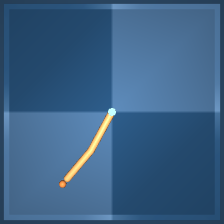}{0} & \qf{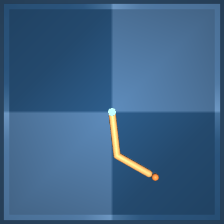}{10} & \qf{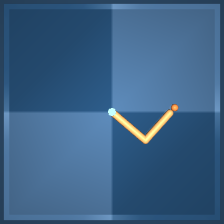}{20} & \qf{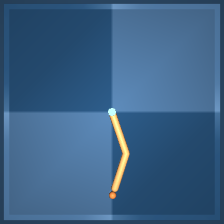}{30} & \qf{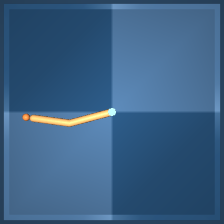}{40} & \qf{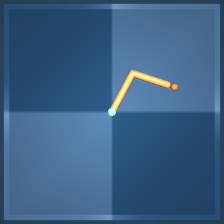}{50} & \\ \noalign{\vspace{3pt}}
\end{tabular}
\caption{\textbf{One test episode under a shift, planned by each method.} Frames from the same start toward the same goal image (right), with the simulator step in each corner. In each condition, the episode shown is the first in which ours succeeds and the fine-tuned backbone fails (28 such PushT episodes and 47 Reacher episodes out of 150). Ours reaches the goal at step 24 on PushT and 34 on Reacher. AdaJEPA and the fine-tuned backbone use all 50 steps without reaching it. On PushT, the goal is the pose of the grey block in the goal image, and the green shape is part of the scene. LeWM backbone.}
\label{fig:supp_qual}
\end{figure*}

%% file: figures/supp_visual.tex
\begin{figure*}[t]
\centering
\input{figures/lm_figstyle}
\setlength{\tabcolsep}{1.2pt}
\begin{tabular}{@{}cc@{\hspace{8pt}}cc@{\hspace{8pt}}cc@{}}
\multicolumn{2}{c}{\lmlab\textbf{PushT}} & \multicolumn{2}{c}{\lmlab\textbf{Reacher}} & \multicolumn{2}{c}{\lmlab\textbf{Cube}} \\
\includegraphics[width=0.158\linewidth]{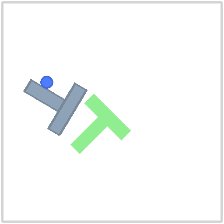} & \includegraphics[width=0.158\linewidth]{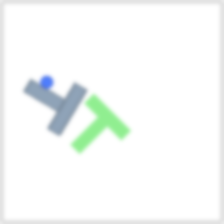} &
\includegraphics[width=0.158\linewidth]{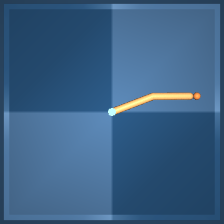} & \includegraphics[width=0.158\linewidth]{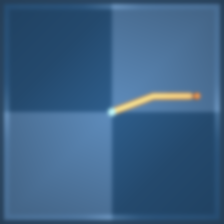} &
\includegraphics[width=0.158\linewidth]{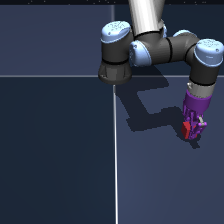} & \includegraphics[width=0.158\linewidth]{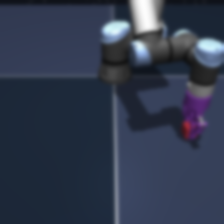} \\
{\lmnote clean} & {\lmnote blurred} & {\lmnote clean} & {\lmnote blurred} & {\lmnote clean} & {\lmnote blurred} \\
\end{tabular}
\caption{\textbf{A change in appearance.} The same frame of each environment, clean and blurred, one of the image changes we test.}
\label{fig:supp_visual}
\end{figure*}

%% file: sections/supp/ethics.tex
All experiments run in simulation or on recorded data, and no physical robot was operated for this work.
\lememoire allows a robot to plan in a world whose dynamics differ from training without changing its weights. It could make deployed robots more reliable when loads, surfaces or actuators change.
The same ability carries a risk. A robot that infers the dynamics wrongly, for example from a memory that still holds transitions of a world that has since changed (\cref{sec:supp_changing_world}), plans with a wrong model and may act unsafely.
Deployment should keep the safety limits of standard robot controllers, such as bounds on force and workspace and an emergency stop. Deployment can clear the memory when the predictor's error on recent transitions rises.

%% file: main.bbl
\begin{thebibliography}{47}
\providecommand{\natexlab}[1]{#1}
\providecommand{\url}[1]{\texttt{#1}}
\expandafter\ifx\csname urlstyle\endcsname\relax
  \providecommand{\doi}[1]{doi: #1}\else
  \providecommand{\doi}{doi: \begingroup \urlstyle{rm}\Url}\fi

\bibitem[Assran et~al.(2025)Assran, Bardes, Fan, Garrido, Howes, Komeili,
  Muckley, Rizvi, Roberts, Sinha, et~al.]{assran2025vjepa2}
Mahmoud Assran, Adrien Bardes, David Fan, Quentin Garrido, Russell Howes,
  Mojtaba Komeili, Matthew Muckley, Ammar Rizvi, Claire Roberts, Koustuv Sinha,
  et~al.
\newblock {V-JEPA} 2: Self-supervised video models enable understanding,
  prediction and planning.
\newblock \emph{arXiv preprint arXiv:2506.09985}, 2025.

\bibitem[Dai et~al.(2026)Dai, Fang, Gao, Guo, Rong, Huang, and
  Li]{dai2026sprii}
Ji Dai, Quan Fang, Junyu Gao, Rongfeng Guo, Haoyan Rong, Yiping Huang, and
  Yongxi Li.
\newblock Shaping persistent representations from independent interactions.
\newblock \emph{arXiv preprint arXiv:2609.34604}, 2026.

\bibitem[Doshi-Velez and Konidaris(2016)]{doshivelez2016hipmdp}
Finale Doshi-Velez and George Konidaris.
\newblock Hidden parameter {Markov} decision processes: A semiparametric
  regression approach for discovering latent task parametrizations.
\newblock In \emph{International Joint Conference on Artificial Intelligence
  (IJCAI)}, 2016.

\bibitem[Duan et~al.(2016)Duan, Schulman, Chen, Bartlett, Sutskever, and
  Abbeel]{duan2016rl2}
Yan Duan, John Schulman, Xi Chen, Peter~L. Bartlett, Ilya Sutskever, and Pieter
  Abbeel.
\newblock {RL$^2$}: Fast reinforcement learning via slow reinforcement
  learning.
\newblock \emph{arXiv preprint arXiv:1611.02779}, 2016.

\bibitem[Evans et~al.(2022)Evans, Thankaraj, and Pinto]{evans2022iida}
Ben Evans, Abitha Thankaraj, and Lerrel Pinto.
\newblock Context is everything: Implicit identification for dynamics
  adaptation.
\newblock In \emph{IEEE International Conference on Robotics and Automation
  (ICRA)}, 2022.

\bibitem[Ghosh et~al.(2021)Ghosh, Gupta, Reddy, Fu, Devin, Eysenbach, and
  Levine]{ghosh2019gcsl}
Dibya Ghosh, Abhishek Gupta, Ashwin Reddy, Justin Fu, Coline Devin, Benjamin
  Eysenbach, and Sergey Levine.
\newblock Learning to reach goals via iterated supervised learning.
\newblock In \emph{International Conference on Learning Representations
  (ICLR)}, 2021.

\bibitem[Hafner et~al.(2019)Hafner, Lillicrap, Fischer, Villegas, Ha, Lee, and
  Davidson]{hafner2019planet}
Danijar Hafner, Timothy Lillicrap, Ian Fischer, Ruben Villegas, David Ha,
  Honglak Lee, and James Davidson.
\newblock Learning latent dynamics for planning from pixels.
\newblock In \emph{International Conference on Machine Learning (ICML)}, 2019.

\bibitem[Hallak et~al.(2015)Hallak, Di~Castro, and Mannor]{hallak2015cmdp}
Assaf Hallak, Dotan Di~Castro, and Shie Mannor.
\newblock Contextual {Markov} decision processes.
\newblock \emph{arXiv preprint arXiv:1502.02259}, 2015.

\bibitem[Hewitt and Liang(2019)]{hewitt2019control}
John Hewitt and Percy Liang.
\newblock Designing and interpreting probes with control tasks.
\newblock In \emph{Conference on Empirical Methods in Natural Language
  Processing (EMNLP)}, 2019.

\bibitem[Khazatsky et~al.(2024)Khazatsky, Pertsch, Nair, Balakrishna, Dasari,
  Karamcheti, Nasiriany, Srirama, Chen, Ellis, et~al.]{khazatsky2024droid}
Alexander Khazatsky, Karl Pertsch, Suraj Nair, Ashwin Balakrishna, Sudeep
  Dasari, Siddharth Karamcheti, Soroush Nasiriany, Mohan~Kumar Srirama,
  Lawrence~Yunliang Chen, Kirsty Ellis, et~al.
\newblock {DROID}: A large-scale in-the-wild robot manipulation dataset.
\newblock In \emph{Robotics: Science and Systems (RSS)}, 2024.

\bibitem[Killian et~al.(2017)Killian, Daulton, Konidaris, and
  Doshi-Velez]{killian2017hipmdp}
Taylor Killian, Samuel Daulton, George Konidaris, and Finale Doshi-Velez.
\newblock Robust and efficient transfer learning with hidden parameter {Markov}
  decision processes.
\newblock In \emph{Advances in Neural Information Processing Systems
  (NeurIPS)}, 2017.

\bibitem[Kumar et~al.(2021)Kumar, Fu, Pathak, and Malik]{kumar2021rma}
Ashish Kumar, Zipeng Fu, Deepak Pathak, and Jitendra Malik.
\newblock {RMA}: Rapid motor adaptation for legged robots.
\newblock In \emph{Robotics: Science and Systems (RSS)}, 2021.

\bibitem[LeCun(2022)]{lecun2022path}
Yann LeCun.
\newblock A path towards autonomous machine intelligence.
\newblock OpenReview, version 0.9.2, 2022.
\newblock \url{https://openreview.net/forum?id=BZ5a1r-kVsf}.

\bibitem[Lee et~al.(2020)Lee, Seo, Lee, Lee, and Shin]{lee2020cadm}
Kimin Lee, Younggyo Seo, Seunghyun Lee, Honglak Lee, and Jinwoo Shin.
\newblock Context-aware dynamics model for generalization in model-based
  reinforcement learning.
\newblock In \emph{International Conference on Machine Learning (ICML)}, 2020.

\bibitem[Liu et~al.(2026)Liu, Sun, Baker, Balestriero, and Sous]{liu2026sgjepa}
Andy~Zeyi Liu, Haoran Sun, Lucas Baker, Randall Balestriero, and John Sous.
\newblock {Semigroup-JEPA}: Latent dynamics consistency for zero-shot physics
  generalization.
\newblock \emph{arXiv preprint arXiv:2609.10464}, 2026.

\bibitem[Luo et~al.(2026)Luo, He, Dai, Nam, Fu, Chai, Fazeli, and
  Park]{luo2026physicalworld}
Yunhao Luo, Wentao He, Yinpei Dai, Hyelin Nam, Hongze Fu, Joyce Chai, Nima
  Fazeli, and Jeong~Joon Park.
\newblock {PhysiCalWorld}: In-context physics calibration for world models.
\newblock \emph{arXiv preprint arXiv:2610.30125}, 2026.

\bibitem[Maes et~al.(2026)Maes, Le~Lidec, Scieur, LeCun, and
  Balestriero]{maes2026lewm}
Lucas Maes, Quentin Le~Lidec, Damien Scieur, Yann LeCun, and Randall
  Balestriero.
\newblock {LeWorldModel}: Stable end-to-end joint-embedding predictive
  architecture from pixels.
\newblock In \emph{Advances in Neural Information Processing Systems
  (NeurIPS)}, 2026.

\bibitem[Mete et~al.(2026)Mete, Kalathil, and Kumar]{mete2026optta}
Akshay Mete, Dileep Kalathil, and P.~R. Kumar.
\newblock Adapt for planning, not just prediction: Optimistic test-time world
  model adaptation.
\newblock \emph{arXiv preprint arXiv:2610.35636}, 2026.

\bibitem[Nagabandi et~al.(2019)Nagabandi, Clavera, Liu, Fearing, Abbeel,
  Levine, and Finn]{nagabandi2019learning}
Anusha Nagabandi, Ignasi Clavera, Simin Liu, Ronald~S. Fearing, Pieter Abbeel,
  Sergey Levine, and Chelsea Finn.
\newblock Learning to adapt in dynamic, real-world environments through
  meta-reinforcement learning.
\newblock In \emph{International Conference on Learning Representations
  (ICLR)}, 2019.

\bibitem[Palafox and Fridovich-Keil(2026)]{palafox2026claw}
Fernando Palafox and David Fridovich-Keil.
\newblock Amortized low-rank adaptation for model-based reinforcement learning.
\newblock \emph{arXiv preprint arXiv:2609.12278}, 2026.

\bibitem[Pandya(2026)]{pandya2026epm}
Vedant Pandya.
\newblock {EPM-JEPA}: Operator-side experience modulation in {JEPA}-family
  world models.
\newblock \emph{arXiv preprint arXiv:2606.12979}, 2026.

\bibitem[Park et~al.(2024)Park, Choe, and Veitch]{park2024linear}
Kiho Park, Yo~Joong Choe, and Victor Veitch.
\newblock The linear representation hypothesis and the geometry of large
  language models.
\newblock In \emph{International Conference on Machine Learning (ICML)}, 2024.

\bibitem[Park et~al.(2025)Park, Frans, Eysenbach, and Levine]{ogbench_park2025}
Seohong Park, Kevin Frans, Benjamin Eysenbach, and Sergey Levine.
\newblock {OGBench}: Benchmarking offline goal-conditioned {RL}.
\newblock In \emph{International Conference on Learning Representations
  (ICLR)}, 2025.

\bibitem[Peebles and Xie(2023)]{peebles2023dit}
William Peebles and Saining Xie.
\newblock Scalable diffusion models with transformers.
\newblock In \emph{Proceedings of the IEEE/CVF International Conference on
  Computer Vision (ICCV)}, 2023.

\bibitem[Peng et~al.(2018)Peng, Andrychowicz, Zaremba, and Abbeel]{peng2018dr}
Xue~Bin Peng, Marcin Andrychowicz, Wojciech Zaremba, and Pieter Abbeel.
\newblock Sim-to-real transfer of robotic control with dynamics randomization.
\newblock In \emph{IEEE International Conference on Robotics and Automation
  (ICRA)}, 2018.

\bibitem[Prasanna et~al.(2024)Prasanna, Farid, Rajan, and
  Biedenkapp]{prasanna2024crssm}
Sai Prasanna, Karim Farid, Raghu Rajan, and Andr{\'e} Biedenkapp.
\newblock Dreaming of many worlds: Learning contextual world models aids
  zero-shot generalization.
\newblock \emph{Reinforcement Learning Journal}, 2024.

\bibitem[Rakelly et~al.(2019)Rakelly, Zhou, Quillen, Finn, and
  Levine]{rakelly2019pearl}
Kate Rakelly, Aurick Zhou, Deirdre Quillen, Chelsea Finn, and Sergey Levine.
\newblock Efficient off-policy meta-reinforcement learning via probabilistic
  context variables.
\newblock In \emph{International Conference on Machine Learning (ICML)}, 2019.

\bibitem[R{\"o}der et~al.(2025)R{\"o}der, Benad, Eppe, and
  Banerjee]{roeder2025dali}
Frank R{\"o}der, Jan Benad, Manfred Eppe, and Pradeep~Kr. Banerjee.
\newblock Dynamics-aligned latent imagination in contextual world models for
  zero-shot generalization.
\newblock In \emph{Advances in Neural Information Processing Systems
  (NeurIPS)}, 2025.

\bibitem[Seo et~al.(2020)Seo, Lee, Clavera, Kurutach, Shin, and
  Abbeel]{seo2020tmcl}
Younggyo Seo, Kimin Lee, Ignasi Clavera, Thanard Kurutach, Jinwoo Shin, and
  Pieter Abbeel.
\newblock Trajectory-wise multiple choice learning for dynamics generalization
  in reinforcement learning.
\newblock In \emph{Advances in Neural Information Processing Systems
  (NeurIPS)}, 2020.

\bibitem[Smirnov et~al.(2026)Smirnov, Ennadir, Pertoft, Hjaltason, and
  Karimi]{smirnov2026headroom}
Oleg Smirnov, Sofiane Ennadir, John Pertoft, Bjartur Hjaltason, and Sara
  Karimi.
\newblock When does dynamics randomization test context inference?
\newblock In \emph{Do Robots Need World Models? Workshop at the Conference on
  Robot Learning (CoRL)}, 2026.

\bibitem[Sobal et~al.(2025)Sobal, Zhang, Cho, Balestriero, Rudner, and
  LeCun]{sobal2025pldm}
Vlad Sobal, Wancong Zhang, Kyunghyun Cho, Randall Balestriero, Tim G.~J.
  Rudner, and Yann LeCun.
\newblock Learning from reward-free offline data: A case for planning with
  latent dynamics models.
\newblock In \emph{Advances in Neural Information Processing Systems
  (NeurIPS)}, 2025.

\bibitem[Son et~al.(2025)Son, Lee, and Kim]{son2025dicp}
Jaehyeon Son, Soochan Lee, and Gunhee Kim.
\newblock Distilling reinforcement learning algorithms for in-context
  model-based planning.
\newblock In \emph{International Conference on Learning Representations
  (ICLR)}, 2025.

\bibitem[Soni et~al.(2026)Soni, Sehgal, Dave, and Rueckert]{soni2026sandwich}
Krishnam Soni, Aditya Sehgal, Vedant Dave, and Elmar Rueckert.
\newblock {Sandwich-Residuals}: Parameter-efficient test-time adaptation of
  world models.
\newblock \emph{arXiv preprint arXiv:2609.21740}, 2026.

\bibitem[Sun and Zhu(2026)]{sun2026memwm}
Yuwei Sun and Siyu Zhu.
\newblock {MemWM}: Associative memory-enhanced latent world models for
  long-horizon planning.
\newblock \emph{arXiv preprint arXiv:2610.19683}, 2026.

\bibitem[Terver et~al.(2026)Terver, Yang, Ponce, Bardes, and
  LeCun]{terver2026drives}
Basile Terver, Tsung-Yen Yang, Jean Ponce, Adrien Bardes, and Yann LeCun.
\newblock What drives success in physical planning with joint-embedding
  predictive world models?
\newblock \emph{Transactions on Machine Learning Research (TMLR)}, 2026.

\bibitem[Tobin et~al.(2017)Tobin, Fong, Ray, Schneider, Zaremba, and
  Abbeel]{tobin2017dr}
Josh Tobin, Rachel Fong, Alex Ray, Jonas Schneider, Wojciech Zaremba, and
  Pieter Abbeel.
\newblock Domain randomization for transferring deep neural networks from
  simulation to the real world.
\newblock In \emph{IEEE/RSJ International Conference on Intelligent Robots and
  Systems (IROS)}, 2017.

\bibitem[Wang et~al.(2026{\natexlab{a}})Wang, Chen, Zhong, Zheng, Shao, Yu,
  Liu, Wang, Ding, Cao, and Kang]{wang2026iclwm}
Fan Wang, Zhiyuan Chen, Yuxuan Zhong, Sunjian Zheng, Pengtao Shao, Bo Yu,
  Shaoshan Liu, Jianan Wang, Ning Ding, Yang Cao, and Yu Kang.
\newblock Context and diversity matter: The emergence of in-context learning in
  world models.
\newblock In \emph{International Conference on Learning Representations
  (ICLR)}, 2026{\natexlab{a}}.

\bibitem[Wang et~al.(2022)Wang, Mu, Li, Zhang, Zhao, Zhuang, Luo, Wang, and
  Hao]{wang2022protocad}
Junjie Wang, Yao Mu, Dong Li, Qichao Zhang, Dongbin Zhao, Yuzheng Zhuang, Ping
  Luo, Bin Wang, and Jianye Hao.
\newblock Prototypical context-aware dynamics generalization for
  high-dimensional model-based reinforcement learning.
\newblock \emph{arXiv preprint arXiv:2211.12774}, 2022.

\bibitem[Wang et~al.(2026{\natexlab{b}})Wang, Shi, Fei, Fu, Ji, Gong, and
  Qiu]{wang2026icwm}
Siyin Wang, Junhao Shi, Senyu Fei, Zhaoyang Fu, Li Ji, Jingjing Gong, and
  Xipeng Qiu.
\newblock In-context world modeling for robotic control.
\newblock \emph{arXiv preprint arXiv:2606.26025}, 2026{\natexlab{b}}.

\bibitem[Wang et~al.(2026{\natexlab{c}})Wang, Bounou, LeCun, and
  Ren]{wang2026adajepa}
Ying Wang, Oumayma Bounou, Yann LeCun, and Mengye Ren.
\newblock {AdaJEPA}: An adaptive latent world model.
\newblock \emph{arXiv preprint arXiv:2606.32026}, 2026{\natexlab{c}}.

\bibitem[Wang et~al.(2026{\natexlab{d}})Wang, Bounou, Zhou, Balestriero,
  Rudner, LeCun, and Ren]{wang2026straightening}
Ying Wang, Oumayma Bounou, Gaoyue Zhou, Randall Balestriero, Tim G.~J. Rudner,
  Yann LeCun, and Mengye Ren.
\newblock Temporal straightening for latent planning.
\newblock In \emph{International Conference on Machine Learning (ICML)},
  2026{\natexlab{d}}.

\bibitem[Yu et~al.(2017)Yu, Tan, Liu, and Turk]{yu2017uposi}
Wenhao Yu, Jie Tan, C.~Karen Liu, and Greg Turk.
\newblock Preparing for the unknown: Learning a universal policy with online
  system identification.
\newblock In \emph{Robotics: Science and Systems (RSS)}, 2017.

\bibitem[Zhang et~al.(2026)Zhang, Ye, Agarwal, Mahjoub, Moradi~Pari, Khashabi,
  Shu, and Tadiparthi]{zhang2026jepattt}
Zheyuan Zhang, Suyu Ye, Nakul Agarwal, Hossein~Nourkhiz Mahjoub, Ehsan
  Moradi~Pari, Daniel Khashabi, Tianmin Shu, and Vaishnav Tadiparthi.
\newblock {JEPA-TTT}: Persistent test-time training of latent world models for
  planning under dynamics shifts.
\newblock \emph{arXiv preprint arXiv:2610.00722}, 2026.

\bibitem[Zholus et~al.(2026)Zholus, Beltran-Velez, Yuan, Chandar, Nagarajan,
  Severo, Sinha, Drozdzal, Romero~Soriano, Bohg, Ballas, and
  Assran]{zholus2026robojepa}
Artem Zholus, Nicolas Beltran-Velez, Jianhao Yuan, Sarath Chandar, Tushar
  Nagarajan, Daniel Severo, Koustuv Sinha, Michal Drozdzal, Adriana
  Romero~Soriano, Jeannette Bohg, Nicolas Ballas, and Mahmoud Assran.
\newblock {RoboJEPA}: Scaling robotic latent world models.
\newblock \emph{arXiv preprint arXiv:2610.10515}, 2026.

\bibitem[Zhou et~al.(2026)Zhou, Hong, Huang, Jiang, Huang, Held, Guizilini,
  Wang, Fei-Fei, Wu, Guibas, and Choi]{zhou2026t2z}
Chenyuan Zhou, Yining Hong, Huang Huang, Yunfan Jiang, Wenlong Huang, David
  Held, Vitor~Campagnolo Guizilini, Yue Wang, Li Fei-Fei, Jiajun Wu, Leonidas
  Guibas, and Yejin Choi.
\newblock {T2-z}: Test-time optimization of environment latents for robotic
  world models.
\newblock \emph{arXiv preprint arXiv:2610.22364}, 2026.

\bibitem[Zhou et~al.(2025)Zhou, Pan, LeCun, and Pinto]{zhou2025dinowm}
Gaoyue Zhou, Hengkai Pan, Yann LeCun, and Lerrel Pinto.
\newblock {DINO-WM}: World models on pre-trained visual features enable
  zero-shot planning.
\newblock In \emph{International Conference on Machine Learning (ICML)}, 2025.

\bibitem[Zintgraf et~al.(2019)Zintgraf, Shiarlis, Kurin, Hofmann, and
  Whiteson]{zintgraf2019cavia}
Luisa Zintgraf, Kyriacos Shiarlis, Vitaly Kurin, Katja Hofmann, and Shimon
  Whiteson.
\newblock Fast context adaptation via meta-learning.
\newblock In \emph{International Conference on Machine Learning (ICML)}, 2019.

\end{thebibliography}
